%% file: main.tex
\documentclass{article} 
\usepackage{iclr2023_conference,times}

\input{preamble/packages}
\input{preamble/definitions}
\input{preamble/math}

\title{Learning to Estimate Shapley Values with \\ Vision Transformers}

\author{Ian Covert\thanks{Equal contribution.}, { }Chanwoo Kim$^*$ \& Su-In Lee \\
Paul G. Allen School of Computer Science \& Engineering\\
University of Washington\\
\texttt{\{icovert,chanwkim,suinlee\}@cs.washington.edu} \\
}

\iclrfinalcopy 
\begin{document}

\maketitle

\input{sections/abstract}

\section{Introduction} \label{sec:introduction}
\input{sections/introduction}

\section{Related work} \label{sec:related}
\input{sections/related}

\section{Background} \label{sec:background}
\input{sections/background}

\section{Evaluating vision transformers with partial information} \label{sec:removal}
\input{sections/removal}

\section{Learning to estimate Shapley values} \label{sec:estimation}
\input{sections/estimation}

\section{Experiments} \label{sec:experiments}
\input{sections/experiments}

\section{Conclusion} \label{sec:conclusion}
\input{sections/conclusion}

\section*{Acknowledgements}

We thank Mukund Sudarshan, Neil Jethani, Chester Holtz and the Lee Lab for helpful discussions. This work was funded by NSF DBI-1552309 and DBI-1759487, NIH R35-GM-128638 and R01-NIA-AG-061132.

\bibliography{main}


\clearpage
\appendix

\input{sections/appendix}

\end{document}

%% file: preamble/packages.tex
\usepackage{amssymb, amsmath, amsthm}
\usepackage{mathtools}
\usepackage{xcolor}

\usepackage{microtype}
\usepackage[english]{babel}
\usepackage[parfill]{parskip}
\usepackage{enumitem}
\usepackage{wrapfig}
\usepackage{multicol}
\usepackage{float}
\usepackage{adjustbox}
\usepackage[utf8]{inputenc} 

\usepackage[T1]{fontenc}
\usepackage{amsfonts}
\usepackage{bm}
\usepackage{bbm}
\usepackage{times}

\usepackage{fancyhdr}

\usepackage{graphicx}
\usepackage{wrapfig}
\usepackage{subcaption}
\usepackage{tikz}
\usepackage{forest}
\usetikzlibrary{arrows.meta}

\usepackage{booktabs}
\usepackage{caption}

\usepackage[ruled,vlined]{algorithm2e}
\SetKwInput{KwInput}{Input}
\SetKwInput{KwOutput}{Output}
\usepackage{listings}
\usepackage{fancyvrb}
\fvset{fontsize=\normalsize}

\renewcommand{\bibname}{References}
\renewcommand{\bibsection}{\subsubsection*{\bibname}}

\usepackage[acronym,smallcaps,nowarn]{glossaries}
\glsdisablehyper

\usepackage[linktoc=all]{hyperref}
\usepackage[all]{hypcap}
\hypersetup{linkcolor=black}

\usepackage{url}
\usepackage[nameinlink]{cleveref}

\usepackage{multirow}
\usepackage{nicefrac}


%% file: preamble/definitions.tex
\newtheorem{lemma}{Lemma}

\newtheorem{thm}{Theorem}

\crefname{thmdef}{definition}{definitions}
\crefname{lemma}{lemma}{lemmas}
\crefname{prop}{proposition}{propositions}
\crefname{algorithm}{algorithm}{algorithms}

\newif\ifprintcomments

\definecolor{deepblue}{HTML}{20639B}
\definecolor{seagreen}{HTML}{3CAEA3}
\definecolor{niceyellow}{HTML}{D19654}
\definecolor{nicegreen}{HTML}{6a040f}
\definecolor{niceblue}{HTML}{277DA1}

\newlength\myindent
\forestset{
    .style={
        for tree={
            base=bottom,
            child anchor=north,
            align=center,
            s sep+=1cm,
    straight edge/.style={
        edge path={\noexpand\path[\forestoption{edge},thick,-{Latex}] 
        (!u.parent anchor) -- (.child anchor);}
    },
    if n children={0}
        {tier=word, draw, thick, rectangle}
        {draw, diamond, thick, aspect=2},
    if n=1{%
        edge path={\noexpand\path[\forestoption{edge},thick,-{Latex}] 
        (!u.parent anchor) -| (.child anchor) node[pos=.2, above] {Y};}
        }{
        edge path={\noexpand\path[\forestoption{edge},thick,-{Latex}] 
        (!u.parent anchor) -| (.child anchor) node[pos=.2, above] {N};}
        }
        }
    }
}

\makeatletter
\newcommand\footnoteref[1]{\protected@xdef\@thefnmark{\ref{#1}}\@footnotemark}
\makeatother

\newcommand\tablesize{\fontsize{8pt}{10pt}\selectfont}
\newcommand\figurespacing{\vspace{-0.7cm}}

%% file: preamble/math.tex
\def\eqref#1{equation~\ref{#1}}

\def\1{\bm{1}}

\def\rvm{{\mathbf{m}}}

\def\rvs{{\mathbf{s}}}

\def\rvx{{\mathbf{x}}}

\def\rvy{{\mathbf{y}}}
\def\rvz{{\mathbf{z}}}

\def\vzero{{\bm{0}}}
\def\vone{{\bm{1}}}

\DeclareMathAlphabet{\mathsfit}{\encodingdefault}{\sfdefault}{m}{sl}
\SetMathAlphabet{\mathsfit}{bold}{\encodingdefault}{\sfdefault}{bx}{n}

\newcommand{\R}{\mathbb{R}}

\newcommand{\softmax}{\mathrm{softmax}}

\newcommand{\KL}{D_{\mathrm{KL}}}

\newcommand{\E}{\mathbb{E}}
\newcommand{\phifast}{\phi_{\text{ViT}}}
\newcommand{\inv}{^{-1}}
\newcommand{\psh}{p_{\text{Sh}}}
\newcommand{\hin}{h}
\newcommand{\hout}{h'}
\newcommand{\rvQ}{\mathbf{Q}}
\newcommand{\rvK}{\mathbf{K}}
\newcommand{\rvV}{\mathbf{V}}
\newcommand{\rvA}{\mathbf{A}}
\newcommand{\rvS}{\mathbf{S}}

%% file: sections/abstract.tex
\begin{abstract}
    Transformers have become a default architecture in computer vision, but understanding what drives their predictions remains a challenging problem. Current explanation approaches rely on attention values or input gradients, but these provide a limited view
    of a model's dependencies.
    Shapley values offer a theoretically sound alternative, but their computational cost makes them impractical for large, high-dimensional models.
    In this work, we aim to make Shapley values practical for vision transformers (ViTs). To do so, we first leverage an attention masking approach to evaluate ViTs with partial information, and we then
    develop a procedure to generate
    Shapley value explanations via a
    separate, learned
    explainer model.
    Our experiments compare Shapley values to many
    baseline
    methods (e.g., attention rollout, GradCAM, LRP), and we find that our approach provides more accurate explanations than
    existing
    methods
    for ViTs.
\end{abstract}

%% file: sections/introduction.tex

Transformers~\citep{vaswani2017attention} were originally introduced for NLP, but in recent years they
have been
successfully adapted to a variety of other domains \citep{wang2020transformer, jumper2021highly}.
In computer vision, transformer-based models are now used for problems including image classification, object detection and semantic segmentation \citep{dosovitskiy2020image, touvron2021training, liu2021swin}, and they achieve state-of-the-art performance in many tasks \citep{wortsman2022model}.
The growing use of transformers in computer vision motivates the question of what drives their predictions:
understanding a complex model's dependencies is an important problem in many applications,
but the field has not settled on a solution for the transformer architecture.

Transformers are composed of alternating self-attention and fully-connected layers, where the self-attention operation associates attention
values
with every pair of tokens.
In vision transformers (ViTs) \citep{dosovitskiy2020image}, the tokens represent non-overlapping image patches, typically a total of $14 \times 14 = 196$ patches each of size $16 \times 16$. It is intuitive to view
attention values as indicators
of feature importance \citep{abnar2020quantifying, ethayarajh2021attention},
but
interpreting transformer attention in this way
is potentially misleading. Recent work has raised questions about
the validity of attention as explanation \citep{serrano2019attention, jain2019attention, chefer2021transformer}, arguing that it provides an incomplete picture of
a model's dependence on
each token.

If attention is not a reliable indicator of feature importance, then what is? We consider the perspective that transformers
are no
different from any other architecture,
and that we can explain their predictions using model-agnostic approaches
that are currently
used for other architectures.
Among these methods, Shapley values are a theoretically compelling approach
with feature importance scores
that are designed to satisfy many desirable properties \citep{shapley1953value, lundberg2017unified}.

The main challenge for Shapley values in the transformer context is calculating them efficiently, because a naive calculation has exponential running time in the number of patches. If Shapley values are poorly approximated, they are unlikely to reflect a model's true dependencies, but calculating them with high
accuracy is currently too slow to be practical. Thus, our work aims to make Shapley values practical for transformers, and for ViTs in particular. Our contributions include:



\begin{enumerate}[leftmargin=20pt]
    \item We investigate several approaches for withholding input features from vision transformers, which is a key operation for computing Shapley values.
    We find that ViTs can accommodate missing image patches by masking attention values
    for held-out tokens,
    and
    that training
    with random masking
    is important for models to
    properly handle
    partial information.
    \item We develop a learning-based approach to estimate Shapley values efficiently and accurately.
    Our approach
    involves fine-tuning an existing ViT
    using a loss function designed specifically for Shapley values \citep{jethani2021fastshap}, and
    we prove that our loss bounds the
    estimation error without requiring ground truth values during training.
    Once trained,
    our explainer model
    provides a significant speedup over
    methods
    like
    KernelSHAP \citep{lundberg2017unified}.
    \item Our experiments compare the Shapley value-based approach to several groups of competing methods:
    attention-based, gradient-based and removal-based explanations. We find that our approach provides the best overall performance, correctly identifying influential and non-influential
    patches for both target and non-target classes. We verify this using three image datasets,
    and the results are consistent across multiple metrics.
\end{enumerate}

Overall, our work shows that Shapley values can be made
practical for vision transformers, and that they provide a compelling alternative to current attention- and gradient-based approaches.

%% file: sections/related.tex
Understanding neural network predictions is a challenging problem that has been actively researched for the last decade \citep{simonyan2013deep, zeiler2014visualizing, ribeiro2016should}. 
We focus on \textit{feature attribution}, or identifying the specific input features that influence a prediction, but prior work has also considered other
problems \citep{olah2017feature, kim2018interpretability}.
The various techniques that have been developed can be grouped into several categories, which we describe below.

\paragraph{Attention-based explanations}
Transformers use self-attention to associate weights with each pair of tokens \citep{vaswani2017attention}, and a natural idea is to assess which tokens receive the most attention
\citep{clark2019does, rogers2020primer, vig2020bertology}. There are several versions of this approach, including \textit{attention rollout} and \textit{attention flow}  \citep{abnar2020quantifying}, which analyze attention across multiple layers. Attention is a popular interpretation tool, but it is only one component in a sequence of nonlinear operations that provides an incomplete picture of a model's dependencies \citep{serrano2019attention, jain2019attention, wiegreffe2019attention}, and direct usage of attention weights
has not been shown to perform well in vision tasks \citep{chefer2021transformer}.

\paragraph{Gradient-based methods}
For other deep learning models such as CNNs, gradient-based explanations are a popular family of approaches. There are many variations on the idea of calculating input gradients, including methods that modify the input (e.g., SmoothGrad, IntGrad) \citep{smilkov2017smoothgrad, sundararajan2017axiomatic, xu2020attribution}, operate at intermediate network layers (GradCAM) \citep{selvaraju2017grad}, or design modified backpropagation rules (e.g., LRP, DeepLift) \citep{bach2015pixel, shrikumar2016not, chefer2021transformer}. Although they are efficient to compute for arbitrary network architectures, gradient-based explanations achieve mixed results in quantitative benchmarks, including for object localization and the removal of influential features \citep{petsiuk2018rise, hooker2019benchmark, saporta2021benchmarking, jethani2021fastshap},
and they have been shown to be insensitive to the randomization of model parameters \citep{adebayo2018sanity}.

\paragraph{Removal-based explanations}
Finally, \textit{removal-based explanations} are those that quantify feature importance by explicitly withholding inputs from the model \citep{covert2021explaining}. For models that require all features to make predictions, several options for removing features include setting them to default values \citep{zeiler2014visualizing}, sampling replacement values \citep{agarwal2020explaining} and blurring images \citep{fong2017interpretable}. These methods work with any model type, but they tend to be slow because they require making many predictions. The simplest approach of removing individual features (known as \textit{leave-one-out}, \citealt{ethayarajh2021attention}) is relatively fast, but the computational cost increases as we examine more feature subsets.

Shapley values \citep{shapley1953value} are an influential approach within the removal-based explanation framework. By examining all feature subsets, they provide a nuanced view of each feature's influence and satisfy many desirable properties \citep{lundberg2017unified}. They are approximated in practice using methods like TreeSHAP and KernelSHAP \citep{lundberg2020local, covert2021improving}, but these
approaches either are not applicable or do not scale to large ViTs. Recent work highlighted a connection between attention flow and Shapley values \citep{ethayarajh2021attention}, but this approach is fundamentally different from SHAP \citep{lundberg2017unified}: attention flow treats each feature's influence on the model as strictly additive, which is computationally convenient but
fails to represent feature interactions. Our work instead focuses on the original formulation \citep{lundberg2017unified} and
aims to make Shapley values based on feature removal practical for ViTs.

%% file: sections/background.tex
Here, we define notation used throughout the paper and briefly introduce Shapley values.

\subsection{Notation}

Our focus is vision transformers trained for classification tasks, where $\rvx \in \R^{224 \times 224 \times 3}$ denotes an image and $\rvy \in \{1, \ldots, K\}$ denotes the class. We write the image patches as $\rvx = (\rvx_1, \ldots, \rvx_d)$, where ViTs typically have $\rvx_i \in \R^{16 \times 16 \times 3}$ and $d = 196$. The model is given by $f(\rvx; \eta) \in [0, 1]^K$ and $f_y(\rvx; \eta) \in [0, 1]$ represents the probability for the $y$th class. Shapley values involve feature subsets, so we use $s \in \{0, 1\}^d$ to denote a subset of indices and $\rvx_s = \{\rvx_i : s_i = 1\}$ a subset of image patches. We also use $\vzero$ and $\vone \in \R^d$ to denote vectors of zeros and ones, and $e_i \in \R^d$ is a vector with a one in the $i$th position and zeros elsewhere. Finally, bold symbols $\rvx, \rvy$ are random variables, $x, y$ are possible values, and $p(\rvx, \rvy)$ denotes the data distribution.




\subsection{Shapley values}

Shapley values were developed in game theory for allocating
credit in coalitional games \citep{shapley1953value}. A coalitional game is represented by a set function, and the value for each subset indicates the profit achieved when the corresponding players participate. Given a game with $d$ players, or a set function $v: \{0, 1\}^d \mapsto \R$, the Shapley values are denoted by $\phi_1(v), \ldots, \phi_d(v) \in \R$ for each player, and the value $\phi_i(v)$ for the $i$th player is defined as follows:
\begin{equation}
    \phi_i(v) = \frac{1}{d} \sum_{s : s_i = 0} \binom{d - 1}{\vone^\top s}\inv \big( v(s + e_i) - v(s) \big). \label{eq:shapley}
\end{equation}
Intuitively, \cref{eq:shapley} represents the change in profit from introducing the $i$th player, averaged across all possible subsets to which $i$ can be added. Shapley values are defined in this way to satisfy many reasonable properties: for example, the credits sum to the value when all players participate, players with equivalent contributions receive equal credit, and players with no contribution receive zero credit \citep{shapley1953value}. These properties make Shapley values attractive in many settings: they have been applied with coalitional games that represent a model's prediction given a subset of features (SHAP) \citep{vstrumbelj2010efficient, lundberg2017unified}, as well as several other use-cases in machine learning \citep{ghorbani2019data, ghorbani2020neuron, covert2020understanding}.

There are two main challenges when using Shapley values to explain individual predictions \citep{chen2022algorithms}. The first is properly withholding feature information, and we explore how to address this challenge in the ViT context (\Cref{sec:removal}). The second is calculating Shapley values efficiently, because their computation scales exponentially with the number of inputs $d$. Traditionally, they are approximated using sampling-based estimators like KernelSHAP \citep{castro2009polynomial, vstrumbelj2010efficient, lundberg2017unified}, but we build on a more efficient learning-based approach (FastSHAP) recently introduced by \cite{jethani2021fastshap} (\Cref{sec:estimation}).

%% file: sections/removal.tex
The basic idea behind Shapley values, as well as other removal-based explanations \citep{covert2021explaining}, is to evaluate the model with partial feature information and analyze how a prediction changes. Most models need values for all the features to make predictions, so in practice we require a mechanism to represent feature removal. For example, we can set held-out image regions to zero, or we can average the prediction across randomly sampled replacement values.

With vision transformers, the options for removing features are slightly different. Recent work has demonstrated the robustness of ViTs to randomly zeroed pixel values \citep{naseer2021intriguing}, but the self-attention operation enables a more elegant approach: we can simply ignore tokens for image patches we wish to remove \citep{jain2021missingness}. We achieve this by masking attention values at each self-attention layer, or setting them to a large negative value before applying the softmax operation (see Appendix~\ref{app:masking}). This resembles causal attention masking in transformer language models like
GPT-3 \citep{brown2020language}, but we use masking for a different purpose. Alternatively, we could use a unique token value as in masked language models such as BERT \citep{devlin2018bert}, which would involve simply setting held-out tokens to the mask value.

Using this attention masking approach, we can evaluate a ViT model $f(\rvx; \eta)$ given subsets of image patches, denoted by $\rvx_\rvs$.
However, because these partial inputs represent off-manifold examples, the predictions with partial information may not behave as desired.
We have two options to correct this: 1)~we can ensure that the
model is trained with random masking, or 2)~we can fine-tune the model to encourage sensible behavior with missing patches.
The first option is more direct, but it does not allow us to explain models trained without masking. For the latter option, we can create an updated model denoted by $g(\rvx_\rvs; \beta)$ that we fine-tune using the following loss,
\begin{equation}
    \min_\beta \quad \E_{p(\rvx)}\E_{p(\rvs)} 
       \Big[ \KL \big( f(\rvx; \eta) \mid\mid g(\rvx_\rvs; \beta) \big) \Big], \label{eq:surrogate}
\end{equation}
where $p(\rvs)$ is a distribution over subsets. In practice, we sample the cardinality $\rvm = \vone^\top \rvs$ from $\rvm \sim \text{Unif}(0, d)$ and then sample $\rvm$ patches uniformly at random. Intuitively, \cref{eq:surrogate} encourages $g(\rvx_\rvs; \beta)$ to preserve the original model's predictions even with missing features. We use this loss because it satisfies the desirable property that the optimal model $g(\rvx_\rvs; \beta^*)$ outputs the expected prediction given the available information \citep{covert2021explaining}, or
\begin{equation}
    g(x_s; \beta^*) = \E[f(\rvx; \eta) \mid \rvx_s = x_s].
\end{equation}
Note that this represents a best-effort prediction, because if $f(\rvx; \eta) = p(\rvy \mid \rvx)$ then we have $g(\rvx_s; \beta^*) = p(\rvy \mid \rvx_s)$. Similarly, in the case where $f(\rvx_s; \eta)$ is trained directly with random masking, the training process estimates $f(\rvx_s; \eta) \approx p(\rvy \mid \rvx_s)$ (see \Cref{app:masked_training}). We refer to the fine-tuned model $g(\rvx_\rvs; \beta)$ as a \textit{surrogate}, following the naming in prior work \citep{frye2020shapley}. Whether we use the original model or a version fine-tuned with random masking, our attention masking approach enables us to probe how individual predictions change as we remove groups of image patches.

%% file: sections/estimation.tex
Given our approach for evaluating ViTs with partial information, we can use Shapley values to identify influential image patches for an input $x$ and class $y$. This involves evaluating the model with many feature subsets $x_s$, so we define a coalitional game $v_{xy}(s) = g_y(x_s; \beta)$. Alternatively, if we use a model trained with masking, we can define the coalitional game as $v_{xy}(s) = f_y(x_s; \eta)$.
Common Shapley value approximations are based on sampling feature permutations \citep{castro2009polynomial, vstrumbelj2010efficient} or fitting a weighted least squares model \citep{lundberg2017unified, covert2021improving}, but these can require hundreds or thousands of model evaluations to explain a single prediction.\footnote{The number of model evaluations depends on how fast the estimators converge, and we find that KernelSHAP requires >100,000 samples to converge for ViTs (\Cref{app:results}).} Instead, we develop a learning-based estimation approach for ViTs.

Our goal is to obtain an explainer model that estimates Shapley values directly. To do so, we train a new vision transformer $\phifast(\rvx, \rvy; \theta) \in \R^d$ that outputs approximate Shapley values for an input-output pair $(x, y)$ in a single forward pass. Crucially, rather than training the model using a dataset of ground truth Shapley value explanations, we train it by minimizing the following objective,

\vskip -0.15in
\begin{align}
    \mathcal{L}(\theta) =\; &\E_{p(\rvx, \rvy)}
    \E_{\psh(\rvs)} \Big[ \big( v_{\rvx \rvy}(\rvs) - v_{\rvx \rvy}(\vzero) - \rvs^\top \phifast(\rvx, \rvy; \theta) \big)^2 \Big] \label{eq:objective} \\
    &\text{s.t.} \quad \vone^\top \phifast(x, y; \theta) = v_{x y}(\vone) - v_{x y}(\vzero) \quad \forall \; (x, y), \nonumber
\end{align}

where $\psh(\rvs)$ is a distribution defined as $\psh(s) \propto (\vone^\top s - 1)!(d - \vone^\top s - 1)!$ for $0 < \vone^\top s < d$ and $\psh(\vone) = \psh(\vzero) = 0$. Intuitively, \cref{eq:objective} encourages the explainer model to output feature scores that provide an additive approximation for the predictions with partial information, where the predictions are represented by $v_{\rvx\rvy}(\rvs)$ and the additive approximation by $v_{\rvx \rvy}(\vzero) + \rvs^\top \phifast(\rvx, \rvy; \theta)$.



The loss in \cref{eq:objective} was introduced by \cite{jethani2021fastshap} and is derived from an optimization-based characterization of the Shapley value \citep{charnes1988extremal}.
To rigorously justify this
training approach, we derive
new results that show how this objective
controls
the Shapley value estimation error.
Proofs
are
in \Cref{app:proofs}.
First, we show that the explainer's
loss for a single input is strongly convex in the prediction, a result that implies the existence of unique optimal predictions.

\begin{lemma} \label{lemma1}
For a single input-output pair $(x, y)$,
the expected loss under \cref{eq:objective} for the prediction
$\phifast(x, y; \theta)$
is $\mu$-strongly convex
with $\mu = H_{d - 1}\inv$, where $H_{d - 1}$ is the $(d - 1)$th harmonic number.
\end{lemma}

Next, we utilize the strong convexity property from \Cref{lemma1} to prove our main result: that the explainer model's loss function upper bounds the distance between the exact and approximated
Shapley values.
This is notable because we do not utilize ground truth values during training.

\begin{thm} \label{theorem1}
For a model $\phifast(\rvx, \rvy; \theta)$ whose predictions satisfy the constraint in \cref{eq:objective}, the objective value $\mathcal{L}(\theta)$ upper bounds the Shapley value estimation error as follows,
\begin{equation*}
    \E_{p(\rvx, \rvy)} \Big[ \big|\big|\phifast(\rvx, \rvy; \theta) - \phi(v_{\rvx \rvy})\big|\big|_2 \Big]
    \leq \sqrt{2 H_{d - 1} \Big( \mathcal{L}(\theta) - \mathcal{L}^* \Big)},
\end{equation*}
where $\mathcal{L}^*$ represents the loss achieved by the exact Shapley values.
\end{thm}


This shows that our objective is a viable approach for training without exact Shapley values, because optimizing \cref{eq:objective} minimizes an upper bound on the estimation error.
In other words, if we can iteratively optimize the explainer model so that its loss approaches the optimum obtained by the exact Shapley values ($\mathcal{L}(\theta) \to \mathcal{L}^*$), our estimation error will go to zero.

\input{figures/qualitative}

In practice, we train the explainer model $\phifast(\rvx, \rvy; \theta)$ using stochastic gradient descent, and several other steps are important during training. First, we normalize the explainer's unconstrained predictions in order to satisfy the objective's constraint in \cref{eq:objective}; this ensures that the Shapley value's \textit{efficiency} property holds \citep{shapley1953value}. Next, rather than training the explainer from scratch, we fine-tune an existing model that can be either the original classifier or a ViT pre-trained on a different supervised or self-supervised learning task \citep{touvron2021training, he2021masked}; ViTs are more difficult to train than convolutional networks, and we find that fine-tuning is important to train the explainer effectively (\Cref{tab:ablations}).
Finally, we simplify the architecture by estimating Shapley values for all classes simultaneously. Our training approach is described in more detail in \Cref{app:method}.

By using a ViT to estimate Shapley values, we model the true explanation function and learn rich representations that capture not only which class is represented, but where key information is located.
And by fine-tuning an existing model, we allow the explainer to re-use visual features that were informative for other challenging tasks.
Ultimately, the explainer cannot guarantee exact Shapley values, but no approximation algorithm can; instead, it offers a favorable trade-off between accuracy and efficiency, and we find empirically that this approach offers a powerful alternative to the methods currently used for ViTs.

%% file: figures/qualitative.tex
\begin{figure}[t]
    \vskip -0.6cm
    \centering
    \includegraphics[width=0.95\linewidth]{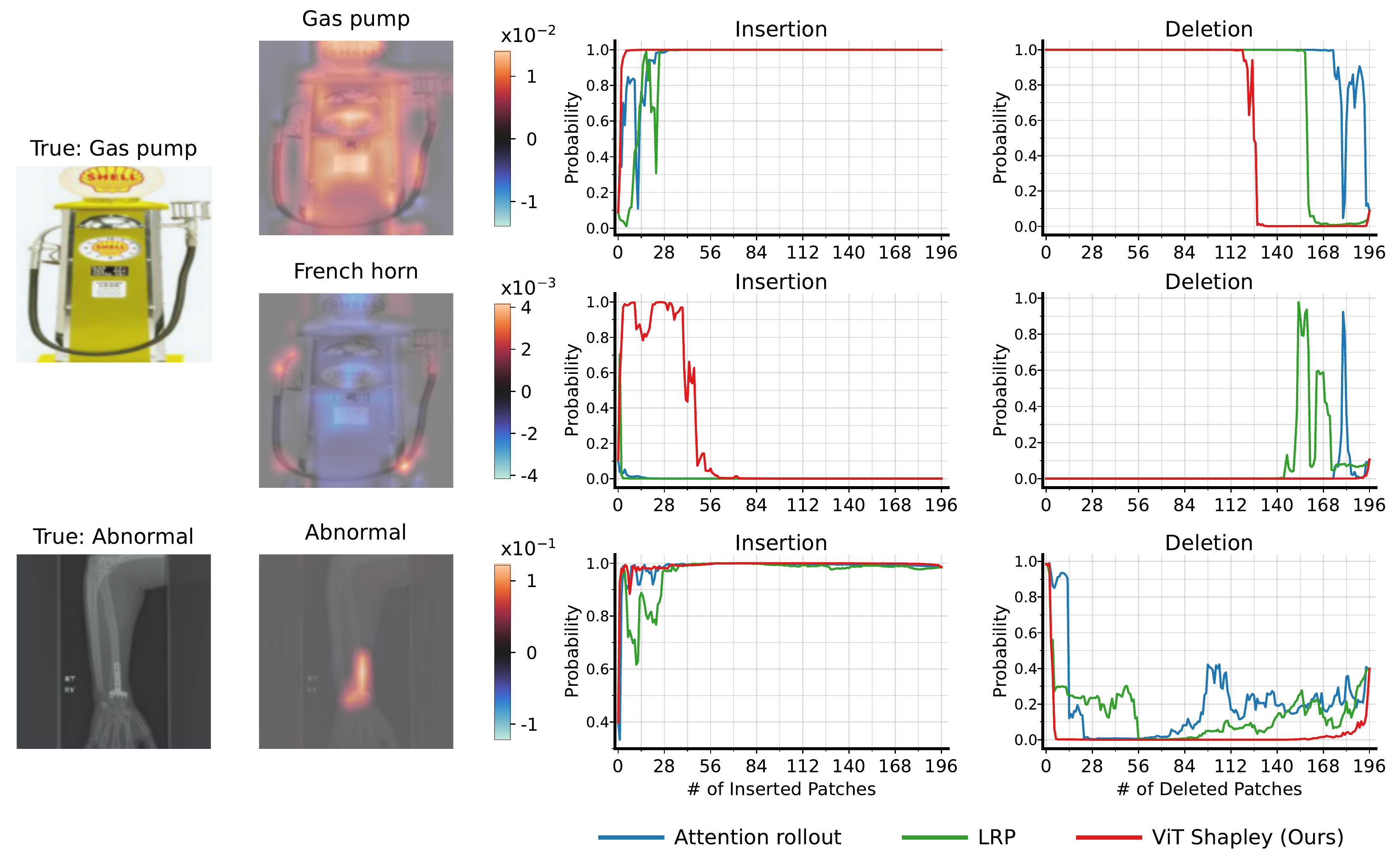}
    \vskip -0.2cm
    \caption{
    Explanations where our approach identifies relevant information for target and non-target classes.
    \textbf{Left:} original images from the ImageNette and MURA datasets.
    \textbf{Middle left:} explanations generated by ViT Shapley
    for specific classes.
    \textbf{Right:} probability of the class being explained after the insertion or deletion of important patches (higher is better for insertion, lower for deletion).
    }
    \label{fig:qualitative}
    \vspace{-.2cm}
\end{figure}

%% file: sections/experiments.tex
We now demonstrate the effectiveness of our approach, termed \textit{ViT Shapley}.\footnote{\url{https://github.com/suinleelab/vit-shapley}}
First, we evaluate attention masking for handling held-out patches in ViTs (\Cref{sec:removal_metrics}). Next, we compare explanations from ViT Shapley to several existing methods (\Cref{sec:explanation_metrics}). Our baselines include attention-, gradient- and removal-based explanations, and we compare these methods via several metrics for explanation quality, including insertion/deletion of important features \citep{petsiuk2018rise},
sensitivity-n \citep{ancona2018towards}, faithfulness \citep{bhatt2021evaluating} and ROAR \citep{hooker2019benchmark}.

Our experiments are based on three image datasets: ImageNette, a natural image dataset consisting of ten ImageNet classes \citep{howard2020fastai, deng2009imagenet}, MURA, a medical image dataset of musculoskeletal radiographs classified as normal or abnormal \citep{rajpurkar2017mura}, and the Oxford IIIT-Pets dataset, which has 37 classes \citep{parkhi2012cats}. See \Cref{fig:qualitative} for example images. 
The main text shows results for ImageNette and MURA,
and Pets results are in \Cref{app:results}.
We use ViT-Base models \citep{rw2019timm} as classifiers for all datasets, unless otherwise specified.

\subsection{Evaluating image patch removal} \label{sec:removal_metrics}

Our initial experiments test whether attention masking is effective for handling held-out image patches. We fine-tuned the classifiers for each dataset
following the procedure described in \Cref{sec:removal}, and we also tested several approaches without performing any fine-tuning: attention masking, attention masking applied after the softmax operation (how dropout is often implemented for ViTs, \citealt{rw2019timm}), setting input patches to zero \citep{naseer2021intriguing}, setting token embeddings to zero, and replacing with random patches from the dataset. Finally, we performed identical fine-tuning while replacing input patches with zeros, which is equivalent to introducing a fixed mask token.

As a measure of how well missing patches are handled, we calculated the KL divergence relative to the full-image predictions as random patches are removed. This can be interpreted as a divergence measure between the masked predictions and the predictions with patches marginalized out (see \Cref{app:masked_training}), or how close we are to correctly removing patch information. The metric is calculated
with randomly generated patch subsets, and it represents whether the model makes reasonable predictions given partial inputs. Similar results for top-1 accuracy are in \Cref{app:results}.

\input{figures/removal}

\Cref{fig:removal} shows the results. Most methods perform well with $<$25\% of patches missing, leading to only small increases in KL divergence. This is especially true for ImageNette, where large objects make the model more robust to missing patches. However, the methods with no fine-tuning begin to diverge as larger numbers of patches are removed and the partial inputs become increasingly off-manifold.
Thus, fine-tuning becomes necessary to properly account for partial inputs
as more patches are removed.

For all datasets, we find that fine-tuning with either attention masking or input patches set to zero provide comparable performance, and that these perform best across all numbers of patches. This means that fine-tuning makes attention masking significantly more effective for marginalizing out missing patches, and these results suggest that training ViTs with held-out tokens may be necessary to enable robustness to partial information. As prior work suggests, properly handling held-out information is crucial for generating informative explanations \citep{frye2020shapley, covert2021explaining}, so the remainder of our experiments proceed with the fine-tuned attention masking approach.

\subsection{Evaluating explanation accuracy} \label{sec:explanation_metrics}

Next, we implemented ViT Shapley by training explainer models for both datasets. We used the fine-tuned classifiers from \Cref{sec:removal_metrics} to handle partial information, and we used the ViT-Base architecture with extra output layers to generate Shapley values for all patches. The explainer models were trained by optimizing \cref{eq:objective} using stochastic gradient descent (see details in \Cref{app:method}), and once trained, the explainer outputs approximate Shapley values in a single forward pass (\Cref{fig:qualitative}).

As comparisons for ViT Shapley, we considered
a large number of baselines.
For attention-based methods, we use attention rollout and the last layer's attention directed to the class token \citep{abnar2020quantifying}. Similar to prior work \citep{chefer2021transformer}, we did not use attention flow due to the computational cost. Next, for gradient-based methods, we use Vanilla Gradients \citep{simonyan2013deep},
IntGrad \citep{sundararajan2017axiomatic}, SmoothGrad \citep{smilkov2017smoothgrad}, VarGrad \citep{hooker2019benchmark}, LRP \citep{chefer2021transformer} and GradCAM \citep{selvaraju2017grad}.
For removal-based methods, we use the leave-one-out approach \citep{zeiler2014visualizing} and RISE \citep{petsiuk2018rise}.
\Cref{app:baselines} describes the baselines in more detail, including how several
were modified to provide patch-level results, and \Cref{app:results} shows the running time for each method.

Given our set of baselines, we used several metrics to evaluate ViT Shapley. Evaluating explanation accuracy is difficult when the true importance is not known a priori, so we rely on metrics that test how removing (un)important features affects a model's predictions. Intuitively, removing influential features for a particular class should reduce the class probability, and removing non-influential features should not affect or even increase the class probability. Removal-based explanations are implicitly related to such metrics \citep{covert2021explaining}, but attention- and gradient-based methods may be hoped to provide strong performance with lower computational cost.

\input{figures/metrics}

First, we implemented the widely used \textit{insertion} and \textit{deletion} metrics \citep{petsiuk2018rise}. For these, we generate predictions while inserting/removing features in order of most to least important, and we then evaluate the area under the curve of prediction probabilities (see \Cref{fig:qualitative}). Here, we average the results across 1,000 images for their true class. We use random test set images for ImageNette, and for MURA we use test examples that were classified as abnormal because these are more important in practice. When removing information, we use the fine-tuned classifier because this represents the closest approximation to properly
removing information from the model (\Cref{sec:removal}).

\Cref{tab:metrics} displays the results, and we find that ViT Shapley offers the best performance on both datasets. RISE and LRP tend to be the most competitive baselines, and perhaps surprisingly, certain other methods fail to outperform a random baseline (GradCAM, SmoothGrad, VarGrad). The baselines are sometimes competitive with ViT Shapley on insertion, but the gap for the deletion metric is larger. Practically, this means that ViT Shapley identifies important features that quickly drive the prediction towards a given class, and that quickly reduce the prediction probability when deleted.

\input{figures/metrics_nontarget}

Next, we modified these metrics to address a common issue with model explanations: that their results are not specific to each class \citep{rudin2019stop}. ViT Shapley produces separate explanations for each class, so it can identify relevant patches even for non-target classes (see \Cref{fig:qualitative}). \Cref{tab:metrics_nontarget} shows insertion/deletion results averaged across all non-target classes for ImageNette. The attention-based methods do not produce class-specific explanations, and the remaining baselines generally provide poor results. Empirically, this is because the explanations are often similar across classes (see \Cref{app:qualitative}). ViT Shapley performs best, particularly on insertion, and RISE is the best-performing baseline. \Cref{app:results} shows results for MURA, as well as the curves used to calculate these results.

The insertion and deletion metrics
only test importance rankings, so we require other
metrics
to test the specific attribution values. \textit{Sensitivity-n} \citep{ancona2018towards} was proposed for this purpose, and it measures
whether attributions correlate with the impact on a model's prediction when a feature is removed. The correlation is
typically
calculated
across
subsets of a fixed size, and then averaged across many predictions.
\textit{Faithfulness} \citep{bhatt2021evaluating} is a similar
metric where the correlation is calculated
across subsets of all sizes.

\Cref{tab:metrics} and \Cref{tab:metrics_nontarget} show faithfulness results.
Among the baselines, RISE and LRP remain most competitive, but ViT Shapley again performs best for both datasets.
\Cref{fig:sensitivity} shows sensitivity-n results calculated across a range of subset sizes.
Leave-one-out naturally performs best for large subset sizes, but ViT Shapley
performs the best overall, particularly with smaller subsets.
The sensitivity-n results focus on the target class,
but \Cref{app:results} shows
results for non-target classes where ViT Shapley's advantage over many baselines (including LRP) is even larger.

\input{figures/sensitivity}

Finally, we performed an evaluation inspired by ROAR \citep{hooker2019benchmark}, which tests how a model's accuracy degrades as important features are removed. ROAR suggests retraining
with masked inputs, but this is
unnecessary here
because the fine-tuned classifier is designed to handle
held-out patches. We therefore generated multiple versions of the metric. First, we evaluated accuracy while using the fine-tuned classifier to handle masked patches. Second, we
repeated the evaluation
using a separate evaluator model trained directly with held-out patches,
similar to EVAL-X \citep{jethani2021have}. Third, we performed masked retraining as described by ROAR. The first version represents the original classifier's best-effort prediction, and the second is a best-effort prediction disconnected from the original model; masked retraining is similar, but the retrained model can exploit information communicated by the masking, such as the shape and position of the removed object.

\input{figures/roar_imagenette}

\Cref{fig:roar} shows the results when removing important patches.
ViT Shapley consistently outperforms the baselines across the first two versions of the metric, yielding
faster degradation when important patches are removed. ViT Shapley also performs best when inserting important patches, yielding a faster increase in accuracy
(\Cref{app:results}).
It is outperformed by several baselines
with masked retraining
in the deletion direction (\Cref{fig:roar} bottom left), but we find that this is likely
due to spatial information leaked by
ViT Shapley's deleted patches;
indeed, when we retrained
\textit{without positional embeddings},
we found that ViT Shapley achieved the fastest degradation with a small number of deleted patches (\Cref{fig:roar} bottom right).
Interestingly, positional embeddings in ViTs
offer a unique approach to alleviate
ROAR's known
information leakage issue
\citep{jethani2021have}.


In addition to these experiments, we include many further results in the supplement (\Cref{app:results}). First, we observe similar benefits for ViT Shapley when using the Oxford-IIIT Pets dataset. Next, regarding the choice of architecture, we observe consistent results when replacing ViT-Base with ViT-Tiny, -Small or -Large. We also replicate our results when using a classifier trained directly with random masking, an approach discussed in prior work to accommodate partial input information \citep{covert2021explaining}. We then generated metrics comparing ViT Shapley's approximation quality with KernelSHAP \citep{lundberg2017unified}, and we found that ViT Shapley's accuracy is equivalent to running KernelSHAP for roughly 120,000 model evaluations (\Cref{app:results}). Lastly, we provide qualitative examples in \Cref{app:qualitative}, including comparisons with the baselines.
Overall, these results show that ViT Shapley is a practical and effective approach for explaining ViT predictions.


%% file: figures/removal.tex
\begin{figure}[t]
    \figurespacing
    \centering
    \includegraphics[width=0.9\linewidth]{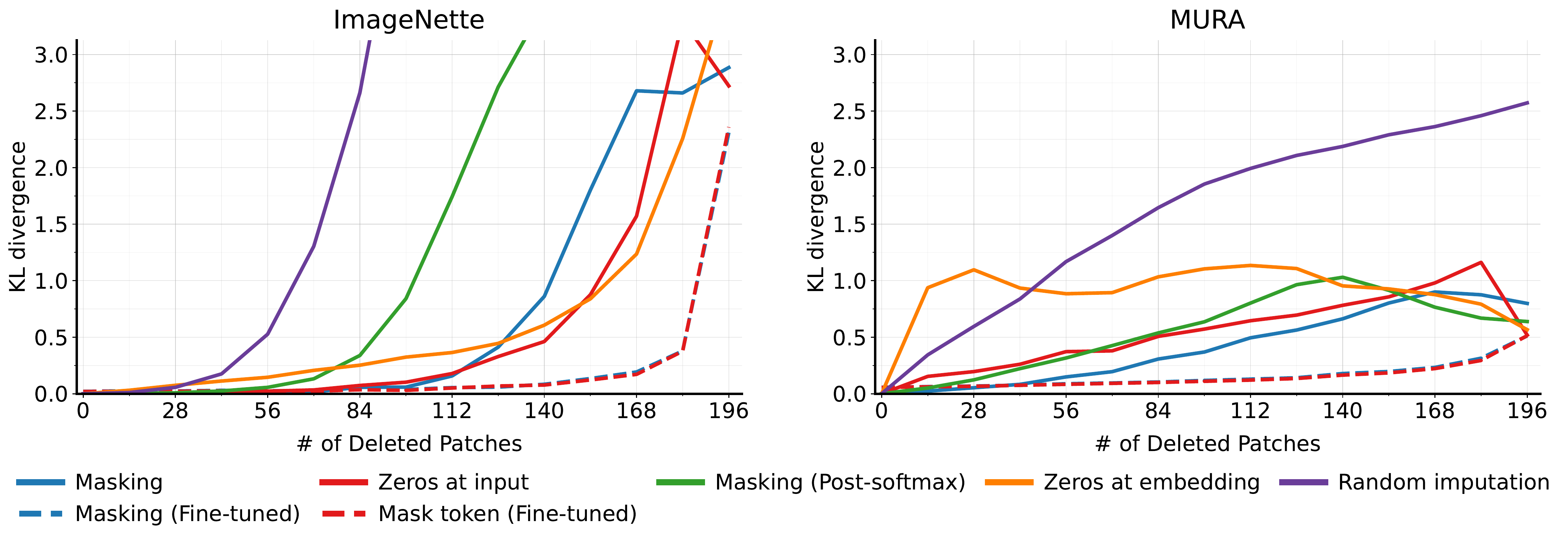}
    \caption{
    ViT predictions given
    partial information.
    We delete patches at random using several removal mechanisms, and then measure the quality of the resulting predictions via the
    KL divergence relative to
    the original, full-image
    predictions (lower is better).
    }
    \label{fig:removal}
    \vspace{-.2cm}
\end{figure}

%% file: figures/metrics.tex
\begin{table}[t]
\centering
\vskip - 0.4cm
\caption{Evaluating ViT Shapley using standard explanation metrics, with explanations calculated for the target class only. Methods that fail to outperform the random baseline are shown in gray, and the best results are shown in bold (accounting for 95\% confidence intervals).
} \label{tab:metrics}
\begin{center}
{\tablesize
    \begin{tabular}{lcccccc}
    \toprule
     & \multicolumn{3}{c}{ImageNette} & \multicolumn{3}{c}{MURA} \\
    \cmidrule(lr){2-4} \cmidrule(lr){5-7}
     & Ins. ($\uparrow$) & Del. ($\downarrow$) & Faith. ($\uparrow$) & Ins. ($\uparrow$) & Del. ($\downarrow$) & Faith. ($\uparrow$) \\
    \midrule
    Attention last & 0.962 (0.004) & 0.793 (0.013) & \textbf{0.694 (0.015)} & 0.890 (0.010) & 0.592 (0.013) & 0.635 (0.016) \\
    Attention rollout & \textcolor{gray}{0.938 (0.005)} & 0.880 (0.010) & \textbf{0.704 (0.015)} & \textcolor{gray}{0.845 (0.011)} & 0.692 (0.014) & 0.618 (0.016) \\\midrule
    GradCAM & \textcolor{gray}{0.914 (0.006)} & 0.937 (0.008) & 0.680 (0.015) & 0.899 (0.009) & 0.681 (0.015) & 0.631 (0.016) \\
    IntGrad & 0.967 (0.004) & 0.930 (0.008) & 0.403 (0.024) & 0.897 (0.010) & 0.796 (0.015) & 0.201 (0.022) \\
    Vanilla & \textcolor{gray}{0.950 (0.004)} & 0.808 (0.013) & \textbf{0.703 (0.015)} & 0.890 (0.010) & 0.537 (0.014) & 0.629 (0.016) \\
    SmoothGrad & \textcolor{gray}{0.947 (0.005)} & \textcolor{gray}{0.942 (0.006)} & \textbf{0.703 (0.015)} & 0.870 (0.010) & 0.813 (0.011) & 0.617 (0.016) \\
    VarGrad & \textcolor{gray}{0.949 (0.005)} & \textcolor{gray}{0.946 (0.005)} & \textbf{0.700 (0.015)} & \textcolor{gray}{0.857 (0.011)} & 0.823 (0.011) & 0.615 (0.016) \\
    LRP & 0.967 (0.004) & 0.779 (0.014) & \textbf{0.705 (0.015)} & 0.900 (0.009) & 0.551 (0.013) & 0.646 (0.016) \\\midrule
    Leave-one-out & 0.969 (0.002) & 0.917 (0.010) & 0.140 (0.040) & 0.926 (0.008) & 0.694 (0.017) & 0.308 (0.032) \\
    RISE & 0.977 (0.001) & 0.860 (0.014) & \textbf{0.704 (0.015)} & 0.957 (0.004) & 0.573 (0.018) & 0.618 (0.016) \\
    \textbf{ViT Shapley} & \textbf{0.985 (0.002)} & \textbf{0.691 (0.014)} & \textbf{0.711 (0.015)} & \textbf{0.971 (0.002)} & \textbf{0.307 (0.013)} & \textbf{0.707 (0.013)} \\\midrule
    Random & \textcolor{gray}{0.951 (0.005)} & \textcolor{gray}{0.951 (0.005)} & - & \textcolor{gray}{0.849 (0.010)} & \textcolor{gray}{0.847 (0.010)} & - \\
    \bottomrule
    \end{tabular}
}
\end{center}
\end{table}

%% file: figures/metrics_nontarget.tex
\begin{wraptable}[22]{r}{0.59\textwidth}
    \vskip -0.15in
    \caption{Evaluating ViT Shapley for explaining non-target classes.
    Methods that fail to outperform the random baseline
    are shown in gray, and the best results are
    shown in bold (accounting for 95\% confidence intervals).
    } \label{tab:metrics_nontarget}
    \vskip 0.1in
    \begin{center}
    {\tablesize
        \begin{tabular}{lccc}
        \toprule
         & \multicolumn{3}{c}{ImageNette} \\
        \cmidrule(lr){2-4}
         & Ins. ($\uparrow$) & Del. ($\downarrow$) & Faith. ($\uparrow$) \\
        \midrule
        Attention last & - & - & - \\
        Attention rollout & - & - & - \\\midrule
        GradCAM & 0.021 (0.002) & \textcolor{gray}{0.005 (0.000)} & -0.672 (0.015) \\
        IntGrad & 0.008 (0.001) & 0.004 (0.000) & 0.294 (0.022) \\
        Vanilla & \textcolor{gray}{0.006 (0.001)} & \textcolor{gray}{0.020 (0.001)} & -0.682 (0.015) \\
        SmoothGrad & \textcolor{gray}{0.006 (0.001)} & \textcolor{gray}{0.006 (0.001)} & -0.683 (0.015) \\
        VarGrad & \textcolor{gray}{0.006 (0.001)} & \textcolor{gray}{0.006 (0.001)} & -0.680 (0.015) \\
        LRP & \textcolor{gray}{0.004 (0.001)} & \textcolor{gray}{0.022 (0.001)} & -0.680 (0.015) \\\midrule
        Leave-one-out & 0.013 (0.002) & 0.003 (0.000) & -0.017 (0.028) \\
        RISE & 0.023 (0.003) & 0.002 (0.000) & -0.681 (0.015) \\
        \textbf{ViT Shapley} & \textbf{0.093 (0.004)} & \textbf{0.001 (0.000)} & \textbf{0.672 (0.014)} \\\midrule
        Random & \textcolor{gray}{0.005 (0.001)} & \textcolor{gray}{0.005 (0.001)} & - \\
        \bottomrule
        \end{tabular}
    }
    \end{center}
\end{wraptable}

%% file: figures/sensitivity.tex
\begin{figure}[t]
    \figurespacing
    \centering
    \includegraphics[width=0.95\linewidth]{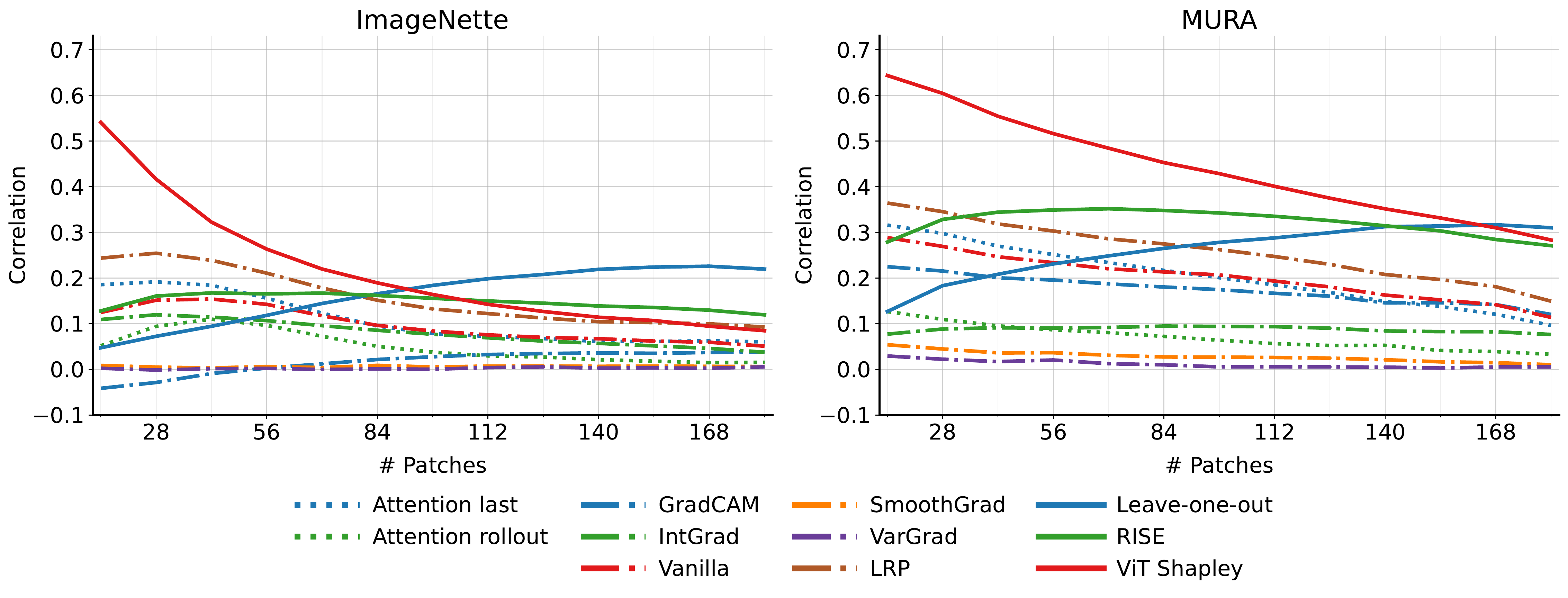}
    \vskip -0.2cm
    \caption{
    Sensitivity-n evaluation
    for different subset sizes. The metric is generated separately for a range of subset sizes,
    whereas faithfulness is calculated jointly over subsets of all sizes.
    }
    \label{fig:sensitivity}
    \vskip -0.3cm
\end{figure}

%% file: figures/roar_imagenette.tex
\begin{figure}[t]
    \vskip -0.1in
    \figurespacing
    \centering
    \includegraphics[width=0.95\linewidth]{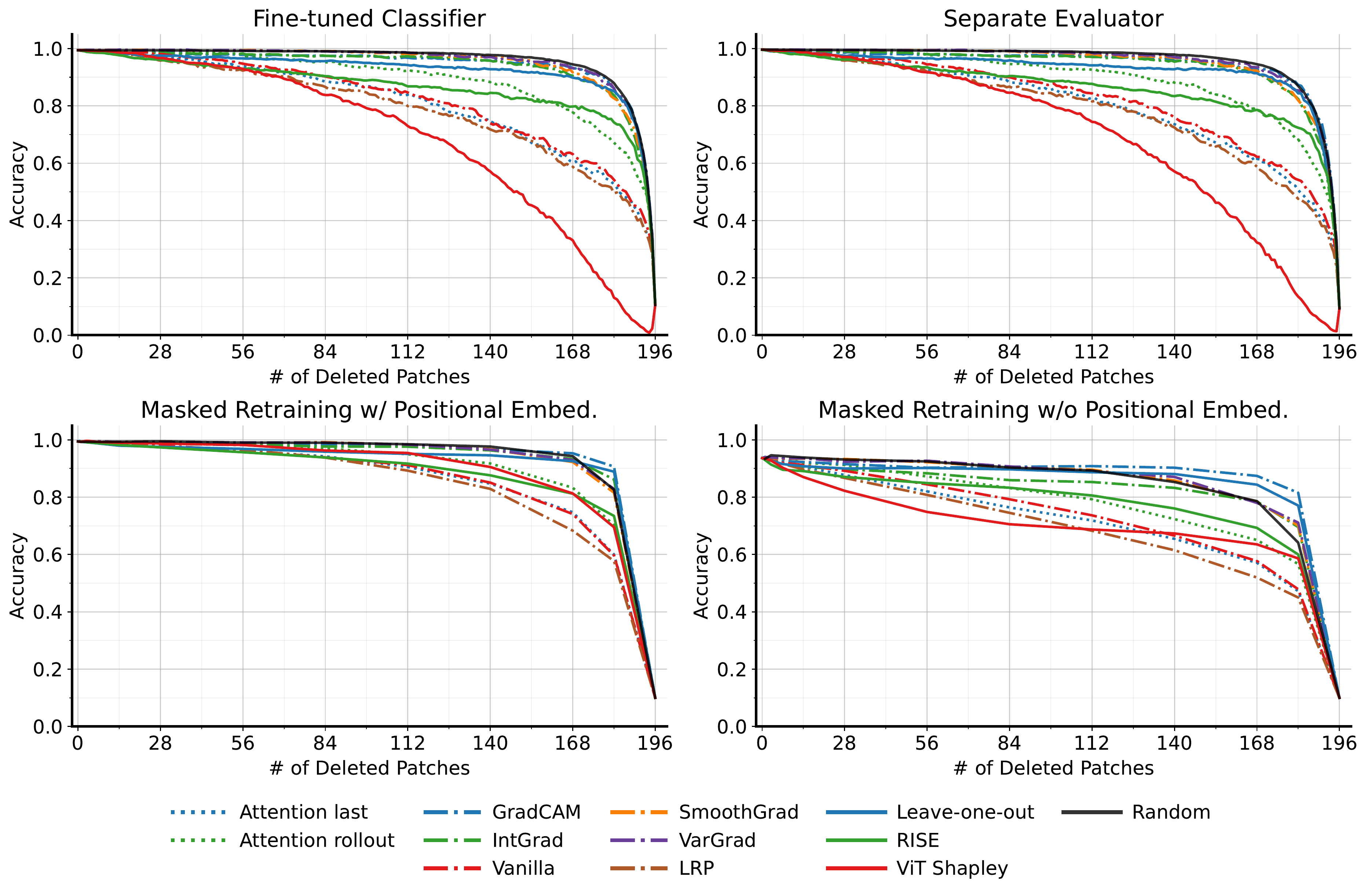}
    \caption{ImageNette accuracy when removing features in order of
    their importance, run with four evaluation strategies: fine-tuned classifier, separate evaluator, masked retraining, and masked retraining without positional embeddings. 
    }
    \label{fig:roar}
    \vskip -0.2cm
\end{figure}

%% file: sections/conclusion.tex
In this work, we developed a learning-based approach to generate Shapley values for ViTs.
Our approach involves training a separate explainer model
using an objective that does not require ground truth supervision,
and ViT Shapley outperforms a variety of attention- and gradient-based methods across a range of
accuracy
metrics.
Our experiments provide empirical support
for Shapley values as an alternative to attention-based approaches, which remain popular for transformer models.
Future directions involve extending ViT Shapley to NLP models,
operating with arbitrary token groups or superpixels, and accelerating or otherwise improving the explainer model's training.

%% file: sections/appendix.tex
\section{Attention masking} \label{app:masking}








This section describes our attention masking approach in
detail. First, recall that ViTs use query-key-value self-attention \citep{vaswani2017attention, dosovitskiy2020image}, which accepts a set of input tokens and produces a weighted sum of learned token values. Given an input $\rvz \in \R^{d \times \hin}$ and parameters $\mathrm{U}_{qkv} \in \R^{\hin \times 3\hout}$, we compute the self attention output $\mathrm{SA}(\rvz)$ for a single head as follows:

\begin{align}
    [\rvQ, \rvK, \rvV] &= \rvz \mathrm{U}_{qkv} \\
    \rvA &= \softmax (\rvQ \rvK^\top / \sqrt{\hout} ) \label{eq:attention} \\
    \mathrm{SA}(\rvz) &= \rvA \rvV.
\end{align}

In multihead self-attention, we perform this operation in parallel over $k$ attention heads and project the concatenated outputs. Denoting each head's output as $\mathrm{SA}_i(\rvz)$ and the projection matrix as $\mathrm{U}_{msa} \in \R^{k \cdot \hout \times \hin}$, the multihead self-attention output $\mathrm{MSA}(\rvz)$ is

\begin{equation}
    \mathrm{MSA}(\rvz) = [\mathrm{SA}_1(\rvz), \ldots, \mathrm{SA}_k(\rvz)]\mathrm{U}_{msa}.
\end{equation}

Multihead self-attention can operate with any number of tokens, so given a subset $\rvs \in \{0, 1\}^d$ and an input $\rvx$, we can evaluate a ViT using only tokens for the patches $\rvx_\rvs = \{\rvx_i : \rvs_i = 1\}$.
However, for implementation purposes it is preferable to maintain the same number of tokens within
a minibatch. We therefore provide all tokens to the model and
achieve the same effect using attention masking. Our exact approach is described below.

Let $\rvz \in \R^{d \times \hin}$ represent the full token set for an input $\rvx$ and let $\rvs$ be a subset. At each self-attention layer, we construct a mask matrix $\rvS = [\rvs, \ldots, \rvs]^\top \in \{0, 1\}^{d \times d}$ and calculate the masked self-attention output $\mathrm{SA}(\rvz, \rvs)$ as follows:

\begin{align}
    \rvA &= \softmax ( (\rvQ \rvK^\top - (1 - \rvS) \cdot \infty) / \sqrt{\hout} ) \label{eq:masked_attention} \\
    \mathrm{SA}(\rvz, \rvs) &= \rvA \rvV.
\end{align}

The masked multihead self-attention output is then calculated similarly to the original version:

\begin{equation}
    \mathrm{MSA}(\rvz, \rvs) = [\mathrm{SA}_1(\rvz, \rvs), \ldots, \mathrm{SA}_k(\rvz, \rvs)]\mathrm{U}_{msa}.
\end{equation}

Due to the masking in \cref{eq:masked_attention}, each output token in $\mathrm{MSA}(\rvz, \rvs)$ is guaranteed not to attend to tokens from $\rvx_{\vone - \rvs} = \{\rvx_i : \rvs_i = 0\}$.
We use masked self-attention in all layers of the network, so that the tokens for $\rvx_\rvs$ remain invariant to those for $\rvx_{\vone - \rvs}$ throughout the entire model, including after the layer norm and fully-connected layers.
When the final prediction is calculated using the class token,
the
output is
equivalent to
using only the tokens for
$\rvx_\rvs$.
If the final prediction is instead produced using global average pooling \citep{beyer2022better}, we can modify the average to account only for tokens we wish to include.

\section{Masked training} \label{app:masked_training}

In this section, we provide proofs
to justify training a ViT classifier with held-out tokens, either as part of the original
training
or as part of a post-hoc fine-tuning procedure (the surrogate model training described in \Cref{sec:removal}).
Our proofs
are similar to those in prior work that discusses marginalizing
out features
using their conditional distribution \citep{covert2021explaining}.

First, consider a model trained directly with masking. Given a subset distribution $p(\rvs)$ and the data distribution $p(\rvx, \rvy)$, we can train a model $f(\rvx_\rvs ; \eta)$ with cross-entropy loss and random masking by minimizing the following:

\begin{equation}
    \min_\eta \quad \E_{p(\rvx, \rvy)} \E_{p(\rvs)} [- \log f_\rvy(\rvx_\rvs; \eta)]. \label{eq:masked_training}
\end{equation}

To understand the global optimizer for this loss function, consider the expected loss for the prediction given a fixed model input $x_s$:

\begin{equation}
    \E_{p(\rvy, \rvx_{\vone - s} \mid x_s)} [- \log f_\rvy(x_s; \eta)] = \E_{p(\rvy \mid x_s)} [- \log f_\rvy(x_s; \eta)]. \label{eq:expected_loss}
\end{equation}

The expression in \cref{eq:expected_loss} is equal to the KL divergence $\KL(p(\rvy \mid x_s) \mid\mid f(x_s; \eta))$ up to a constant value, so
the prediction
that minimizes this loss is $p(\rvy \mid x_s)$.
For any subset $s \in \{0, 1\}^d$ where $p(s) > 0$,
we then
have the following result for the
model $f(\rvx_\rvs ; \eta^*)$ that minimizes \cref{eq:masked_training}:

\begin{equation*}
    f_y(x_s; \eta^*) = p(y \mid x_s) \;\; \text{a.e. in } p(\rvx).
\end{equation*}

Intuitively, this means that training the original model with masking
estimates $f(\rvx_s; \eta) \approx p(\rvy \mid \rvx_s)$. In practice, we use a subset distribution $p(\rvs)$ where
$p(s) > 0$ for all $s \in \{0, 1\}^d$: we set $p(\rvs)$ by sampling the cardinality uniformly at random and then sampling the members, which is equivalent to defining $p(\rvs)$ as

\begin{equation*}
    p(s) = \frac{1}{\binom{d}{\vone^\top s} \cdot (d + 1)}.
\end{equation*}

Alternatively, we can use a
model $f(\rvx; \eta)$ trained without masking and fine-tune it to better
handle held-out features.
In our case, this yields a \textit{surrogate model} \citep{frye2020shapley}
denoted as $g(\rvx_\rvs; \beta)$ that we fine-tune
by minimizing the following loss:

\begin{equation}
    \min_\beta \quad \E_{p(\rvx)} \E_{p(\rvs)} \Big[\KL\big(f(\rvx; \eta) \mid\mid g(\rvx_\rvs; \beta) \big)\Big]. \label{eq:surrogate_app}
\end{equation}

To understand the global optimizer for the above
loss, we can again consider the expected loss given a fixed
input $x_s$:

\begin{equation}
    \E_{p(\rvx_{\vone - s} \mid x_s)} \Big[ \KL\big( f(\rvx; \eta) \mid\mid g(x_s; \beta) \big) \Big] = \KL\big(\E[f(\rvx; \eta) \mid x_s] \mid\mid g(x_s; \beta) \big) + \mathrm{const}.
\end{equation}

The
distribution that minimizes this loss
is the expected
output given the available features, or $\E[f(\rvx; \eta) \mid x_s]$. By the same argument presented above, we then have the following result for the
optimal surrogate $g(\rvx_\rvs; \beta^*)$ that minimizes \cref{eq:surrogate_app}:

\begin{equation*}
    g(x_s; \beta^*) = \E[f(\rvx; \eta) \mid x_s] \;\; \text{a.e. in } p(\rvx).
\end{equation*}

Notice that if the initial model is optimal, or $f(\rvx; \eta) = p(\rvy \mid \rvx)$, then the optimal surrogate satisfies
$g(\rvx_s ; \beta^*) = p(\rvy \mid \rvx_s)$.

\section{Explainer training approach} \label{app:method}

In this section, we summarize our
approach for training
the explainer model and describe
several design choices. Recall that the explainer is a
vision transformer
$\phifast(\rvx, \rvy; \theta) \in \R^d$ that we train by minimizing the following loss:

\begin{align*}
    \min_\theta \quad &\E_{p(\rvx, \rvy)}
    \E_{\psh(\rvs)} \Big[ \big( v_{\rvx \rvy}(\rvs) - v_{\rvx \rvy}(\vzero) - \rvs^\top \phifast(\rvx, \rvy; \theta) \big)^2 \Big] \label{eq:objective} \\
    &\mathrm{s.t.} \quad \vone^\top \phifast(x, y; \theta) = v_{x y}(\vone) - v_{x y}(\vzero) \quad \forall \; (x, y). \nonumber
\end{align*}

\paragraph{Additive efficient normalization} The constraint on the explainer predictions is necessary to ensure that the global optimizer
outputs the exact Shapley values,
and we use the same approach as prior work to enforce this constraint \citep{jethani2021fastshap}.
We allow
the model to make unconstrained predictions that we then modify
using the following transformation:

\begin{equation}
    \phifast(x, y; \theta) \gets \phifast(x, y; \theta) + \frac{v_{xy}(\vone) - v_{xy}(\vzero) - \vone^\top \phifast(x, y; \theta)}{d}.
\end{equation}

This operation is known as the \textit{additive efficient normalization} \citep{ruiz1998family}, and it can be interpreted as projecting the predictions onto the hyperplane where the constraint holds \citep{jethani2021fastshap}. We implement it as an output activation function, similar to how softmax is used to ensure valid probabilistic predictions for
classification
models.

\paragraph{Subset distribution} The specific distribution $\psh(\rvs)$ in our loss function is motivated by the Shapley value's weighted least squares characterization \citep{charnes1988extremal, lundberg2017unified}. This result states that the Shapley values for a game $v: \{0, 1\}^d \mapsto \R$
are the solution to the following optimization problem:

\begin{align*}
    \min_{\phi \in \R^d} \quad & \sum_{0 < \vone^\top s < d} \frac{d - 1}{\binom{d}{\vone^\top s} (\vone^\top s)(d - \vone^\top s)} \Big( v(s) - v(\vzero) - s^\top \phi \Big)^2 \\
    &\mathrm{s.t.} \quad \vone^\top \phi = v(\vone) - v(\vzero).
\end{align*}



We obtain $\psh(\rvs)$ by normalizing the weighting term in the summation, and doing so yields a distribution $\psh(s) \propto (\vone^\top s - 1)!(d - \vone^\top s - 1)!$ for $0 < \vone^\top s < d$ and $\psh(\vone) = \psh(\vzero) = 0$.
To sample from $\psh(\rvs)$,
we calculate the probability mass on each cardinality,
sample a cardinality $\rvm$ from this multinomial distribution, and then select $\rvm$ indices uniformly at random. 

\paragraph{Stochastic gradient descent} As is common in deep learning, we optimize our objective using stochastic gradients rather than exact gradients. To estimate our objective, we require a set of tuples $(\rvx, \rvy, \rvs)$ that we obtain as follows. First, we sample an input $x \sim p(\rvx)$. Next, we sample multiple subsets $s \sim \psh(\rvs)$. To reduce gradient variance, we use the paired sampling trick \citep{covert2021improving} and pair each subset $s$ with its complement $\vone - s$. Then, we use
our explainer to output Shapley values simultaneously for all classes $y \in \{1, \ldots, K\}$. Finally, we minibatch this procedure across multiple inputs $x$ and calculate our loss across the resulting set of tuples $(\rvx, \rvy, \rvs)$.


\paragraph{Fine-tuning} Rather than training the
ViT explainer from scratch, we find that fine-tuning an existing model leads to better performance. This is consistent with recent work that finds ViTs
challenging to train from scratch \citep{dosovitskiy2020image}.
We have several options for initializing the explainer: we can use 1)~the original classifier $f(\rvx; \eta)$, 2)~the fine-tuned classifier
$g(\rvx_s; \beta)$, or 3)~a ViT pre-trained on another task. We treat this choice as a hyperparameter, selecting the initialization that yields
the best performance. We also experiment with freezing certain layers in the model, but we
find that training all the parameters leads to the best performance.

\paragraph{Explainer architecture}
We use standard ViT architectures for the explainer. These typically append a class token to the set of image tokens \citep{dosovitskiy2020image}, and we find it beneficial to preserve this token in pre-trained architectures even though it is unnecessary for
Shapley value estimation.
We require a separate output head from the pre-trained architecture, and our explainer head consists of one additional self-attention block followed by three fully-connected layers.
Each image patch yields one Shapley value estimate per class,
and we discard the results for the class token.

\paragraph{Hyperparameter tuning}
To select hyperparaters related to the learning rate, initialization and architecture,
we use a pre-computed set of tuples $(\rvx, \rvy, \rvs)$ to calculate a validation loss. These are generated
using
inputs $x$ that were not used for training, so our validation loss can be interpreted as an unbiased estimator of the objective function. This approach
serves as an inexpensive alternative to comparing with ground truth Shapley values for a large number of samples.

\paragraph{Training pseudocode} \Cref{alg:training} shows a simplified version of our training algorithm, without minibatching, sampling multiple subsets $s$, or parallelizing across the classes $y$.

\begin{algorithm}[H]
  \SetAlgoLined
  \DontPrintSemicolon
  \KwInput{Coalitional game $v_{\rvx\rvy}(\rvs)$, learning rate $\alpha$}
  \KwOutput{Explainer $\phifast(\rvx, \rvy ; \theta)$}
  initialize $\phifast( \rvx, \rvy ; \theta)$ \;
  \While{not converged}{
    sample $(x, y) \sim p(\rvx, \rvy)$, $s \sim \psh(\rvs)$ \;
    predict $\phi \gets \phifast(x, y; \theta)$ \;
    set $\phi \gets \phi + d^{-1} \left( v_{xy}(\vone) - v_{xy}(\vzero) - \vone^\top \phi \right)$ \;
    calculate $\mathcal{L} \gets \left( v_{xy}(s) - v_{xy}(\vzero) - s^\top \phi \right)^2$ \;
    update $\theta \gets \theta - \alpha \nabla_\theta \mathcal{L}$ \;
  }
\caption{Explainer training}
\label{alg:training}
\end{algorithm}

\subsection{Hyperparameter choices}

When training the original classifier and fine-tuned classifier
models, we used a learning rate of $10^{-5}$
and trained for 25 epochs and 50 epochs, respectively. The MURA classifier was trained with an upweighted loss for negative examples to account for class imbalance. The best model was selected based on the validation criterion, where we used 0-1 accuracy for ImageNette and Oxford-IIIT Pets, and Cohen Kappa for MURA.

When training the explainer model, we used the same ViT-Base architecture as the original classifier and initialized using the fine-tuned classifier,
as this gave the best results. We used the AdamW optimizer \citep{loshchilov2018decoupled} with a cosine
learning rate schedule and a maximum learning rate of $10^{-4}$,
and we trained the model for 100 epochs, selecting the best model based on the validation loss. We used standard data augmentation steps: random resized crops, vertical flips, horizontal flips, and color jittering including brightness, contrast, saturation, and hue. We used minibatches of size 64 with 32 subset samples $s$ per $x$ sample, and we found that using a tanh nonlinearity on the explainer
predictions was helpful to stabilize training.

Finally, we modified the ViT architecture to output Shapley values for each token and each class: we removed the classification head and
added an extra attention layer, followed by three fully-connected layers with width 4 times
the embedding dimension, and we fine-tuned the entire ViT backbone.
These choices were determined by an ablation study with different model configurations, and we also compared with training training the ViT from scratch and training a separate U-Net explainer model \citep{ronneberger2015u} (see \Cref{tab:ablations}).


\input{figures/ablations}

We used a machine with 2 GeForce RTX 2080Ti GPUs
to train the explainer model, and due to GPU memory constraints we loaded the classifier
and explainer to separate GPUs and trained with mixed precision using PyTorch Lightning.\footnote{\url{https://github.com/PyTorchLightning/pytorch-lightning}} Training the explainer model required roughly 19 hours
for the ImageNette dataset and 60 hours
for the MURA dataset.


\clearpage

\section{Proofs} \label{app:proofs}

Here, we re-state and prove our main results from \Cref{sec:estimation}.

\textbf{Lemma 1.} \textit{For a single input-output pair $(x, y)$,
the expected loss under \cref{eq:objective} for the prediction $\phifast(x, y; \theta)$
is $\mu$-strongly convex
with $\mu = H_{d - 1}\inv$, where $H_{d - 1}$ is the $(d - 1)$th harmonic number.
}

\begin{proof}

For an input-output pair $(x, y)$, the expected loss for the prediction $\phi = \phifast(x, y; \theta)$
under our objective
is given by

\begin{align*}
    h_{xy}(\phi) &= \phi^\top \E_{\psh(\rvs)}[\rvs \rvs^\top] \phi - 2 \phi^\top \E_{\psh(\rvs)}\Big[\rvs \big(v_{xy}(\rvs) - v_{xy}(\vzero) \big) \Big] + \E_{\psh(\rvs)}\Big[ \big(v_{xy}(\rvs) - v_{xy}(\vzero)\big)^2 \Big].
\end{align*}

This is a quadratic function of $\phi$ with its Hessian given by

\begin{align*}
    \nabla_\phi^2 h_{xy}(\phi) = 2 \cdot \E_{\psh(\rvs)}[\rvs \rvs^\top].
\end{align*}

The convexity of $h_{xy}(\phi)$ is determined by the Hessian's eigenvalues,
and its
entries can be derived from the subset distribution $\psh(\rvs)$; see similar results in \cite{simon2020projected} and \cite{covert2021improving}.
The distribution assigns equal probability to subsets of equal cardinality, so we define the shorthand notation $p_k \equiv \psh(s)$ for $s$ such that $\vone^\top s = k$. Specifically, we have

\begin{align*}
    p_k &= Q\inv \frac{d - 1}{\binom{d}{k}k(d - k)} \quad\quad \mathrm{and} \quad\quad
    Q = \sum_{k = 1}^{d - 1} \frac{d - 1}{k(d - k)}.
\end{align*}

We can then write $A \equiv \E_{\psh(\rvs)}[\rvs \rvs^\top]$ and derive its entries as follows:

\begin{align*}
    A_{ii} &= \mathrm{Pr}(\rvs_i = 1) = \sum_{k = 1}^d \binom{d - 1}{k - 1} p_k \\
    &= Q\inv \sum_{k = 1}^{d - 1} \frac{(d - 1)}{d(d - k)} \\
    &= \frac{\sum_{k = 1}^{d - 1} \frac{d - 1}{d(d - k)}}{\sum_{k = 1}^{d - 1} \frac{d - 1}{k(d - k)}} \\
    %
    A_{ij} &= \mathrm{Pr}(\rvs_i = \rvs_j = 1) = \sum_{k = 2}^d \binom{d - 2}{k - 2} p_k \\
    &= Q\inv \sum_{k = 2}^{d - 1} \frac{k - 1}{d(d - k)} \\
    &= \frac{\sum_{k = 2}^{d - 1} \frac{k - 1}{d(d - k)}}{\sum_{k = 1}^{d - 1} \frac{d - 1}{k(d - k)}}
\end{align*}

Based on this, we can see that $A$ has the structure $A = (b - c) I_d + c \vone \vone^\top$
for $b \equiv A_{ii} - A_{ij}$ and $c \equiv A_{ij}$. Following the argument by \cite{simon2020projected},
the minimum eigenvalue is then given by $\lambda_{\min}(A) = b - c$.
Deriving the specific value shows that it depends on the $(d - 1)$th harmonic number, $H_{d - 1}$:

\begin{align*}
    \lambda_{\mathrm{min}}(A) &= b - c = A_{ii} - A_{ij} \\
    &= \frac{\frac{1}{d} + \sum_{k = 2}^{d - 1} \Big( \frac{d - 1}{d(d - k)} - \frac{k - 1}{d(d - k)} \Big)}{\sum_{k = 1}^{d - 1} \frac{d - 1}{k(d - k)}} \\
    &= \frac{1}{d \sum_{k = 1}^{d - 1} \frac{1}{k(d - k)}} \\
    &= \frac{1}{2 \sum_{k = 1}^{d - 1} \frac{1}{k}} \\
    &= \frac{1}{2 H_{d - 1}}.
\end{align*}


The minimum eigenvalue is therefore strictly positive, and this implies that $h_{xy}(\phi)$ is $\mu$-strongly convex with $\mu$ given by

\begin{align*}
    \mu &= 2 \cdot \lambda_{\mathrm{min}}(A) = H_{d - 1}\inv.
\end{align*}

Note that the strong convexity constant $\mu$ does not depend on $(x, y)$ and is determined solely by the number of features $d$.

\end{proof}

\textbf{Theorem 1.} \textit{For a model $\phifast(\rvx, \rvy; \theta)$ whose predictions satisfy the constraint in \cref{eq:objective}, the objective value $\mathcal{L}(\theta)$ upper bounds the Shapley value estimation error as follows,
\begin{equation*}
    \E_{p(\rvx, \rvy)} \Big[ \big|\big|\phifast(\rvx, \rvy; \theta) - \phi(v_{\rvx \rvy})\big|\big|_2 \Big]
    \leq \sqrt{2 H_{d - 1} \Big( \mathcal{L}(\theta) - \mathcal{L}^* \Big)},
\end{equation*}
where $\mathcal{L}^*$ represents the loss achieved by the exact Shapley values.
}

\begin{proof}

We first consider a single input-output pair $(x, y)$ with prediction given by $\phi = \phifast(x, y; \theta)$. Rather than writing the expected loss $h_{xy}(\phi)$, we now write the Lagrangian $\mathcal{L}_{xy}(\phi, \gamma)$ to account for the linear constraint
in our objective, see \cref{eq:objective}:

\begin{align*}
    \mathcal{L}_{xy}(\phi, \gamma) = h_{xy}(\phi) + \gamma \big( v_{xy}(\vone) - v_{xy}(\vzero) - \vone^\top \phi \big).
\end{align*}

Regardless of the Lagrange multiplier value $\gamma \in \R$, the Lagrangian $\mathcal{L}_{xy}(\phi, \gamma)$ is a $\mu$-strongly convex quadratic with
the same Hessian as $h_{xy}(\phi)$:

\begin{align*}
    \nabla_\phi^2 \mathcal{L}_{xy}(\phi, \gamma) = \nabla_\phi^2 h_{xy}(\phi).
\end{align*}

Strong convexity enables us to bound $\phi$'s distance to the global minimizer via the Lagrangian's value. First, we denote the Lagrangian's optimizer as $(\phi^*, \gamma^*)$, where $\phi^*$ is given by the exact Shapley values \citep{charnes1988extremal}:

\begin{align*}
    \phi^* = \phi(v_{xy}).
\end{align*}

Next, we use the first-order strong convexity condition to write the following inequality:

\begin{align*}
    \mathcal{L}_{xy}(\phi, \gamma^*) &\geq \mathcal{L}_{xy}(\phi^*, \gamma^*) + (\phi - \phi^*)^\top \nabla_\phi \mathcal{L}_{xy}(\phi^*, \gamma^*) + \frac{\mu}{2} ||\phi - \phi^*||^2.
\end{align*}

According to the Lagrangian's KKT conditions \citep{boyd2004convex}, we have the property that $\nabla_\phi \mathcal{L}_{xy}(\phi^*, \gamma^*) = 0$. The inequality therefore simplifies to

\begin{align*}
    \mathcal{L}_{xy}(\phi, \gamma^*) &\geq \mathcal{L}_{xy}(\phi^*, \gamma^*) + \frac{\mu}{2} ||\phi - \phi^*||^2_2,
\end{align*}

or equivalently,

\begin{align*}
    ||\phi - \phi^*||^2_2 &\leq \frac{2}{\mu} \Big( \mathcal{L}_{xy}(\phi, \gamma^*) - \mathcal{L}_{xy}(\phi^*, \gamma^*) \Big).
\end{align*}

If we constrain $\phi$ to be a feasible solution (i.e., it satisfies our objective's linear constraint),
the KKT primal feasibility condition implies that
the inequality further simplifies to

\begin{align}
    ||\phi - \phi^*||^2_2 &\leq \frac{2}{\mu} \Big( h_{xy}(\phi) - h_{xy}(\phi^*) \Big). \label{eq:upper_bound_single}
\end{align}

Now, we can consider this bound in expectation over $p(\rvx, \rvy)$. First, we denote our full training loss as $\mathcal{L}(\theta)$, which is equal to

\begin{align*}
    \mathcal{L}(\theta) &= \E_{p(\rvx, \rvy)} \E_{\psh(\rvs)} \Big[ \big( v_{\rvx \rvy}(\rvs) - v_{\rvx \rvy}(\vzero) - \rvs^\top \phifast(\rvx, \rvy; \theta) \big)^2 \Big] = \E_{p(\rvx, \rvy)} \Big[ h_{\rvx \rvy}\big(\phifast(\rvx, \rvy; \theta) \big) \Big].
\end{align*}

Next, we denote $\mathcal{L}^*$ as the training loss achieved by the exact Shapley values, or

\begin{align*}
    \mathcal{L}^* = \E_{p(\rvx, \rvy)} \Big[ h_{\rvx \rvy}\big(\phi(v_{\rvx\rvy})\big) \Big].
\end{align*}

Given a network $\phifast(\rvx, \rvy; \theta)$ whose predictions are constrained to satisfy the linear constraint,
taking the bound from \cref{eq:upper_bound_single} in expectation yields the following
bound on the distance between the predicted and exact Shapley values:

\begin{align*}
    \E_{p(\rvx, \rvy)} \Big[ \big|\big|\phifast(\rvx, \rvy; \theta) - \phi(v_{\rvx \rvy})\big|\big|^2_2 \Big] \leq \frac{2}{\mu} \Big( \mathcal{L}(\theta) - \mathcal{L}^* \Big).
\end{align*}

Applying Jensen's inequality to the left side, we can rewrite the bound as follows:

\begin{align*}
    \E_{p(\rvx, \rvy)} \Big[ \big|\big|\phifast(\rvx, \rvy; \theta) - \phi(v_{\rvx \rvy})\big|\big|_2 \Big]
    &\leq \sqrt{\frac{2}{\mu} \Big( \mathcal{L}(\theta) - \mathcal{L}^* \Big)}.
\end{align*}

Substituting in the strong convexity parameter $\mu$ from \Cref{lemma1},
we arrive at the final bound:

\begin{align*}
    \E_{p(\rvx, \rvy)} \Big[ \big|\big|\phifast(\rvx, \rvy; \theta) - \phi(v_{\rvx \rvy})\big|\big|_2 \Big]
    &\leq \sqrt{2 H_{d - 1} \Big( \mathcal{L}(\theta) - \mathcal{L}^* \Big)}.
\end{align*}

\end{proof}

We also present a corollary to \Cref{theorem1}. This result formalizes the intuition that if we can iteratively optimize the explainer such that its loss approaches the optimum, our Shapley value estimation error will go to zero.

\textbf{Corollary 1.} \textit{Given a sequence of models $\phifast(\rvx, \rvy; \theta_1), \phifast(\rvx, \rvy; \theta_2), \ldots$ whose predictions satisfy the constraint in \cref{eq:objective} and where $\mathcal{L}(\theta_n) \to \mathcal{L}^*$, the Shapley value estimation error goes to zero:
\begin{equation*}
    \lim_{n \to \infty} \E_{p(\rvx, \rvy)} \Big[ \big|\big|\phifast(\rvx, \rvy; \theta_n) - \phi(v_{\rvx \rvy})\big|\big|_2 \Big] = 0.
\end{equation*}
}

\begin{proof}


Fix $\epsilon > 0$.
By assumption, there exists a value $n'$ such that $\mathcal{L}(\theta_n) - \mathcal{L}^* < \frac{\mu \epsilon^2}{2}$ for $n > n'$. Following the result in \Cref{theorem1},
we have $\E_{p(\rvx, \rvy)} \Big[ \big|\big|\phifast(\rvx, \rvy; \theta_n) - \phi(v_{\rvx \rvy})\big|\big|_2 \Big] < \epsilon$ for $n > n'$.

\end{proof}

Finally, we also consider the role of our loss function in quantifying the Shapley value estimation error, which we define for a given explainer model $\phifast(\rvx, \rvy; \theta)$ as

\begin{equation*}
    \text{SVE} = \E_{p(\rvx, \rvy)} \Big[ \big|\big|\phifast(\rvx, \rvy; \theta) - \phi(v_{\rvx \rvy})\big|\big|_2 \Big].
\end{equation*}

One natural approach is to use an external dataset (e.g., the test data) consisting of samples $(x_i, y_i)$ for $i = 1, \ldots, n$, calculate their exact Shapley values $\phi(v_{x_iy_i})$, and generate a Monte Carlo estimate as follows:

\begin{equation*}
    \hat{\text{SVE}}_n = \frac{1}{n} \sum_{i = 1}^n \big|\big|\phifast(x_i, y_i; \theta) - \phi(v_{x_i y_i})\big|\big|_2.
\end{equation*}

While standard concentration inequalities allow us to bound $\text{SVE}$ using $\hat{\text{SVE}}_n$, generating the ground truth values can be computationally costly,
particularly for large $n$. Instead, another approach is to use our result from \Cref{theorem1}, which bypasses the need for ground truth Shapley values.
For this, recall that $\mathcal{L}(\theta)$ represents our weighted least squares loss function, where we assume that the explainer $\phifast(\rvx, \rvy; \theta)$ satisfies the constraint in \cref{eq:objective} for all predictions. If we know $\mathcal{L}(\theta)$ exactly, then \Cref{theorem1} yields the following bound with probability 1:

\begin{equation*}
    \text{SVE} \leq \sqrt{2 H_{d - 1} \Big( \mathcal{L}(\theta) - \mathcal{L}^* \Big)}.
\end{equation*}

If we do not know $\mathcal{L}(\theta)$ exactly, we can instead form a Monte Carlo estimate $\hat{\mathcal{L}}(\theta)_n$ using samples $(x_i, y_i, s_i)$ for $i = 1, \ldots, n$. Then, using concentration inequalities like Chebyshev or Hoeffding (the latter only applies if we assume bounded errors), we can get probabilistic bounds of the form $\text{P}(|\mathcal{L}(\theta) - \hat{\mathcal{L}}_n| > \epsilon) \leq \delta$. With these, we can say with probability at least $1 - \delta$ that
$\mathcal{L}(\theta) \leq \hat{\mathcal{L}}(\theta)_n + \epsilon$.
Finally, combining this with the last steps of our \Cref{theorem1} proof, we obtain the following bound with probability at least $1 - \delta$:

\begin{equation}
    \text{SVE} \leq \sqrt{2 H_{d - 1} \Big( \hat{\mathcal{L}}(\theta)_n - \mathcal{L}^* + \epsilon \Big)}.
\end{equation}

Naturally, $\delta$ is a function of $\epsilon$ and the number of samples $n$ used to estimate $\hat{\mathcal{L}}(\theta)_n$, with the rate of convergence to probability 1 depending on the choice of concentration inequality (Chebyshev or Hoeffding).
Although this procedure yields an inexpensive upper bound on the Shapley value estimation error, the bound's looseness, as well as the fact that we do not know $\mathcal{L}^*$ a priori, make it unappealing
as an evaluation metric. The more important takeaways are 1)~that training with the loss $\mathcal{L}(\theta)$ effectively minimizes an upper bound on the Shapley value estimation error, and 2)~that comparing explainer models via their validation loss, which is effectively $\hat{\mathcal{L}}(\theta)_n$, is a principled approach to perform model selection and hyperparameter tuning.


\section{Datasets} \label{app:datasets}

The ImageNette dataset contains 9,469 training examples and 3,925 validation examples, and we split the validation data to obtain validation and test sets containing 1,962 examples each. The MURA dataset contains 36,808 training examples and 3,197 validation examples. We use the validation examples as a test set, and we split the training examples to obtain train and validation sets containing 33,071 and 3,737 examples, ensuring that images from the same patient belong to a single split. The Oxford-IIIT Pets dataset contains 7,349 examples for 37 classes, and we split the data to obtain train, validation, and test sets containing 5,879, 735, and 735 examples, respectively.
For all datasets, the training and validation data were used to train the original classifiers, fine-tuned classifiers and explainer models, and the test data was used only when calculating
performance metrics.



\section{Baseline methods} \label{app:baselines}

This section provides implementation details for the baseline explanation methods. We used a variety of attention-, gradient- and removal-based methods as comparisons for ViT Shapley, and we modified several approaches
to arrive at patch-level feature attribution scores.

\paragraph{Attention last} This approach
calculates the attention directed from each image token into the class token in the final self-attention layer, summed across attention heads \citep{abnar2020quantifying, chefer2021transformer}. The results are
provided at the patch-level, but they are not generated separately for each output class.

\paragraph{Attention rollout} This approach accounts for the flow of attention between tokens by summing across attention heads and multiplying the resulting attention matrices at each layer \citep{abnar2020quantifying}.
Like the previous method, results are not generated separately for each output class. We used an implementation provided by prior work \citep{chefer2021transformer}.


\paragraph{Common gradient-based methods} Several methods that operate via input gradients are Vanilla gradients \citep{simonyan2013deep}, SmoothGrad \citep{smilkov2017smoothgrad}, VarGrad \citep{hooker2019benchmark}, and IntGrad \citep{sundararajan2017axiomatic}. These methods were run using the Captum package \citep{kokhlikyan2020captum}, and we used 10 samples per image for SmoothGrad, VarGrad and IntGrad. We tried applying these
at the level of pixels and patch embeddings, and in both cases
we created
class-specific, patch-level attributions by summing
across the unnecessary dimensions.
We calculated the absolute value before summing for Vanilla and SmoothGrad, VarGrad automatically produces non-negative values, and we preserved the sign for IntGrad because it should be meaningful.

\input{figures/metrics_supplement}

\paragraph{GradCAM} Originally designed for
intermediate convolutional layers \citep{selvaraju2017grad}, GradCAM has since been generalized to the ViT context.
The main operations remain the same, only the representation being analyzed is 
the layer-normed input to the final self-attention layer,
and the aggregation is across the embedding dimension
rather than convolutional channels (GradCAM LN) \citep{jacobgilpytorchcam}.
We also experimented with using a different internal layer for generating explanations
(the attention weights computed in the final self-attention layer, denoted as GradCAM Attn. \citep{chefer2021transformer}).

\paragraph{Layer-wise relevance propagation (LRP)} Originally described as a set of constraints for a modified backpropagation routine \citep{bach2015pixel},
LRP has since been implemented
for a variety of network layers and architectures,
and it was recently adapted to ViTs \citep{chefer2021transformer}.
We used an implementation provided by prior work \citep{chefer2021transformer}.

\paragraph{Leave-one-out} The importance scores in this approach are the difference in prediction probability for the full-image and the iamge with a single patch removed. We removed patches by setting pixels to zero, 
similar to the original version for CNNs \citep{zeiler2014visualizing}.

\paragraph{RISE} This approach involves sampling many
occlusion masks and reporting the mean prediction when each patch is included. The original version for CNNs \citep{petsiuk2018rise} used a complex approach to generate masks,
but we simply sampled subsets of patches. As in the original work, we sample from all subsets with equal probability, and we use 2,000 mask samples per sample to be explained.
We occlude patches by setting pixel values to zero, similar to the original work.

\paragraph{Random} Finally, we included a random baseline as a comparison for the insertion, deletion and ROAR metrics. These metrics only require a ranking of important patches, so we generated ten random orderings 
and averaged the results across these orderings.

\Cref{tab:metrics_supplement} shows the same metrics as \Cref{tab:metrics} with additional results for alternative implementations of several baselines. For the methods based on input gradients, 
we experimented with generating explanations at both the pixel level and embedding level; the preferred approach depends on the method and metric, but both versions
tend to underperform
ViT Shapley, with the exception of faithfulness on ImageNette where the 95\% confidence intervals overlap for many methods. We also experimented with two versions of GradCAM (described above) and find that the GradCAM LN implementation generally performs slightly better. In the main text, we present results only for GradCAM LN and the remaining gradient-based methods generated at the embedding level.

\section{Metrics details} \label{app:metrics}

This section provides additional details about the performance metrics used in the main text experiments (\Cref{sec:experiments}).

\paragraph{Insertion/deletion} These metrics involve repeatedly making predictions while either inserting or deleting features in order of most to least important \citep{petsiuk2018rise}. While the original work removed features by setting them to zero, we use the fine-tuned classifier
that was trained to handle partial information. We calculated the area under the curve for individual predictions and then averaged the results across 1,000 test set examples; we used random examples for ImageNette, and only examples that were predicted to be abnormal for MURA. \Cref{tab:metrics} presents results for the true class only, and \Cref{tab:metrics_nontarget} presents results averaged across all the remaining classes.

\paragraph{Sensitivity-n} This metric samples feature subsets at random and calculates the correlation between the prediction with each subset and the sum of the corresponding features' attribution scores \citep{ancona2018towards}. It typically considers subsets of a fixed size $n$, which means sampling from the following subset distribution $p_n(s)$:

\begin{equation*}
    p_n(s) = \mathbbm{1}(\vone^\top s = n) \binom{d}{n}\inv.
\end{equation*}

Mathematically, the metric is defined for a model $f(\rvx)$, an individual sample $x$ and label $y$, feature attributions $\phi \in \R^d$ and subset size $n$ as follows:

\begin{equation*}
    \mathrm{Sens}(f, x, y, \phi, n) = \mathrm{Corr}_{p_n(\rvs)}\big(\rvs^\top \phi, f_y(x) - f_y(x_{\vone - \rvs})\big).
\end{equation*}

Similar to insertion/deletion, we use the fine-tuned classifier
to handle held-out patches and calculate the metric across
1,000 test set images.
We use 
subset sizes ranging from 14 to 182 patches with step size 14, and we estimate the correlation for each example and subset size using 1,000 subset samples. 

\paragraph{Faithfulness} This metric is nearly identical to sensitivity-n, only it calculates the correlation
across subsets of all sizes \citep{bhatt2021evaluating}. Mathematically, it is defined as

\begin{equation*}
    \mathrm{Faith}(f, x, y, \phi) = \mathrm{Corr}_{p(\rvs)}\big(\rvs^\top \phi, f_y(x) - f_y(x_{\vone - \rvs})\big),
\end{equation*}

and we sample from a distribution with
equal probability mass on all cardinalities, or

\begin{equation*}
    p(s) = \frac{1}{\binom{d}{\vone^\top s} \cdot (d + 1)}.
\end{equation*}

We use the fine-tuned classifier
to handle held-out patches, and we compute faithfulness across 1,000 test set images and with 1,000 subset samples per image.

\paragraph{ROAR} Finally, ROAR evaluates the model's accuracy after removing features in order from most to least important \citep{hooker2019benchmark}. We also experimented with \textit{inserting} features in order of most to least important.
Crucially, the ROAR authors propose handling held-out features by retraining the model with masked inputs. We performed masked retraining by performing test-time augmentations for all training, validation and test set images, generating explanations to identify
the most important patches for the true class, and setting the corresponding pixels to zero.

Because masked retraining leaks
information through the masking, we also replicated this metric using the fine-tuned classifier
model, and with a separate evaluator model trained directly with random masking; the evaluator model trained with random masking has been used in prior work \citep{jethani2021have, jethani2021fastshap}. We generated results for each number of inserted/deleted patches (1, 3, 7, 14, 28, 56, 84, 112, 140, 168, and 182) with the final accuracy computed across the entire test set.

\paragraph{Ground truth metrics} Previous work has considered evaluations involving comparison with ground truth importance, where the ground truth is either identified by humans \citep{chefer2021transformer} or introduced via synthetic dataset modifications \citep{zhou2022feature}. An important issue with such methods is that they test explanations against what a model should depend on rather than what it does depend on, so the results do not reflect the explanation's accuracy for the specific model \citep{petsiuk2018rise}. We thus decided against including such metrics.

\section{Additional results} \label{app:results}

This section provides additional experimental results. We first show results involving similar baselines and metrics as in the main text, and we then show results comparing ViT Shapley to KernelSHAP.

\subsection{Main baselines and metrics}

\Cref{fig:removal_accuracy} shows our evaluation of attention masking for handling held-out image patches using two separate metrics: 1)~KL divergence relative to the full-image predictions (also shown in the main text), and 2)~top-1 accuracy relative to the true labels. The former can be understood as a divergence measure between the predictions with masked inputs and the predictions with patches marginalized out using their conditional distribution (see \Cref{app:masked_training}). The latter is a more intuitive measure of how much the performance degrades given partial inputs. The results are similar between the two metrics, showing that the predictions diverge more quickly if the model is not fine-tuned with random masking.

\Cref{tab:metrics_mura_nontarget} shows insertion, deletion and faithfulness results for the MURA dataset with examples that were predicted to be normal, but while evaluating explanations for the abnormal class. ViT Shapley outperforms the baseline methods, reflecting that our explanations correctly identify patches that influence the prediction towards and against the abnormal class even for normal examples.

\Cref{tab:metrics_supplement_pet} shows insertion, deletion and faithfulness results for the Pets dataset. We observe that ViT Shapley outperforms other methods for all metrics with the exception of faithfulness for target classes, where 95\% confidence intervals overlap for many methods (similar to the other datasets).

\Cref{tab:metrics_supplement_vittiny},
\Cref{tab:metrics_supplement_vitsmall},
and \Cref{tab:metrics_supplement_vitlarge} show insertion, deletion, and faithfulness metrics for ImageNette
when using
other ViT architectures (i.e., ViT-Tiny, -Small, and -Large, respectively) \citep{rw2019timm, dosovitskiy2020image}  for the
classifier
and explainer. They show results for target-class explanations and non-target-class explanations, respectively. The results are consistent with those obtained for ViT-Base,
and ViT Shapley outperforms the baseline methods across all three metrics.
This shows that our explainer model can be trained successfully
with architectures of different sizes, including when using a relatively small number of parameters.

\Cref{tab:metrics_supplement_direct_masking} shows insertion, deletion and faithfulness results for a ViT classifier trained directly with random masking. Whereas our \Cref{sec:experiments} experiments utilize a fine-tuned classifier
to handle missing patches, a classifier trained with random masking allows us to bypass the fine-tuning stage
and train the explainer directly. The results are similar to \Cref{tab:metrics}, and we find that ViT Shapley consistently achieves the best performance.

\Cref{fig:insertion_curves_imagenette}, \Cref{fig:insertion_curves_mura}, and \Cref{fig:insertion_curves_pet} show the average curves used to generate the insertion/deletion AUC results. All sets of plots reflect that explanations from ViT Shapley
identify relevant patches that quickly move the prediction towards or away from a given class. In the case of ImageNette and Pets, we observe that this holds for both target and non-target classes.

\Cref{fig:sensitivity_nontarget} shows the sensitivity-n metric evaluated for non-target classes on the ImageNette dataset. Similarly, these results show that ViT Shapley generates attribution scores that represent
the impact of withholding features from a model, even for non-target classes. In this case, RISE and leave-one-out are more competitive with ViT Shapley, but their performance is less competitive
when the correlation is calculated for subsets of all sizes (see faithfulness in \Cref{tab:metrics_nontarget}).

Next, \Cref{fig:roar_all} shows ROAR results generated in both the insertion and deletion directions, using the four patch removal approaches: 1)~the fine-tuned classifier, 2)~the separate evaluator model trained directly with random masking, 3)~masked retraining, and 4)~masked retraining without positional embeddings. The results show that in addition to strong performance in the deletion direction, ViT Shapley consistently achieves the best results in the insertion direction, even in the case of masked retraining with positional embeddings.


\Cref{fig:roar_vitsmall} shows ROAR results for the same settings, but when using the ViT-Small architecture. We observe the same results obtained with ViT-Base. Except for the deletion direction with masked retraining and positional embeddings enabled, ViT Shapley achieves the best performance among all methods.

Finally, \Cref{tab:times} shows the time required to generate explanations using each approach. Because ViT Shapley requires a single forward pass through the explainer model, it is among the fastest approaches and is paralleled only by the attention-based methods.
The gradient-based methods require
forward and backward passes for all classes, and sometimes for many altered inputs (e.g., with noise injected for SmoothGrad). RISE is the slowest of all the approaches tested because it requires making several thousand predictions to explain each sample. Our evaluation was conducted on a
GeForce RTX 2080 Ti GPU, with minibatches of 16 samples for attention last, attention rollout and ViT Shapley; batch size of 1 for Vanilla Gradients, GradCAM, LRP, leave-one-out and RISE; and internal minibatching for SmoothGrad, IntGrad and VarGrad (implemented via Captum \citep{kokhlikyan2020captum}).

ViT Shapley is the only method considered here to require training time, and as described in \Cref{app:method}, training the explainer models required roughly 0.8 days for ImageNette and 2.5 days for MURA. The training time is not insignificant, but investing time in training the explainer is worthwhile if 1)~high-quality explanations are required, 2)~there are many examples to be explained (e.g., an entire dataset), or 3)~fast explanations are required during a model’s deployment.

\input{figures/removal_accuracy.tex}

\input{figures/metrics_mura_nontarget}

\input{figures/metrics_supplement_pet}

\input{figures/metrics_supplement_vittiny}

\input{figures/metrics_supplement_vitsmall}

\input{figures/metrics_supplement_vitlarge} 

\input{figures/metrics_supplement_direct_masking}

\input{figures/insertion_curves_imagenette}

\input{figures/insertion_curves_mura}

\input{figures/insertion_curves_pet}

\input{figures/sensitivity_nontarget}

\input{figures/roar_all.tex}

\input{figures/roar_vitsmall}

\input{figures/times}

\clearpage
\subsection{KernelSHAP comparisons}

Here, we provide two results comparing ViT Shapley with KernelSHAP.

First, \Cref{fig:kernelshap_error} compares the approximation quality of Shapley value estimates produced by ViT Shapley and KernelSHAP. The estimates are evaluated in terms of L2 distance, Pearson correlation and Spearman (rank) correlation, and our ground truth is generated by running KernelSHAP for a large number of iterations. Specifically, we use the convergence detection approach described by \cite{covert2021improving}
with a threshold of $t = 0.1$. The results are computed using just 100 randomly selected ImageNette images due to the significant computational cost.

Based on \Cref{fig:kernelshap_error}, we observe that the original version of KernelSHAP takes roughly 120,000 model evaluations to reach the accuracy that ViT Shapley reaches with a single model evaluation. KernelSHAP with paired sampling \citep{covert2021improving} converges faster, and it requires roughly 40,000 model evaluations on average. ViT Shapley's estimates are not perfect, but they reach nearly 0.8 correlation with the
ground truth for the target class, and nearly 0.7 correlation on average across non-target classes.

Next, \Cref{tab:kernelshap_insertdelete} compares the
ViT Shapley estimates and the fully converged estimates from KernelSHAP via the insertion and deletion metrics. The fully converged KernelSHAP estimates performed better than ViT Shapley on both metrics, and the gap is largest for deletion with the target class. The results were also computed using only 100 ImageNette examples due to the computational cost.
These results reflect that there is room for further improvement
if ViT Shapley's estimates can be made more accurate. KernelSHAP itself is not a viable option in practice, as we found that its estimates took between 30 minutes and 2 hours to converge when using paired sampling \citep{covert2021improving} (equivalent to roughly 300k and 1,200k model evaluations), but it represents an upper bound on how well ViT Shapley could
perform with near-perfect estimation quality.

\input{figures/kernelshap_insertion_deletion.tex}

\input{figures/kernelshap_error.tex}

\clearpage
\section{Qualitative examples} \label{app:qualitative}

This section provides qualitative examples for ViT Shapley and the baseline methods. 
When visualizing explanations from each method, we used the \textit{icefire} color palette, a diverging colormap implemented in the Python Seaborn package \citep{Waskom2021}.
Negative influence is an important feature for ViT Shapley, and a diverging colormap allows us to highlight
both positive and negative contributions.
To generate the plots shown in this paper, we first calculated the maximum
absolute value of an explanation and then rescaled the values
to each end of the color map; next, we plotted the original image with an alpha of 0.85, and finally we performed bilinear upsampling on the explanation and overlaid
the color-mapped result with an alpha of 0.9. The alpha values can be tuned to control the visibility of the original image.

\Cref{fig:kernelshap} shows a comparison between
ViT Shapley and KernelSHAP explanations for several examples from the ImageNette dataset. The results are nearly identical, but ViT Shapley produces explanations in a single forward pass while KernelSHAP is considerably slower. Determining the number of samples required for KernelSHAP to converge is challenging, and we used the approach proposed by prior work with a convergence threshold of $t = 0.2$ \citep{covert2021improving}. With this setting and with acceleration using paired sampling, the KernelSHAP explanations required between
30 minutes to 1 hour to generate per image, versus a single forward pass for ViT Shapley.

\Cref{fig:qualitative_baselines_imagenette_1}, \Cref{fig:qualitative_baselines_imagenette_2} and \Cref{fig:qualitative_baselines_imagenette_3} show comparisons between ViT Shapley and the baselines on ImageNette samples. We only show results for attention last, attention rollout, Vanilla Gradients, Integrated Gradients, SmoothGrad, LRP, leave-one-out, and ViT Shapley;
we excluded VarGrad, GradCAM and RISE because their results were less visually appealing.
The explanations are shown only for the target class, and we observe that ViT Shapley often highlights the main object of interest. We also find that the model is prone to
confounders, 
as ViT Shapley often highlights parts of the background that are correlated with the true class (e.g., the face of a man holding a tench, the clothes of a man holding a chainsaw, the sky in a parachute image).

Similarly, \Cref{fig:qualitative_baselines_MURA_1}, \Cref{fig:qualitative_baselines_MURA_2} and \Cref{fig:qualitative_baselines_MURA_3} compare ViT Shapley to the baselines on example images
from the MURA dataset. The ViT Shapley explanations almost always highlight clear signs of abnormality,
which are only sometimes highlighted by the baselines. Among those shown here, LRP is most similar to ViT Shapley, but they disagree in several cases.

Next, \Cref{fig:qualitative_baselines_pet_1}, \Cref{fig:qualitative_baselines_pet_2}, \Cref{fig:qualitative_baselines_pet_3}, and \Cref{fig:qualitative_baselines_pet_4} compare ViT Shapley to the baselines on example images from the Pets dataset. We showed one randomly sampled image per each class (i.e., breed). We observe that ViT Shapley highlights distinctive features of breeds (e.g., the mouth of Boxer or the fur pattern of Egyptian Mau), and rarely puts significant importance on background patches. 

Finally, \Cref{fig:qualitative_baselines_nontarget} shows several examples of non-target class explanations. These results show that ViT Shapley highlights patches that can push the model's prediction towards certain non-target classes, which is not the case for other methods. These results corroborate those in \Cref{tab:metrics_nontarget} and \Cref{tab:metrics_mura_nontarget}, which show that ViT Shapley offers the most accurate
class-specific explanations.



\input{figures/kernelshap}

\input{figures/qualitative_baselines_imagenette_1}
\input{figures/qualitative_baselines_imagenette_2}
\input{figures/qualitative_baselines_imagenette_3}

\input{figures/qualitative_baselines_mura_1}
\input{figures/qualitative_baselines_mura_2}
\input{figures/qualitative_baselines_mura_3}

\input{figures/qualitative_baselines_pet_1}
\input{figures/qualitative_baselines_pet_2}
\input{figures/qualitative_baselines_pet_3}
\input{figures/qualitative_baselines_pet_4}

\input{figures/qualitative_baselines_nontarget}

%% file: figures/ablations.tex
\begin{table}[h]
\centering
\caption{
Ablation experiments for ViT Shapley explainer architecture on the ImageNette dataset, with and without fine-tuning.
} \label{tab:ablations}
\vskip 0.1in
\begin{center}
\begin{small}
    \begin{tabular}{ccccc}
    \toprule
    Fine-tuning config. & Extra attention block & Frozen backbone & Val loss & Test loss \\
    \midrule
    A & False & True & 4.332 & 4.351 \\
    B & False & False & 4.331 & 4.351 \\
    C & True & True & 4.319 & 4.339 \\
    D & True & False & \textbf{4.309} & \textbf{4.318} \\
    \midrule
    \multicolumn{3}{c}{ViT trained from scratch} & 4.331 & 4.341 \\
    \midrule
    \multicolumn{3}{c}{U-Net trained from scratch} & 4.332 & 4.338 \\
    \bottomrule
    \end{tabular}
\end{small}
\end{center}
\end{table}

%% file: figures/metrics_supplement.tex
\begin{table}[t]
\centering
\caption{Performance metrics for target-class explanations
with additional baselines. Methods that fail to outperform the random baseline are shown in gray, and the best results are shown in bold (accounting for 95\% confidence intervals).
} \label{tab:metrics_supplement}
\vskip 0.1in
\begin{center}
{\tablesize
    \begin{tabular}{lcccccc}
    \toprule
     & \multicolumn{3}{c}{ImageNette} & \multicolumn{3}{c}{MURA} \\
    \cmidrule(lr){2-4} \cmidrule(lr){5-7}
     & Ins. ($\uparrow$) & Del. ($\downarrow$) & Faith. ($\uparrow$) & Ins. ($\uparrow$) & Del. ($\downarrow$) & Faith. ($\uparrow$) \\
    \midrule
    Attention last & 0.962 (0.004) & 0.793 (0.013) & \textbf{0.694 (0.015)} & 0.890 (0.010) & 0.592 (0.013) & 0.635 (0.016) \\
    Attention rollout & \textcolor{gray}{0.938 (0.005)} & 0.880 (0.010) & \textbf{0.704 (0.015)} & \textcolor{gray}{0.845 (0.011)} & 0.692 (0.014) & 0.618 (0.016) \\\midrule
    GradCAM (LN) & \textcolor{gray}{0.914 (0.006)} & 0.937 (0.008) & 0.680 (0.015) & 0.899 (0.009) & 0.681 (0.015) & 0.631 (0.016) \\
    GradCAM (Attn) & \textcolor{gray}{0.938 (0.006)} & \textcolor{gray}{0.948 (0.006)} & 0.656 (0.014) & \textcolor{gray}{0.843 (0.012)} & \textcolor{gray}{0.835 (0.011)} & 0.580 (0.016) \\
    IntGrad (Pixel) & 0.967 (0.004) & 0.930 (0.008) & 0.403 (0.024) & 0.897 (0.010) & 0.796 (0.015) & 0.201 (0.022) \\
    IntGrad (Embed.) & 0.967 (0.004) & 0.930 (0.008) & 0.403 (0.024) & 0.897 (0.010) & 0.796 (0.015) & 0.201 (0.022) \\
    Vanilla (Pixel) & \textcolor{gray}{0.938 (0.005)} & 0.860 (0.011) & \textbf{0.700 (0.015)} & 0.890 (0.010) & 0.561 (0.014) & 0.627 (0.016) \\
    Vanilla (Embed.) & \textcolor{gray}{0.950 (0.004)} & 0.808 (0.013) & \textbf{0.703 (0.015)} & 0.890 (0.010) & 0.537 (0.014) & 0.629 (0.016) \\
    SmoothGrad (Pixel) & 0.960 (0.005) & 0.779 (0.013) & \textbf{0.706 (0.015)} & 0.873 (0.010) & 0.634 (0.014) & 0.618 (0.016) \\
    SmoothGrad (Embed.) & \textcolor{gray}{0.947 (0.005)} & \textcolor{gray}{0.942 (0.006)} & \textbf{0.703 (0.015)} & 0.870 (0.010) & 0.813 (0.011) & 0.617 (0.016) \\
    VarGrad (Pixel) & \textcolor{gray}{0.958 (0.005)} & 0.796 (0.013) & \textbf{0.682 (0.015)} & 0.871 (0.010) & 0.660 (0.013) & 0.577 (0.015) \\
    VarGrad (Embed.) & \textcolor{gray}{0.949 (0.005)} & \textcolor{gray}{0.946 (0.005)} & \textbf{0.700 (0.015)} & \textcolor{gray}{0.857 (0.011)} & 0.823 (0.011) & 0.615 (0.016) \\
    LRP & 0.967 (0.004) & 0.779 (0.014) & \textbf{0.705 (0.015)} & 0.900 (0.009) & 0.551 (0.013) & 0.646 (0.016) \\\midrule
    Leave-one-out & 0.969 (0.002) & 0.917 (0.010) & 0.140 (0.040) & 0.926 (0.008) & 0.694 (0.017) & 0.308 (0.032) \\
    RISE & 0.977 (0.001) & 0.860 (0.014) & \textbf{0.704 (0.015)} & 0.957 (0.004) & 0.573 (0.018) & 0.618 (0.016) \\
    \textbf{ViT Shapley} & \textbf{0.985 (0.002)} & \textbf{0.691 (0.014)} & \textbf{0.711 (0.015)} & \textbf{0.971 (0.002)} & \textbf{0.307 (0.013)} & \textbf{0.707 (0.013)} \\\midrule
    Random & \textcolor{gray}{0.951 (0.005)} & \textcolor{gray}{0.951 (0.005)} & - & \textcolor{gray}{0.849 (0.010)} & \textcolor{gray}{0.847 (0.010)} & - \\    
    \bottomrule
    \end{tabular}
}
\end{center}
\end{table}

%% file: figures/removal_accuracy.tex
\begin{figure}[t]
    \figurespacing
    \centering
    \vskip -0.2in
    \includegraphics[width=1.0\linewidth]{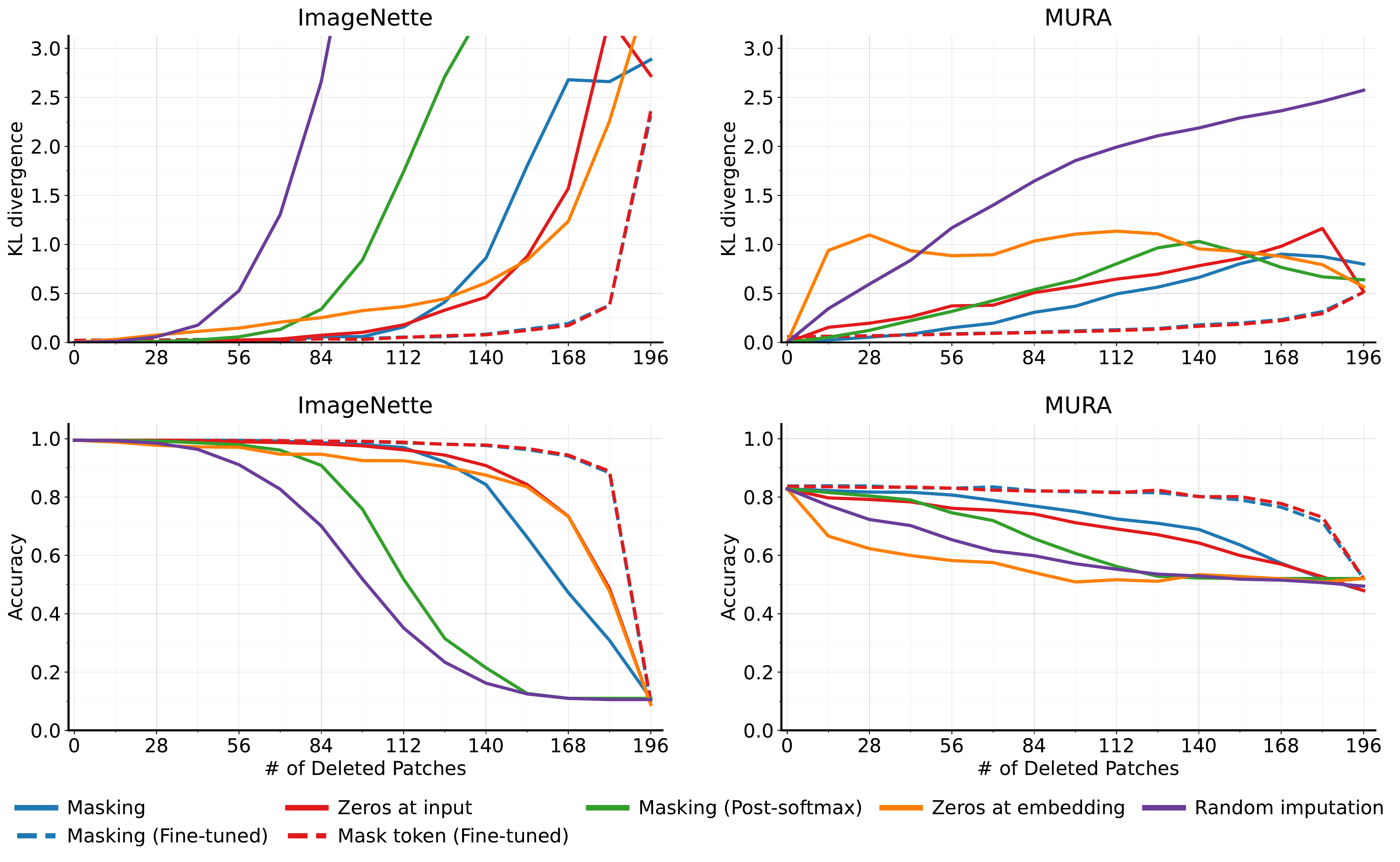}
    \caption{
    ViT predictions given
    partial information.
    We delete patches at random using several removal mechanisms, and we then measure the quality of the resulting predictions via two metrics: the
    KL divergence relative to
    the original, full-image
    predictions (top), and the top-1 accuracy relative to the true labels (bottom).
    }
    \label{fig:removal_accuracy}
\end{figure}

%% file: figures/metrics_mura_nontarget.tex
\begin{table}[t]
\centering
\caption{
MURA non-target metrics for images that were predicted to be normal. Methods that fail to outperform the random baseline are shown in gray, and the best results are shown in bold (accounting for 95\% confidence intervals).
} \label{tab:metrics_mura_nontarget}
\vskip 0.1in
\begin{center}
{\tablesize
    \begin{tabular}{lccc}
    \toprule
     & \multicolumn{3}{c}{MURA} \\
    \cmidrule(lr){2-4} 
     & Ins. ($\uparrow$) & Del. ($\downarrow$) & Faith. ($\uparrow$) \\
    \midrule

    Attention last & \textcolor{gray}{0.157 (0.011)} & \textcolor{gray}{0.210 (0.009)} & -0.455 (0.022) \\
    Attention rollout & 0.199 (0.011) & \textcolor{gray}{0.169 (0.010)} & -0.480 (0.023) \\\midrule
    GradCAM & 0.197 (0.012) & \textcolor{gray}{0.152 (0.010)} & -0.449 (0.022) \\
    IntGrad & 0.209 (0.014) & \textcolor{gray}{0.151 (0.010)} & 0.103 (0.023) \\
    Vanilla & \textcolor{gray}{0.174 (0.011)} & \textcolor{gray}{0.197 (0.009)} & -0.477 (0.023) \\
    SmoothGrad & \textcolor{gray}{0.169 (0.011)} & \textcolor{gray}{0.176 (0.011)} & -0.480 (0.023) \\
    VarGrad & \textcolor{gray}{0.166 (0.011)} & \textcolor{gray}{0.175 (0.011)} & -0.478 (0.023) \\
    LRP & \textcolor{gray}{0.169 (0.012)} & \textcolor{gray}{0.200 (0.009)} & -0.461 (0.023) \\\midrule
    Leave-one-out & 0.326 (0.016) & 0.093 (0.007) & 0.272 (0.029) \\
    RISE & 0.406 (0.018) & 0.075 (0.006) & -0.479 (0.023) \\
    \textbf{ViT Shapley} & \textbf{0.580 (0.016)} & \textbf{0.039 (0.003)} & \textbf{0.642 (0.014)} \\\midrule
    Random & \textcolor{gray}{0.169 (0.011)} & \textcolor{gray}{0.170 (0.011)} & - \\    
    \bottomrule
    \end{tabular}
}
\end{center}
\end{table}

%% file: figures/metrics_supplement_pet.tex
\begin{table}[t]
\centering
\caption{Performance metrics
for ViT-Base on Pets. Methods that fail to outperform the random baseline are shown in gray, and the best results are shown in bold (accounting for 95\% confidence intervals).
} \label{tab:metrics_supplement_pet}
\vskip 0.1in
\begin{center}
\begin{small}
{\tablesize
    \begin{tabular}{lcccccc}
    \toprule
     & \multicolumn{3}{c}{Target} & \multicolumn{3}{c}{Non-Target} \\
    \cmidrule(lr){2-4} \cmidrule(lr){5-7}
     & Ins. ($\uparrow$) & Del. ($\downarrow$) & Faith. ($\uparrow$) & Ins. ($\uparrow$) & Del. ($\downarrow$) & Faith. ($\uparrow$) \\
    \midrule
    Attention last & 0.876 (0.011) & 0.494 (0.016) & 0.545 (0.017) & - & - & - \\
    Attention rollout & \textcolor{gray}{0.837 (0.012)} & 0.672 (0.017) & \textbf{0.559 (0.017)} & - & - & - \\\midrule
    GradCAM (Attn) & \textcolor{gray}{0.837 (0.013)} & 0.737 (0.018) & 0.527 (0.016) & \textcolor{gray}{0.005 (0.000)} & \textcolor{gray}{0.008 (0.000)} & -0.499 (0.013) \\
    GradCAM (LN) & \textcolor{gray}{0.837 (0.011)} & 0.717 (0.019) & 0.553 (0.017) & 0.012 (0.001) & 0.004 (0.000) & -0.511 (0.014) \\
    IntGrad (Pixel) & 0.899 (0.010) & 0.802 (0.016) & 0.437 (0.017) & 0.006 (0.001) & 0.003 (0.000) & 0.126 (0.008) \\
    IntGrad (Embed.) & 0.899 (0.010) & 0.802 (0.016) & 0.437 (0.017) & 0.006 (0.001) & 0.003 (0.000) & 0.126 (0.008) \\
    Vanilla (Pixel) & \textcolor{gray}{0.862 (0.011)} & 0.596 (0.018) & \textbf{0.559 (0.017)} & \textcolor{gray}{0.004 (0.000)} & \textcolor{gray}{0.011 (0.000)} & -0.526 (0.014) \\
    Vanilla (Embed.) & 0.876 (0.011) & 0.519 (0.017) & \textbf{0.561 (0.017)} & \textcolor{gray}{0.004 (0.000)} & \textcolor{gray}{0.013 (0.000)} & -0.529 (0.014) \\
    SmoothGrad (Pixel) & 0.889 (0.011) & 0.488 (0.016) & \textbf{0.564 (0.017)} & \textcolor{gray}{0.003 (0.000)} & \textcolor{gray}{0.014 (0.000)} & -0.532 (0.014) \\
    SmoothGrad (Embed.) & \textcolor{gray}{0.840 (0.013)} & \textcolor{gray}{0.834 (0.013)} & \textbf{0.558 (0.017)} & \textcolor{gray}{0.004 (0.000)} & \textcolor{gray}{0.005 (0.000)} & -0.526 (0.014) \\
    VarGrad (Pixel) & 0.891 (0.011) & 0.513 (0.016) & \textbf{0.557 (0.017)} & \textcolor{gray}{0.003 (0.000)} & \textcolor{gray}{0.013 (0.000)} & -0.522 (0.014) \\
    VarGrad (Embed.) & \textcolor{gray}{0.847 (0.012)} & \textcolor{gray}{0.835 (0.013)} & 0.555 (0.017) & \textcolor{gray}{0.004 (0.000)} & \textcolor{gray}{0.005 (0.000)} & -0.523 (0.014) \\
    LRP & 0.890 (0.011) & 0.466 (0.016) & \textbf{0.567 (0.017)} & \textcolor{gray}{0.004 (0.000)} & \textcolor{gray}{0.013 (0.000)} & -0.529 (0.014) \\\midrule
    Leave-one-out & \textbf{0.936 (0.006)} & 0.653 (0.022) & 0.215 (0.036) & 0.018 (0.001) & 0.001 (0.000) & 0.138 (0.023) \\
    RISE & \textbf{0.946 (0.005)} & 0.505 (0.022) & \textbf{0.559 (0.017)} & 0.032 (0.002) & 0.001 (0.000) & -0.523 (0.014) \\
    \textbf{ViT Shapley} & \textbf{0.945 (0.006)} & \textbf{0.388 (0.017)} & \textbf{0.590 (0.016)} & \textbf{0.052 (0.002)} & \textbf{0.001 (0.000)} & \textbf{0.529 (0.012)} \\\midrule
    Random & \textcolor{gray}{0.848 (0.012)} & \textcolor{gray}{0.845 (0.012)} & - & \textcolor{gray}{0.004 (0.000)} & \textcolor{gray}{0.004 (0.000)} & - \\
    \bottomrule
    \end{tabular}
}
\end{small}
\end{center}
\end{table}

%% file: figures/metrics_supplement_vittiny.tex
\begin{table}[t]
\centering
\caption{Performance metrics
for ViT-Tiny on ImageNette. Methods that fail to outperform the random baseline are shown in gray, and the best results are shown in bold (accounting for 95\% confidence intervals).
} \label{tab:metrics_supplement_vittiny}
\vskip 0.1in
\begin{center}
{\tablesize
    \begin{tabular}{lcccccc}
    \toprule
     & \multicolumn{3}{c}{Target} & \multicolumn{3}{c}{Non-Target} \\
    \cmidrule(lr){2-4} \cmidrule(lr){5-7}
     & Ins. ($\uparrow$) & Del. ($\downarrow$) & Faith. ($\uparrow$) & Ins. ($\uparrow$) & Del. ($\downarrow$) & Faith. ($\uparrow$) \\
    \midrule
    Attention last & \textcolor{gray}{0.917 (0.008)} & 0.702 (0.015) & 0.649 (0.014) & - & - & - \\
    Attention rollout & \textcolor{gray}{0.901 (0.009)} & 0.698 (0.015) & 0.658 (0.015) & - & - & - \\\midrule
    GradCAM (LN) & \textcolor{gray}{0.918 (0.007)} & 0.809 (0.014) & 0.659 (0.014) & 0.034 (0.002) & 0.006 (0.000) & -0.616 (0.013) \\
    GradCAM (Attn) & \textcolor{gray}{0.898 (0.009)} & 0.814 (0.013) & 0.636 (0.014) & \textcolor{gray}{0.011 (0.001)} & \textcolor{gray}{0.020 (0.001)} & -0.607 (0.012) \\
    IntGrad (Pixel) & 0.928 (0.006) & 0.852 (0.013) & 0.293 (0.025) & 0.018 (0.002) & 0.008 (0.001) & 0.094 (0.014) \\
    IntGrad (Embed.) & 0.928 (0.006) & 0.852 (0.013) & 0.293 (0.025) & 0.018 (0.002) & 0.008 (0.001) & 0.094 (0.014) \\
    Vanilla (Pixel) & \textcolor{gray}{0.884 (0.009)} & 0.801 (0.013) & 0.656 (0.015) & 0.015 (0.001) & \textcolor{gray}{0.020 (0.001)} & -0.629 (0.013) \\
    Vanilla (Embed.) & \textcolor{gray}{0.902 (0.008)} & 0.752 (0.015) & 0.659 (0.015) & 0.012 (0.001) & \textcolor{gray}{0.025 (0.001)} & -0.632 (0.013) \\
    SmoothGrad (Pixel) & \textcolor{gray}{0.908 (0.009)} & 0.733 (0.015) & 0.659 (0.015) & \textcolor{gray}{0.011 (0.001)} & \textcolor{gray}{0.029 (0.002)} & -0.632 (0.013) \\
    SmoothGrad (Embed.) & \textcolor{gray}{0.917 (0.008)} & 0.756 (0.014) & 0.659 (0.015) & \textcolor{gray}{0.010 (0.001)} & \textcolor{gray}{0.025 (0.002)} & -0.632 (0.013) \\
    VarGrad (Pixel) & \textcolor{gray}{0.914 (0.008)} & 0.754 (0.014) & 0.648 (0.014) & \textcolor{gray}{0.011 (0.001)} & \textcolor{gray}{0.027 (0.002)} & -0.620 (0.012) \\
    VarGrad (Embed.) & \textcolor{gray}{0.921 (0.008)} & 0.777 (0.014) & 0.648 (0.014) & \textcolor{gray}{0.010 (0.001)} & \textcolor{gray}{0.023 (0.001)} & -0.618 (0.012) \\
    LRP & 0.938 (0.007) & 0.648 (0.016) & \textbf{0.673 (0.014)} & 0.014 (0.001) & \textcolor{gray}{0.024 (0.001)} & -0.623 (0.013) \\\midrule
    Leave-one-out & 0.956 (0.004) & 0.737 (0.019) & 0.221 (0.039) & 0.052 (0.004) & 0.003 (0.000) & 0.145 (0.025) \\
    RISE & 0.970 (0.003) & 0.602 (0.019) & 0.658 (0.015) & 0.092 (0.005) & 0.002 (0.000) & -0.628 (0.013) \\
    \textbf{ViT Shapley} & \textbf{0.981 (0.002)} & \textbf{0.457 (0.015)} & \textbf{0.694 (0.014)} & \textbf{0.198 (0.006)} & \textbf{0.001 (0.000)} & \textbf{0.641 (0.011)} \\\midrule
    Random & \textcolor{gray}{0.908 (0.008)} & \textcolor{gray}{0.907 (0.008)} & - & \textcolor{gray}{0.010 (0.001)} & \textcolor{gray}{0.010 (0.001)} & - \\
    \bottomrule
    \end{tabular}
}
\end{center}
\end{table}

%% file: figures/metrics_supplement_vitsmall.tex
\begin{table}[t]
\centering
\caption{Performance metrics
for ViT-Small on ImageNette. Methods that fail to outperform the random baseline are shown in gray, and the best results are shown in bold (accounting for 95\% confidence intervals).
} \label{tab:metrics_supplement_vitsmall}
\vskip 0.1in
\begin{center}
\begin{small}
{\tablesize
    \begin{tabular}{lcccccc}
    \toprule
     & \multicolumn{3}{c}{Target} & \multicolumn{3}{c}{Non-Target} \\
    \cmidrule(lr){2-4} \cmidrule(lr){5-7}
     & Ins. ($\uparrow$) & Del. ($\downarrow$) & Faith. ($\uparrow$) & Ins. ($\uparrow$) & Del. ($\downarrow$) & Faith. ($\uparrow$) \\
    \midrule
    Attention last & 0.949 (0.006) & 0.706 (0.014) & \textbf{0.781 (0.011)} & - & - & - \\
    Attention rollout & \textcolor{gray}{0.914 (0.007)} & 0.814 (0.013) & \textbf{0.787 (0.011)} & - & - & - \\\midrule
    GradCAM (LN) & \textcolor{gray}{0.908 (0.006)} & 0.898 (0.011) & 0.770 (0.010) & 0.027 (0.002) & 0.006 (0.000) & -0.740 (0.009) \\
    GradCAM (Attn.) & \textcolor{gray}{0.904 (0.009)} & 0.911 (0.009) & 0.724 (0.010) & \textcolor{gray}{0.008 (0.001)} & \textcolor{gray}{0.016 (0.001)} & -0.717 (0.009) \\
    IntGrad (Pixel) & 0.951 (0.006) & 0.908 (0.010) & 0.541 (0.022) & 0.011 (0.001) & 0.005 (0.001) & 0.384 (0.018) \\
    IntGrad (Embed.) & 0.951 (0.006) & 0.908 (0.010) & 0.541 (0.022) & 0.011 (0.001) & 0.005 (0.001) & 0.384 (0.018) \\
    Vanilla (Pixel) & \textcolor{gray}{0.905 (0.007)} & 0.830 (0.012) & \textbf{0.784 (0.010)} & 0.011 (0.001) & \textcolor{gray}{0.018 (0.001)} & -0.754 (0.009) \\
    Vanilla (Embed.) & \textcolor{gray}{0.921 (0.007)} & 0.793 (0.013) & \textbf{0.786 (0.011)} & 0.009 (0.001) & \textcolor{gray}{0.022 (0.001)} & -0.756 (0.009) \\
    SmoothGrad (Pixel) & \textcolor{gray}{0.943 (0.006)} & 0.744 (0.014) & \textbf{0.786 (0.011)} & \textcolor{gray}{0.007 (0.001)} & \textcolor{gray}{0.027 (0.001)} & -0.756 (0.009) \\
    SmoothGrad (Embed.) & 0.946 (0.006) & 0.753 (0.014) & \textbf{0.787 (0.011)} & \textcolor{gray}{0.006 (0.001)} & \textcolor{gray}{0.026 (0.001)} & -0.757 (0.009) \\
    VarGrad (Pixel) & 0.946 (0.006) & 0.766 (0.013) & 0.742 (0.010) & \textcolor{gray}{0.007 (0.001)} & \textcolor{gray}{0.025 (0.001)} & -0.709 (0.009) \\
    VarGrad (Embed.) & 0.947 (0.006) & 0.774 (0.013) & 0.758 (0.010) & \textcolor{gray}{0.007 (0.001)} & \textcolor{gray}{0.024 (0.001)} & -0.723 (0.009) \\
    LRP & 0.957 (0.005) & 0.695 (0.015) & \textbf{0.793 (0.010)} & \textcolor{gray}{0.007 (0.001)} & \textcolor{gray}{0.027 (0.001)} & -0.753 (0.009) \\\midrule
    Leave-one-out & 0.967 (0.002) & 0.855 (0.015) & 0.044 (0.042) & 0.024 (0.002) & 0.003 (0.000) & 0.079 (0.030) \\
    RISE & 0.976 (0.002) & 0.752 (0.018) & \textbf{0.787 (0.010)} & 0.049 (0.004) & 0.002 (0.000) & -0.755 (0.009) \\
    \textbf{ViT Shapley} & \textbf{0.982 (0.002)} & \textbf{0.591 (0.015)} & \textbf{0.801 (0.010)} & \textbf{0.130 (0.005)} & \textbf{0.001 (0.000)} & \textbf{0.747 (0.009)} \\\midrule
    Random & \textcolor{gray}{0.933 (0.006)} & \textcolor{gray}{0.934 (0.006)} & - & \textcolor{gray}{0.007 (0.001)} & \textcolor{gray}{0.007 (0.001)} & - \\
    \bottomrule
    \end{tabular}
}
\end{small}
\end{center}
\end{table}

%% file: figures/metrics_supplement_vitlarge.tex
\begin{table}[t]
\centering
\caption{Performance metrics
for ViT-Large on ImageNette. Methods that fail to outperform the random baseline are shown in gray, and the best results are shown in bold (accounting for 95\% confidence intervals).
} \label{tab:metrics_supplement_vitlarge}
\vskip 0.1in
\begin{center}
\begin{small}
{\tablesize
    \begin{tabular}{lcccccc}
    \toprule
     & \multicolumn{3}{c}{Target} & \multicolumn{3}{c}{Non-Target} \\
    \cmidrule(lr){2-4} \cmidrule(lr){5-7}
     & Ins. ($\uparrow$) & Del. ($\downarrow$) & Faith. ($\uparrow$) & Ins. ($\uparrow$) & Del. ($\downarrow$) & Faith. ($\uparrow$) \\
    \midrule
    Attention last & \textcolor{gray}{0.915 (0.005)} & 0.896 (0.008) & 0.401 (0.022) & - & - & - \\
    Attention rollout & \textcolor{gray}{0.929 (0.005)} & 0.928 (0.007) & \textbf{0.423 (0.023)} & - & - & - \\\midrule
    GradCAM (LN) & \textcolor{gray}{0.955 (0.004)} & 0.924 (0.009) & \textbf{0.424 (0.023)} & 0.012 (0.001) & \textcolor{gray}{0.005 (0.000)} & -0.372 (0.020) \\
    GradCAM (Attn) & \textcolor{gray}{0.935 (0.005)} & 0.930 (0.007) & 0.395 (0.022) & 0.007 (0.001) & \textcolor{gray}{0.006 (0.001)} & -0.351 (0.018) \\
    IntGrad (Pixel) & 0.965 (0.003) & \textcolor{gray}{0.947 (0.006)} & 0.195 (0.020) & \textcolor{gray}{0.006 (0.001)} & 0.004 (0.000) & 0.114 (0.013) \\
    IntGrad (Embed.) & 0.965 (0.003) & \textcolor{gray}{0.948 (0.006)} & 0.198 (0.020) & \textcolor{gray}{0.006 (0.001)} & 0.004 (0.000) & 0.117 (0.013) \\
    Vanilla (Pixel) & \textcolor{gray}{0.898 (0.007)} & 0.921 (0.007) & \textbf{0.411 (0.022)} & 0.012 (0.001) & \textcolor{gray}{0.008 (0.001)} & -0.370 (0.020) \\
    Vanilla (Embed.) & \textcolor{gray}{0.911 (0.006)} & 0.905 (0.009) & \textbf{0.415 (0.023)} & 0.010 (0.001) & \textcolor{gray}{0.010 (0.001)} & -0.373 (0.020) \\
    SmoothGrad (Pixel) & \textcolor{gray}{0.949 (0.004)} & 0.861 (0.011) & \textbf{0.424 (0.023)} & 0.006 (0.001) & \textcolor{gray}{0.014 (0.001)} & -0.382 (0.020) \\
    SmoothGrad (Embed.) & \textcolor{gray}{0.956 (0.004)} & 0.857 (0.011) & \textbf{0.423 (0.023)} & \textcolor{gray}{0.005 (0.000)} & \textcolor{gray}{0.015 (0.001)} & -0.380 (0.020) \\
    VarGrad (Pixel) & \textcolor{gray}{0.942 (0.005)} & 0.878 (0.010) & 0.387 (0.021) & 0.007 (0.001) & \textcolor{gray}{0.013 (0.001)} & -0.350 (0.018) \\
    VarGrad (Embed.) & \textcolor{gray}{0.944 (0.005)} & 0.875 (0.010) & 0.383 (0.020) & 0.007 (0.001) & \textcolor{gray}{0.013 (0.001)} & -0.343 (0.017) \\
    LRP & \textcolor{gray}{0.946 (0.004)} & 0.869 (0.010) & \textbf{0.424 (0.023)} & 0.006 (0.001) & \textcolor{gray}{0.013 (0.001)} & -0.381 (0.020) \\\midrule
    Leave-one-out & 0.970 (0.003) & 0.938 (0.007) & 0.123 (0.031) & 0.009 (0.001) & 0.003 (0.000) & 0.118 (0.020) \\
    RISE & 0.978 (0.002) & 0.882 (0.013) & \textbf{0.425 (0.023)} & 0.019 (0.003) & 0.002 (0.000) & -0.381 (0.020) \\
    \textbf{ViT Shapley} & \textbf{0.986 (0.002)} & \textbf{0.742 (0.013)} & \textbf{0.439 (0.023)} & \textbf{0.077 (0.003)} & \textbf{0.001 (0.000)} & \textbf{0.391 (0.019)} \\\midrule
    Random & \textcolor{gray}{0.957 (0.004)} & \textcolor{gray}{0.956 (0.004)} & - & \textcolor{gray}{0.005 (0.000)} & \textcolor{gray}{0.005 (0.000)} & - \\
    \bottomrule
    \end{tabular}
}
\end{small}
\end{center}
\end{table}

%% file: figures/metrics_supplement_direct_masking.tex
\begin{table}[t]
\centering
\caption{Performance metrics for target classes explanations
for a ViT classifier trained with random masking. Methods that fail to outperform the random baseline are shown in gray, and the best results are shown in bold (accounting for 95\% confidence intervals).
} \label{tab:metrics_supplement_direct_masking}
\vskip 0.1in
\begin{center}
{\tablesize
    \begin{tabular}{lccc}
    \toprule
     & \multicolumn{3}{c}{ImageNette} \\
    \cmidrule(lr){2-4}
     & Ins. ($\uparrow$) & Del. ($\downarrow$) & Faith. ($\uparrow$) \\
    \midrule
    Attention last & \textcolor{gray}{0.963 (0.003)} & 0.839 (0.011) & \textbf{0.593 (0.017)} \\
    Attention rollout & \textcolor{gray}{0.945 (0.004)} & 0.903 (0.008) & \textbf{0.597 (0.017)} \\\midrule
    GradCAM (LN) & \textcolor{gray}{0.955 (0.003)} & 0.898 (0.011) & \textbf{0.585 (0.016)} \\
    GradCAM (Attn) & \textcolor{gray}{0.945 (0.005)} & 0.942 (0.006) & 0.564 (0.016) \\
    IntGrad (Pixel) & 0.977 (0.002) & 0.933 (0.008) & 0.474 (0.017) \\
    IntGrad (Embed.) & 0.977 (0.002) & 0.933 (0.008) & 0.474 (0.017) \\
    Vanilla (Pixel) & \textcolor{gray}{0.934 (0.005)} & 0.901 (0.009) & \textbf{0.593 (0.017)} \\
    Vanilla (Embed.) & \textcolor{gray}{0.948 (0.004)} & 0.864 (0.011) & \textbf{0.595 (0.017)} \\
    SmoothGrad (Pixel) & 0.966 (0.004) & 0.809 (0.012) & \textbf{0.599 (0.017)} \\
    SmoothGrad (Embed.) & \textcolor{gray}{0.954 (0.004)} & \textcolor{gray}{0.950 (0.005)} & \textbf{0.597 (0.017)} \\
    VarGrad (Pixel) & 0.964 (0.004) & 0.826 (0.011) & \textbf{0.573 (0.016)} \\
    VarGrad (Embed.) & \textcolor{gray}{0.953 (0.005)} & \textcolor{gray}{0.954 (0.004)} & \textbf{0.594 (0.017)} \\
    LRP & 0.972 (0.002) & 0.806 (0.013) & \textbf{0.603 (0.017)} \\\midrule
    Leave-one-out & 0.977 (0.001) & 0.881 (0.014) & 0.134 (0.034) \\
    RISE & 0.982 (0.001) & 0.793 (0.017) & \textbf{0.599 (0.017)} \\
    \textbf{ViT Shapley} & \textbf{0.988 (0.001)} & \textbf{0.736 (0.013)} & \textbf{0.605 (0.017)} \\\midrule
    Random & \textcolor{gray}{0.957 (0.004)} & \textcolor{gray}{0.957 (0.004)} & - \\
    
    \bottomrule
    \end{tabular}
}
\end{center}
\end{table}

%% file: figures/insertion_curves_imagenette.tex
\begin{figure}[t]
    \centering
    \includegraphics[width=1.0\linewidth]{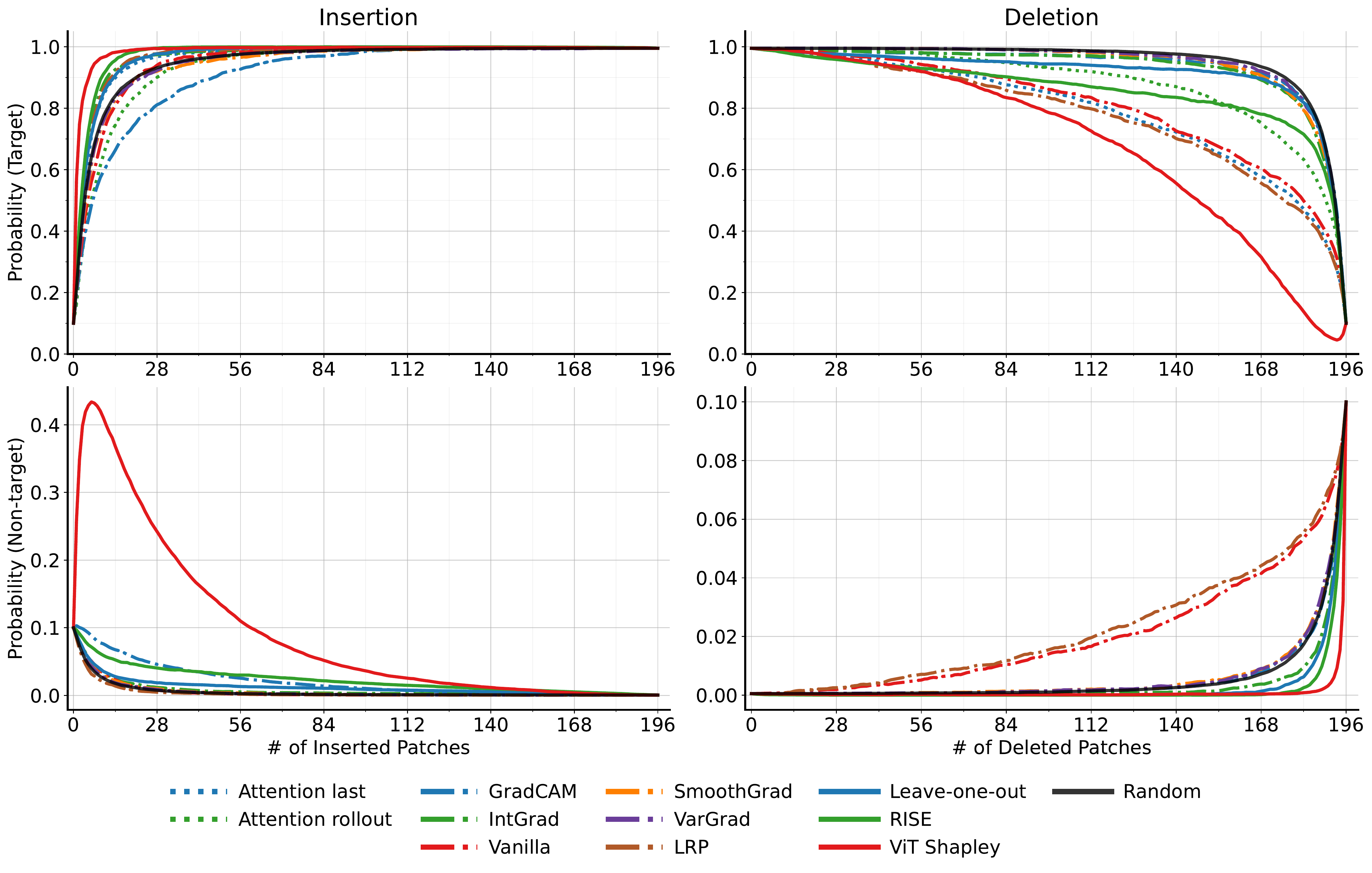}
    \caption{
    ImageNette average insertion/deletion curves. \textbf{Top:} the mean prediction probability for the target class as features are inserted or deleted in order of most to least important.
    \textbf{Bottom:} the mean prediction probability, averaged across all non-target classes.
    }
    \label{fig:insertion_curves_imagenette}
\end{figure}

%% file: figures/insertion_curves_mura.tex
\begin{figure}[t]
    \centering
    \includegraphics[width=1.0\linewidth]{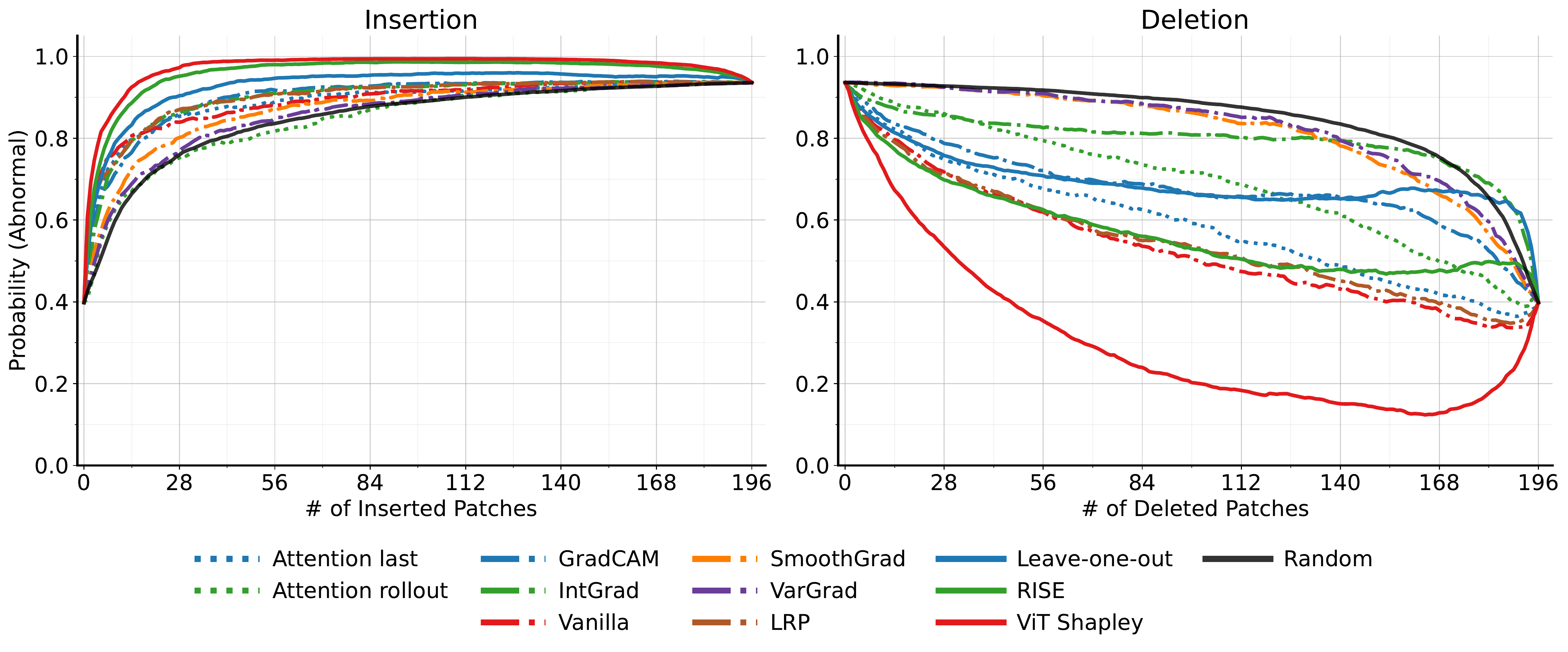}
    \caption{
    MURA average insertion/deletion curves. The mean prediction probability for the target class is plotted as features are inserted or deleted in order of most to least important.
    }
    \label{fig:insertion_curves_mura}
\end{figure}

%% file: figures/insertion_curves_pet.tex
\begin{figure}[t]
    \centering
    \includegraphics[width=1.0\linewidth]{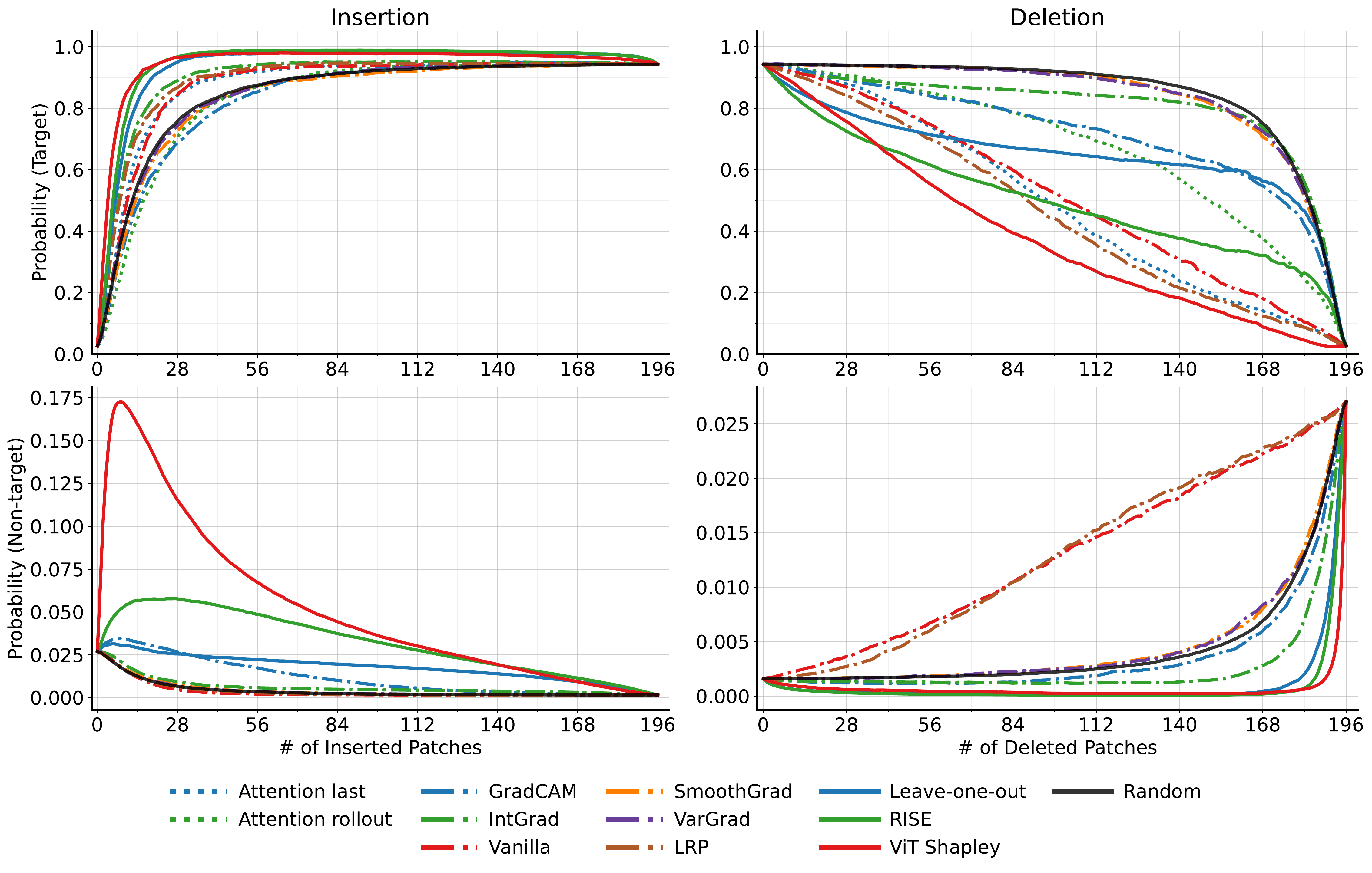}
    \caption{
    Pets average insertion/deletion curves.
    \textbf{Top:} the mean prediction probability for the target class as features are inserted or deleted in order of most to least important.
    \textbf{Bottom:} the mean prediction probability, averaged across all non-target classes.
    }
    \label{fig:insertion_curves_pet}
\end{figure}

%% file: figures/sensitivity_nontarget.tex
\begin{figure}[t]
    \centering
    \includegraphics[width=0.7\linewidth]{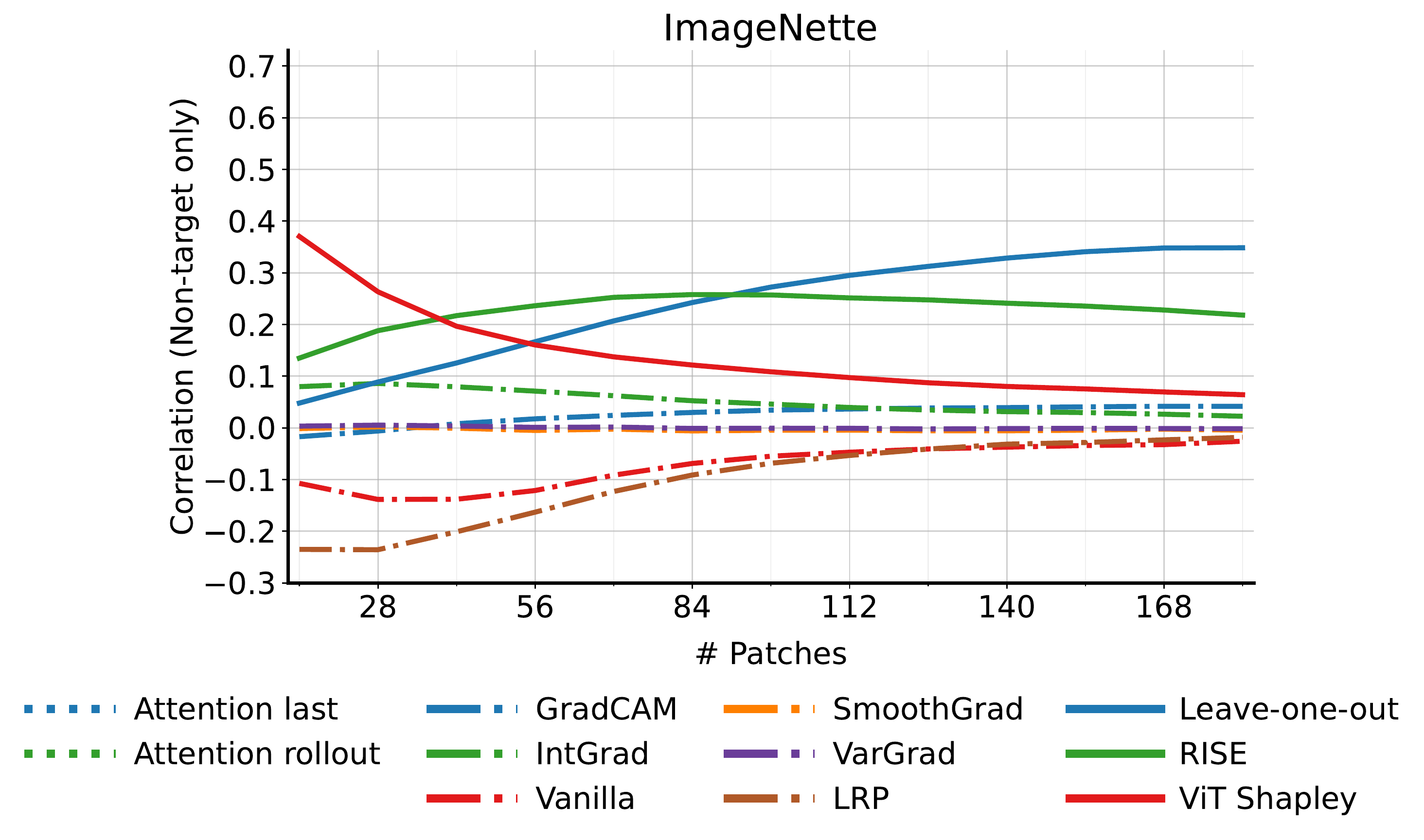}
    \caption{
    Sensitivity-n evaluation for non-target classes. The results are generated separately for each subset size and averaged across all non-target classes.
    }
    \label{fig:sensitivity_nontarget}
\end{figure}

%% file: figures/roar_all.tex
\begin{figure}[t]
    \figurespacing
    \centering
    \includegraphics[width=\linewidth]{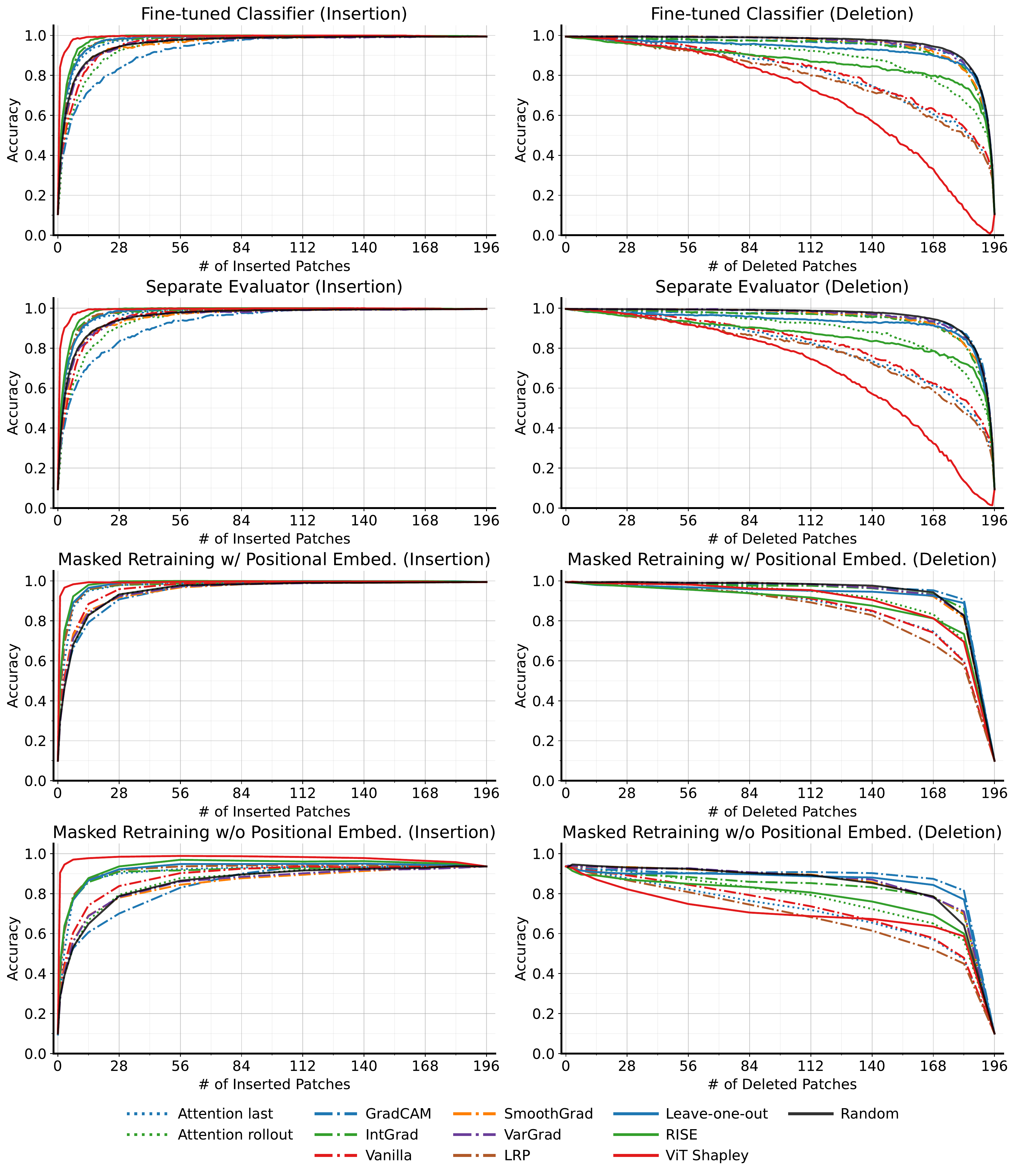}
    \caption{ImageNette accuracy when removing features in order of
    their importance, run with four evaluation strategies (fine-tuned classifier, separate evaluator, masked retraining
    and masked retraining without positional embeddings) on the ViT-Base architecture.
    \textbf{Left:} inserting features from
    most to least important.
    \textbf{Right:} removing features from
    most to least important.
    }
    \label{fig:roar_all}
\end{figure}

%% file: figures/roar_vitsmall.tex
\begin{figure}[t]
    \figurespacing
    \centering
    \includegraphics[width=\linewidth]{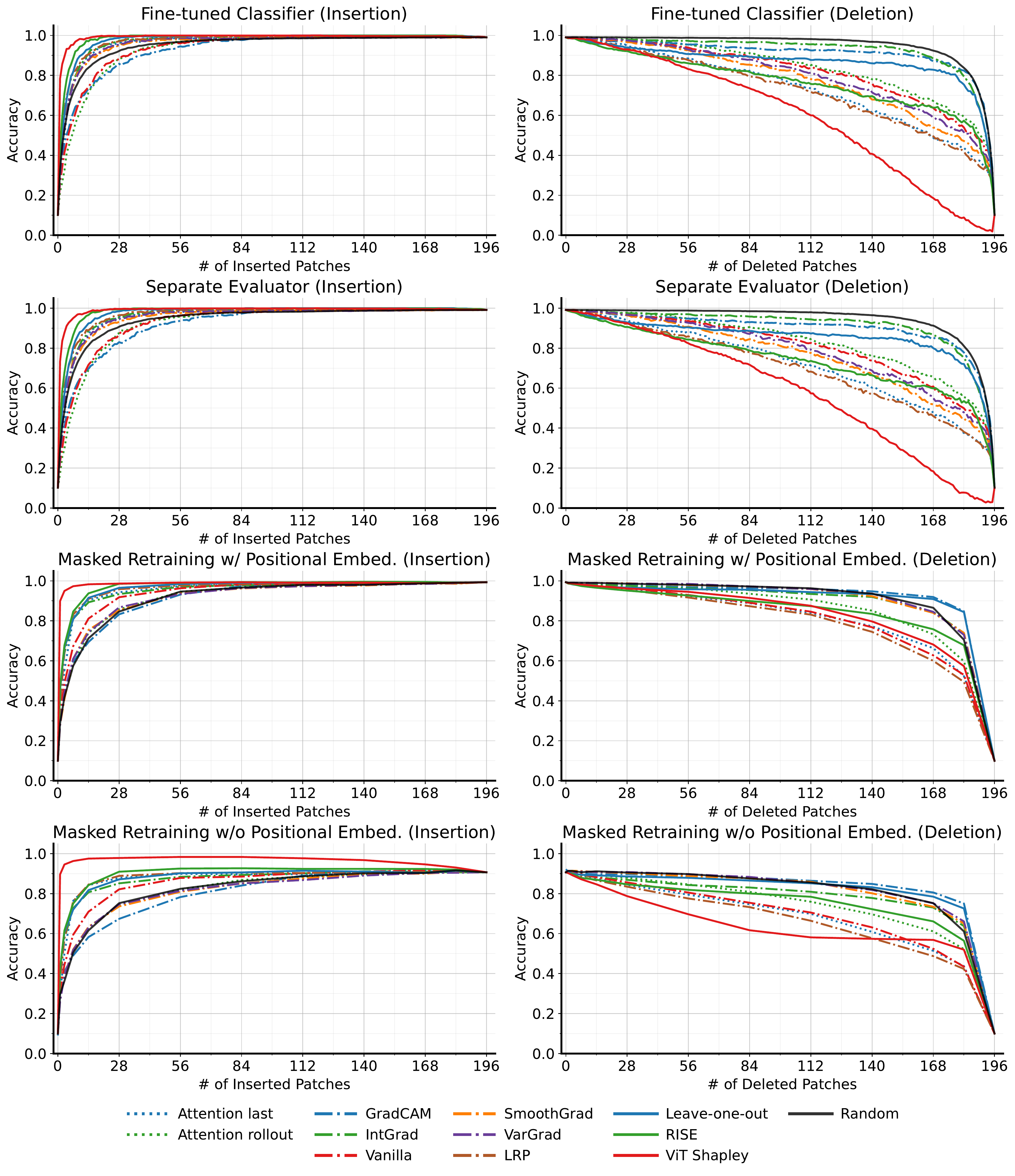}
    \caption{ImageNette accuracy when removing features in order of
    their importance, run with four evaluation strategies (fine-tuned classifier, separate evaluator, masked retraining
    and masked retraining without positional embeddings) on the ViT-Small architecture.
    \textbf{Left:} inserting features from
    most to least important.
    \textbf{Right:} removing features from
    most to least important.
    }
    \label{fig:roar_vitsmall}
    \vspace{-.2cm}
\end{figure}

%% file: figures/times.tex
\begin{table}[t]
\centering
\caption{
Time to generate explanations for a single sample (in milliseconds). For class-specific explanations, the running time involves generating explanations for all classes.
} \label{tab:times}
\vskip 0.1in
\begin{center}
\begin{small}
    \begin{tabular}{lrrr}
    \toprule
     & ImageNette & MURA & Pets\\
    \midrule
    Attention last & 6.8 & 6.4 & 6.2\\
    Attention rollout & 10.6 & 10.0 & 11.1\\\midrule
    GradCAM & 275.9 & 26.2 & 986.4 \\
    IntGrad & 1236.8 & 123.3 & 5004.9 \\
    Vanilla & 230.1 & 23.0 & 810.7 \\
    SmoothGrad & 1218.0 & 121.2 & 4854.8 \\
    VarGrad & 1218.9 & 121.3 & 4882.5 \\
    LRP & 1551.4 & 155.7 & 5583.5 \\\midrule
    Leave-one-out & 810.0 & 815.9 & 922.5 \\
    RISE & 8213.4 & 8171.1 & 8919.8 \\
    \textbf{ViT Shapley} & 10.1 & 10.2 & 10.8 \\
    \bottomrule
    \end{tabular}
\end{small}
\end{center}
\end{table}

%% file: figures/kernelshap_insertion_deletion.tex
\begin{table}[t]
\centering
\caption{Comparing the quality of Shapley value estimates obtained using ViT Shapley and KernelSHAP via insertion/deletion scores. 
} \label{tab:kernelshap_insertdelete}
\vskip 0.1in
\begin{center}
\begin{small}
\begin{tabular}{lcccccc}

\toprule
 & \multicolumn{2}{c}{Target} & \multicolumn{2}{c}{Non-target} \\
\cmidrule(lr){2-3} \cmidrule(lr){4-5}
 & Ins. (↑) & Del. (↓) & Ins. (↑) & Del. (↓) \\
\midrule
ViT Shapley & 0.985 (0.002) & 0.691 (0.014) & 0.093 (0.004) & 0.001 (0.000) \\
KernelSHAP & 0.990 (0.002) & 0.589 (0.053) & 0.158 (0.021) & 0.001 (0.000) \\
\bottomrule
\end{tabular}
\end{small}
\end{center}
\end{table}

%% file: figures/kernelshap_error.tex
\begin{figure}[t]
    \centering
    \includegraphics[width=\linewidth]{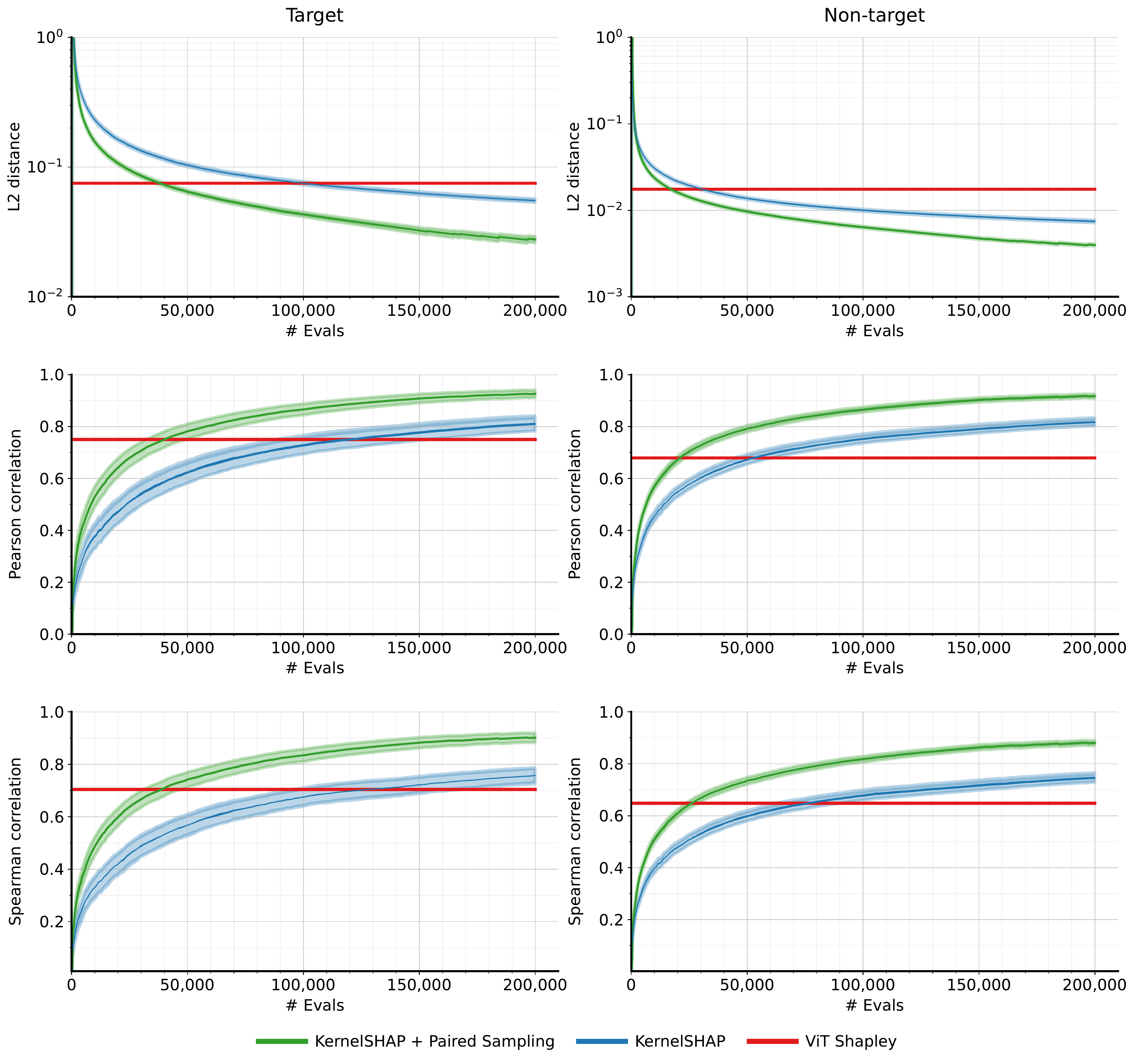}
    \caption{
    Comparing the quality of Shapley value estimates obtained by ViT Shapley and KernelSHAP. The shaded areas represent 95\% confidence intervals.
    }
    \label{fig:kernelshap_error}
\end{figure}

%% file: figures/kernelshap.tex
\begin{figure}[t]
    \centering
    \includegraphics[width=\linewidth]{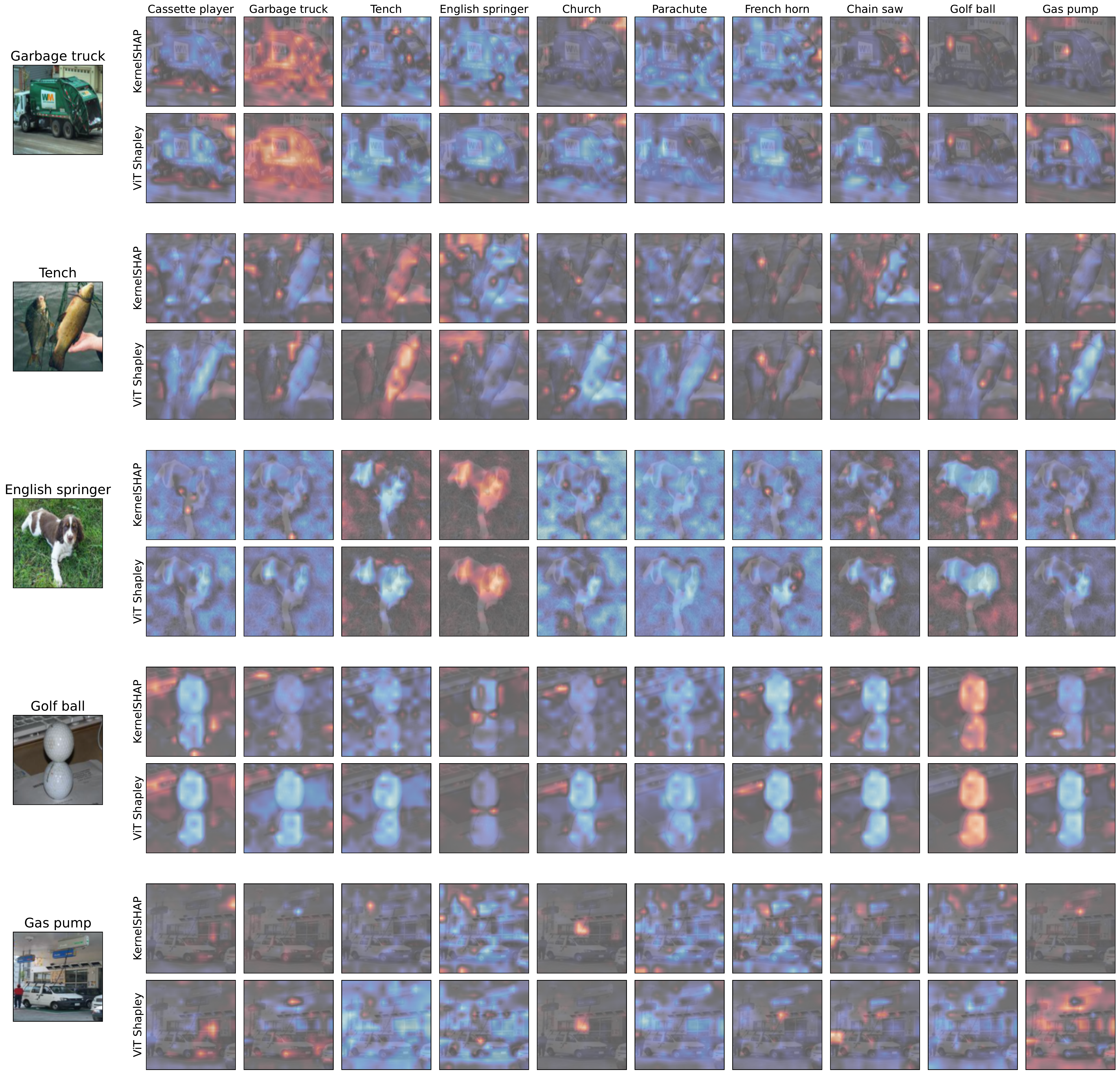}
    \caption{
    ViT Shapley vs. KernelSHAP comparison (ImageNette). 
    }
    \label{fig:kernelshap}
\end{figure}

%% file: figures/qualitative_baselines_imagenette_1.tex
\begin{figure}[t]
    \vspace{-.2cm}
    \centering
    \includegraphics[width=1.0\linewidth]{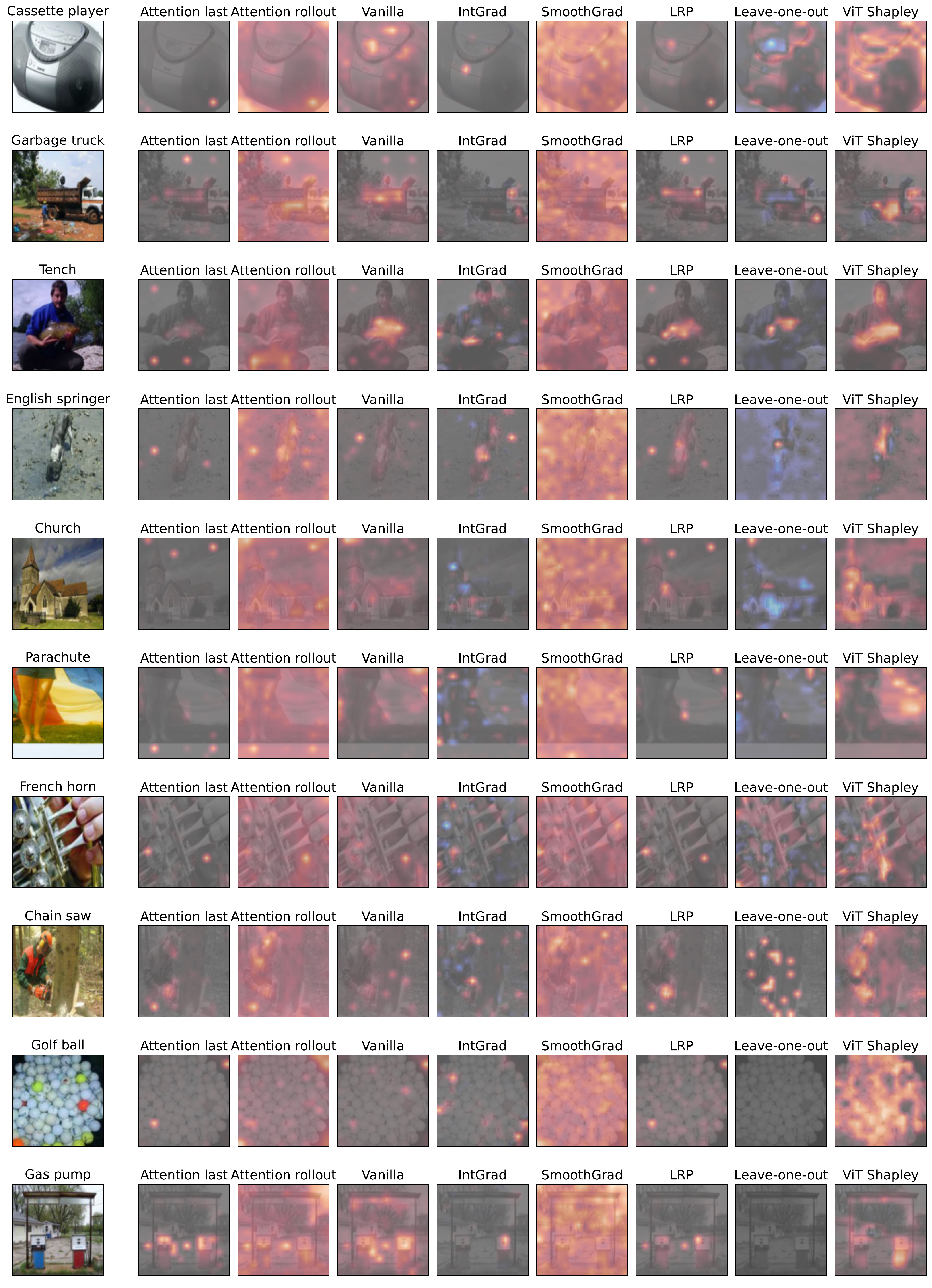}
    \caption{
    ViT Shapley vs. baselines comparison (ImageNette).
    }
    \label{fig:qualitative_baselines_imagenette_1}
\end{figure}

%% file: figures/qualitative_baselines_imagenette_2.tex
\begin{figure}[t]
    \vspace{-.2cm}
    \centering
    \includegraphics[width=1.0\linewidth]{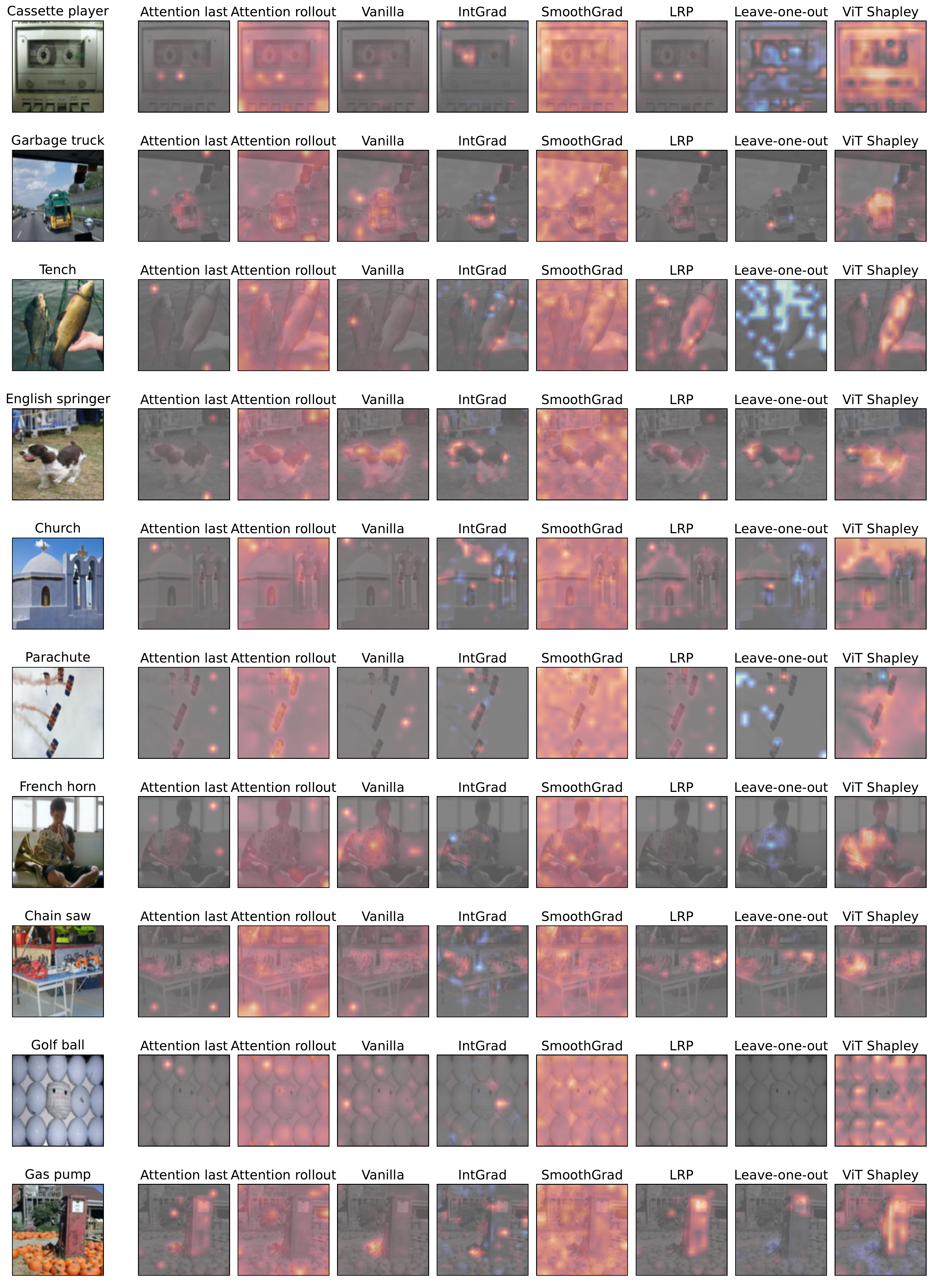}
    \caption{
    ViT Shapley vs. baselines comparison (ImageNette).
    }
    \label{fig:qualitative_baselines_imagenette_2}
\end{figure}

%% file: figures/qualitative_baselines_imagenette_3.tex
\begin{figure}[t]
    \vspace{-.2cm}
    \centering
    \includegraphics[width=1.0\linewidth]{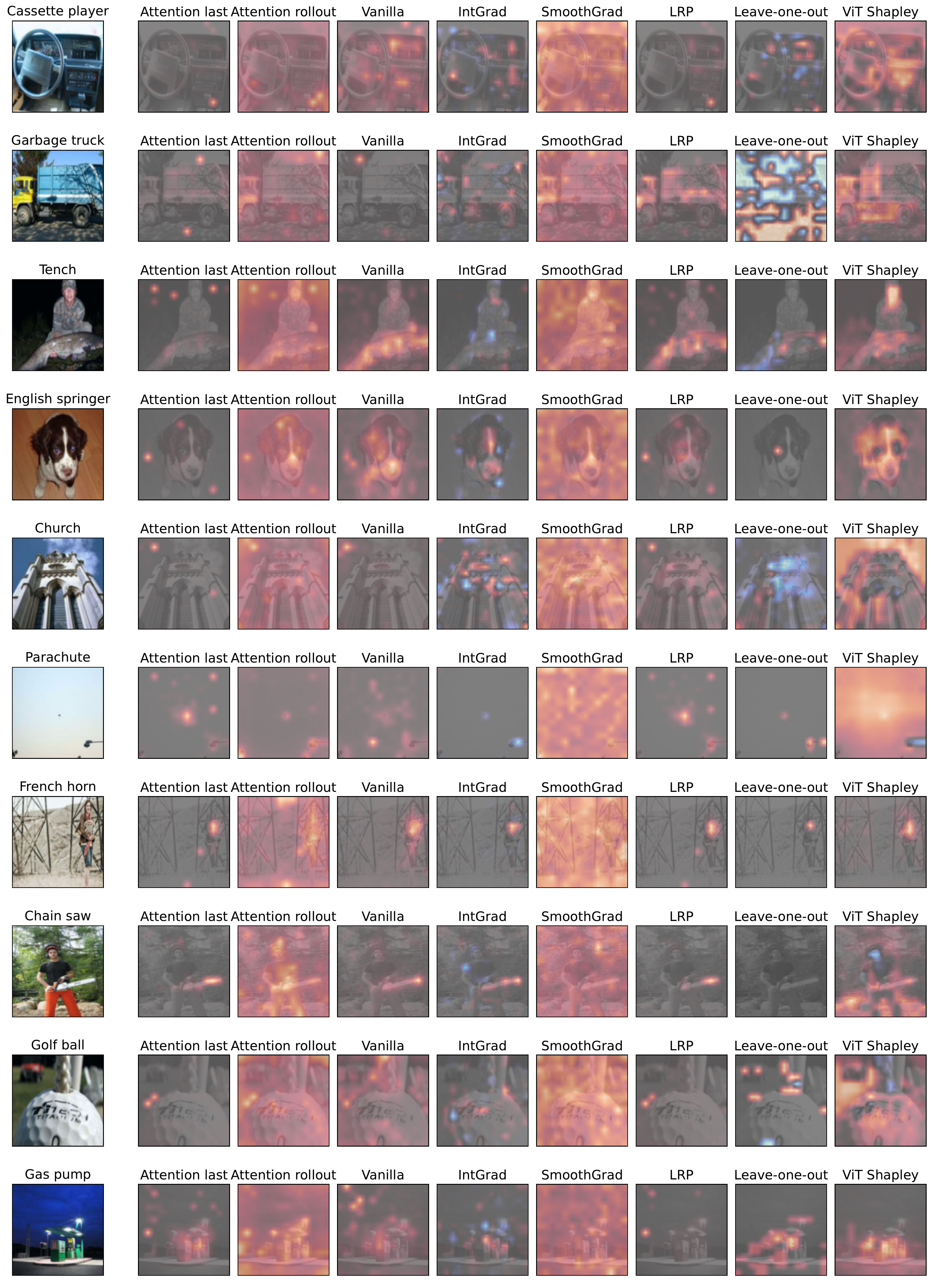}
    \caption{
    ViT Shapley vs. baselines comparison (ImageNette).
    }
    \label{fig:qualitative_baselines_imagenette_3}
\end{figure}

%% file: figures/qualitative_baselines_mura_1.tex
\begin{figure}[t]
    \vspace{-.2cm}
    \centering
    \includegraphics[width=1.0\linewidth]{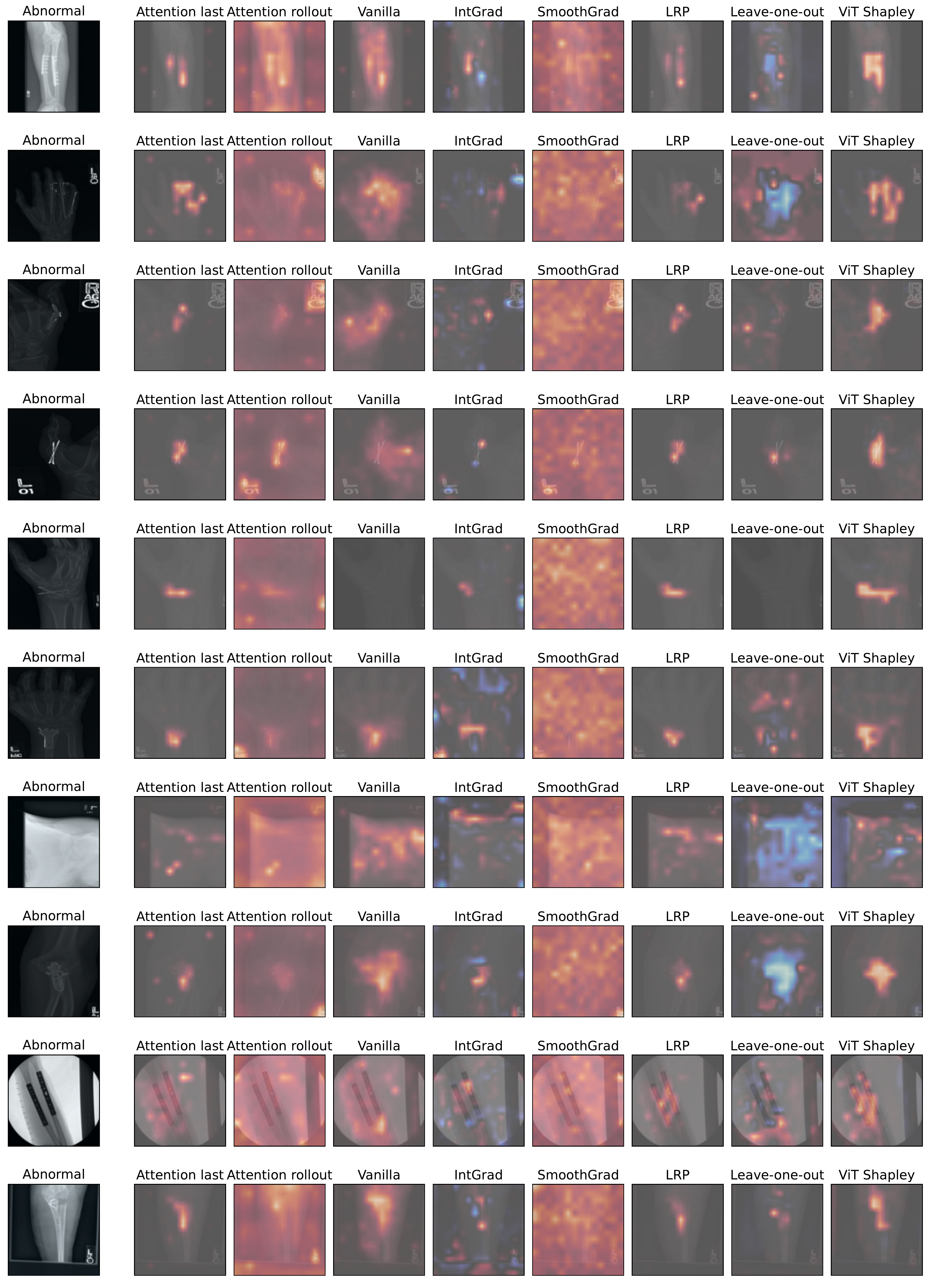}
    \caption{
    ViT Shapley vs. baselines comparison (MURA).
    }
    \label{fig:qualitative_baselines_MURA_1}
\end{figure}

%% file: figures/qualitative_baselines_mura_2.tex
\begin{figure}[t]
    \vspace{-.2cm}
    \centering
    \includegraphics[width=1.0\linewidth]{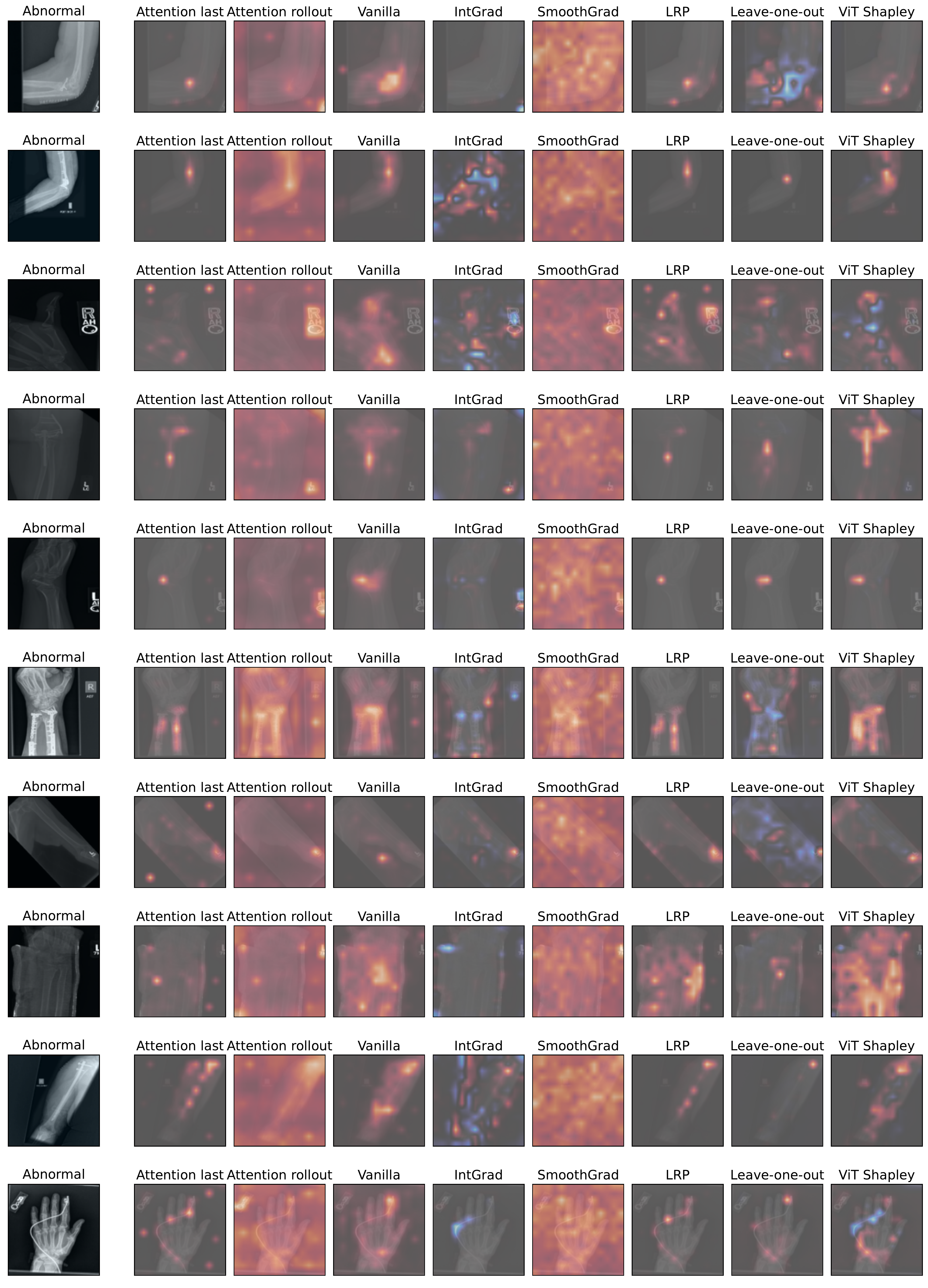}
    \caption{
    ViT Shapley vs. baselines comparison (MURA).
    }
    \label{fig:qualitative_baselines_MURA_2}
\end{figure}

%% file: figures/qualitative_baselines_mura_3.tex
\begin{figure}[t]
    \vspace{-.2cm}
    \centering
    \includegraphics[width=1.0\linewidth]{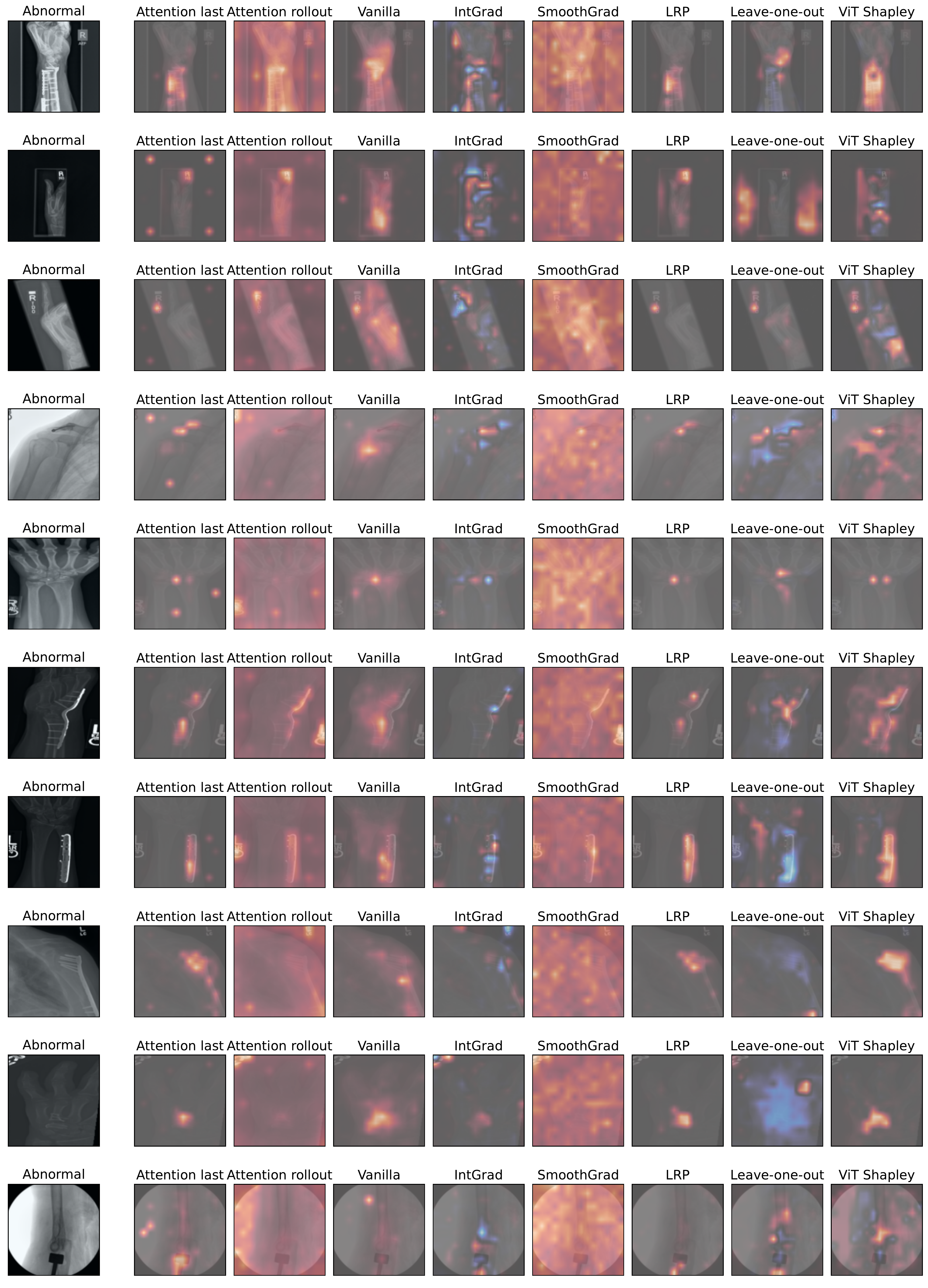}
    \caption{
    ViT Shapley vs. baselines comparison (MURA).
    }
    \label{fig:qualitative_baselines_MURA_3}
\end{figure}

%% file: figures/qualitative_baselines_pet_1.tex
\begin{figure}[t]
    \vspace{-.2cm}
    \centering
    \includegraphics[width=1.0\linewidth]{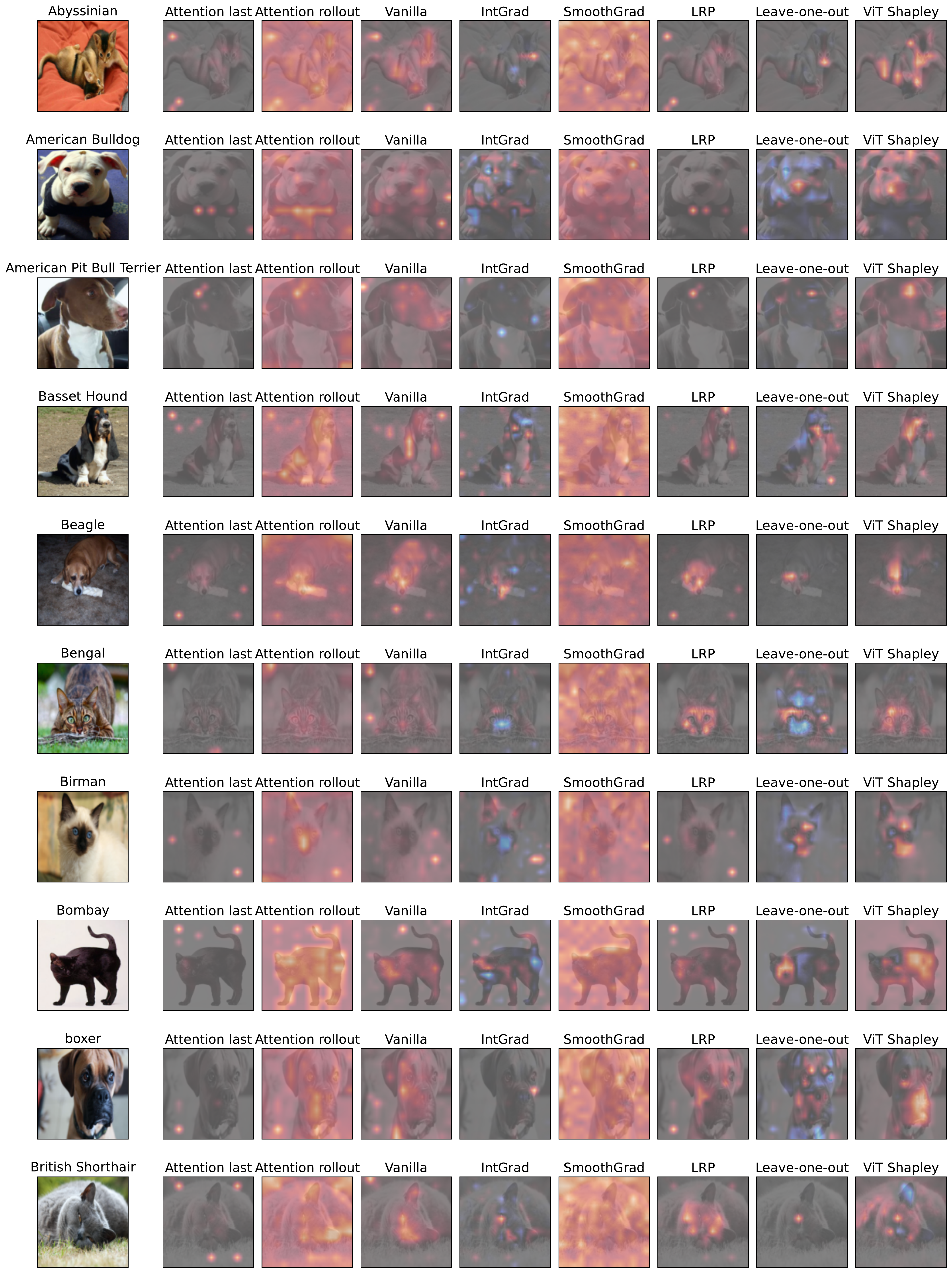}
    \caption{
    ViT Shapley vs. baselines comparison (Pets).
    }
    \label{fig:qualitative_baselines_pet_1}
\end{figure}

%% file: figures/qualitative_baselines_pet_2.tex
\begin{figure}[t]
    \vspace{-.2cm}
    \centering
    \includegraphics[width=1.0\linewidth]{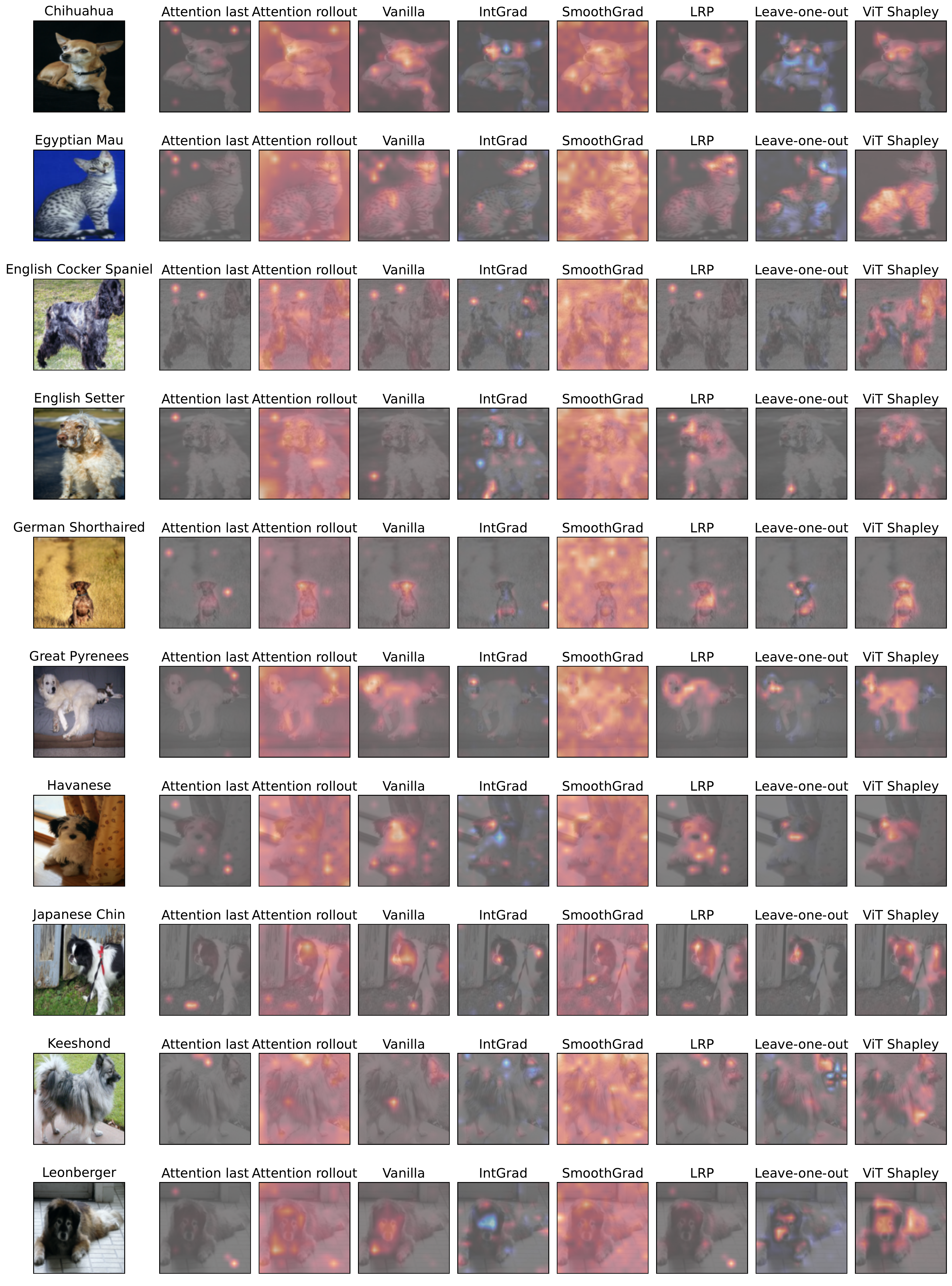}
    \caption{
    ViT Shapley vs. baselines comparison (Pets).
    }
    \label{fig:qualitative_baselines_pet_2}
\end{figure}

%% file: figures/qualitative_baselines_pet_3.tex
\begin{figure}[t]
    \vspace{-.2cm}
    \centering
    \includegraphics[width=1.0\linewidth]{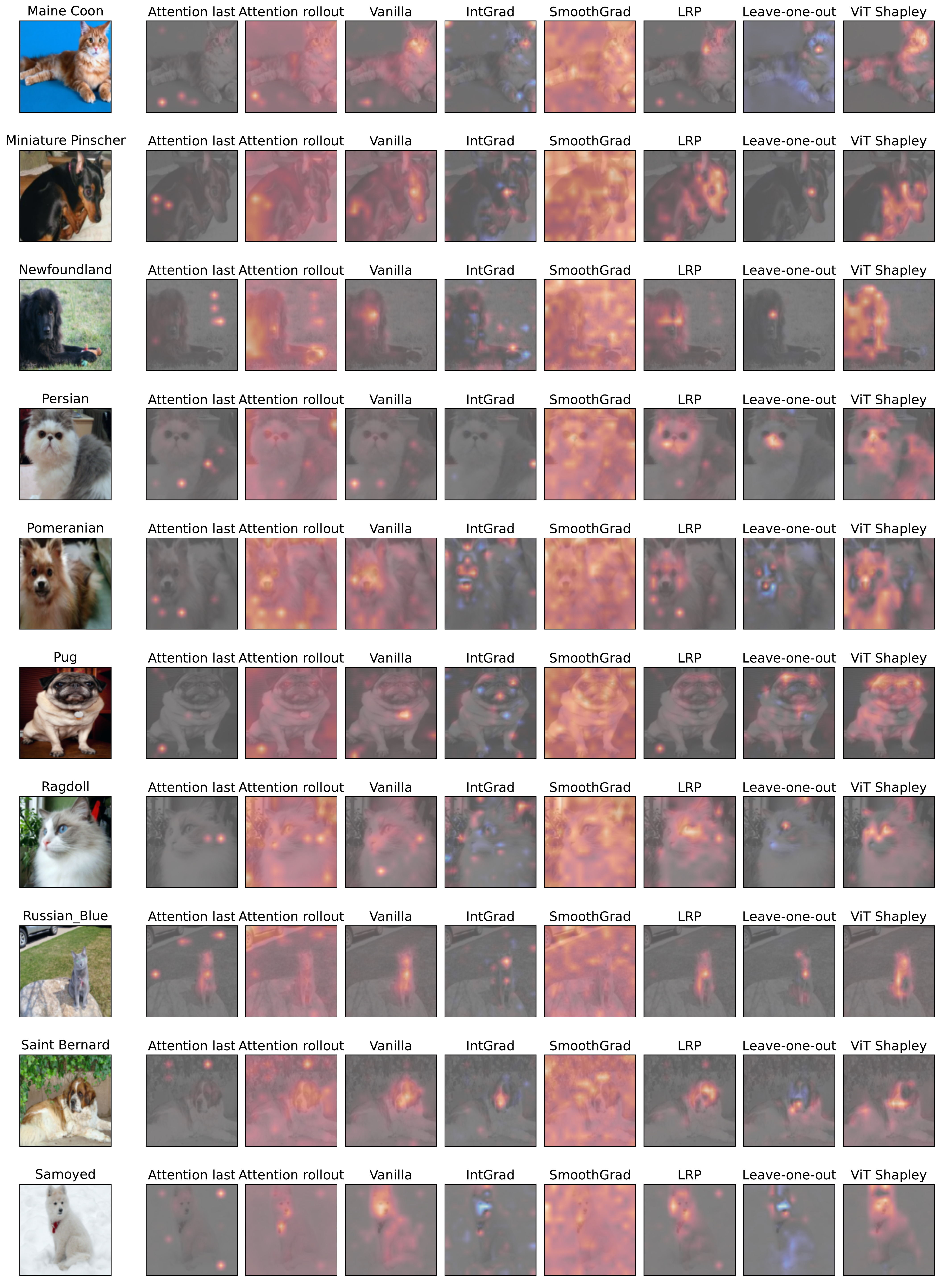}
    \caption{
    ViT Shapley vs. baselines comparison (Pets).
    }
    \label{fig:qualitative_baselines_pet_3}
\end{figure}

%% file: figures/qualitative_baselines_pet_4.tex
\begin{figure}[t]
    \vspace{-.2cm}
    \centering
    \includegraphics[width=1.0\linewidth]{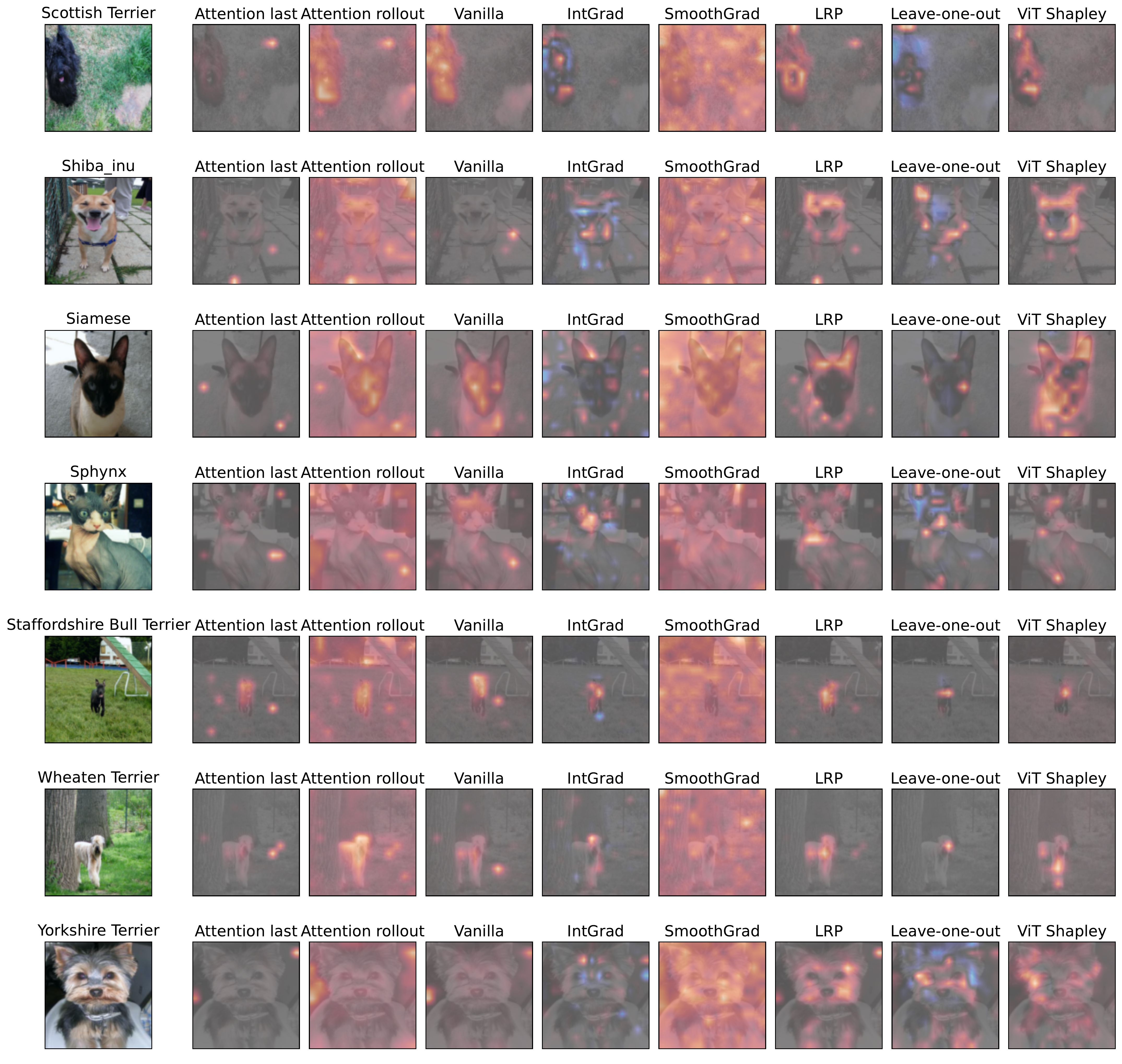}
    \caption{
    ViT Shapley vs. baselines comparison (Pets).
    }
    \label{fig:qualitative_baselines_pet_4}
\end{figure}

%% file: figures/qualitative_baselines_nontarget.tex
\begin{figure}[t]
     \vspace{-.2cm}
    \centering
    \includegraphics[width=0.3\linewidth]{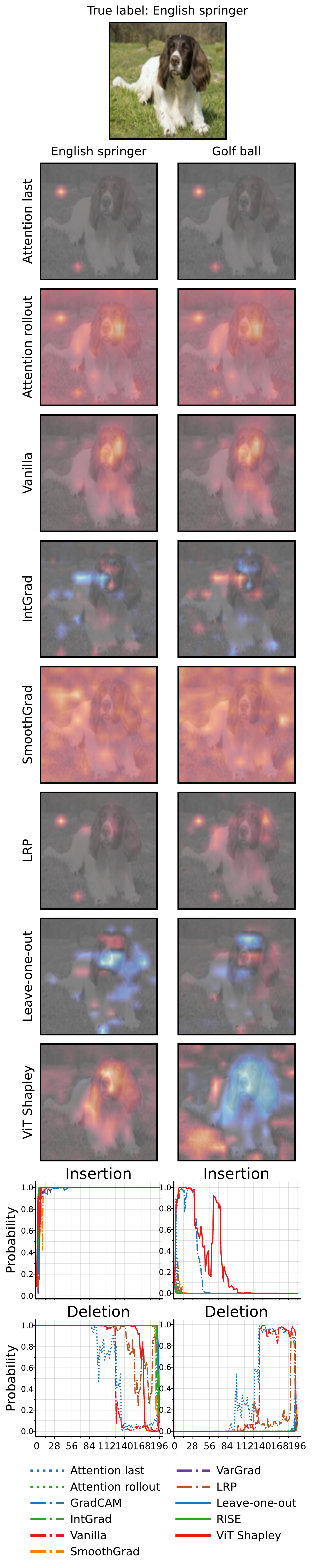}
    \includegraphics[width=0.3\linewidth]{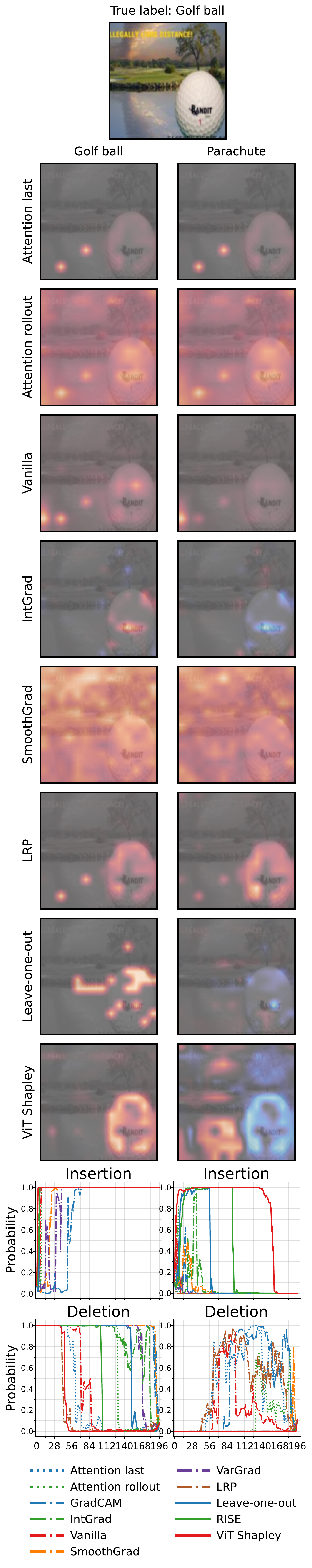}
    \includegraphics[width=0.3\linewidth]{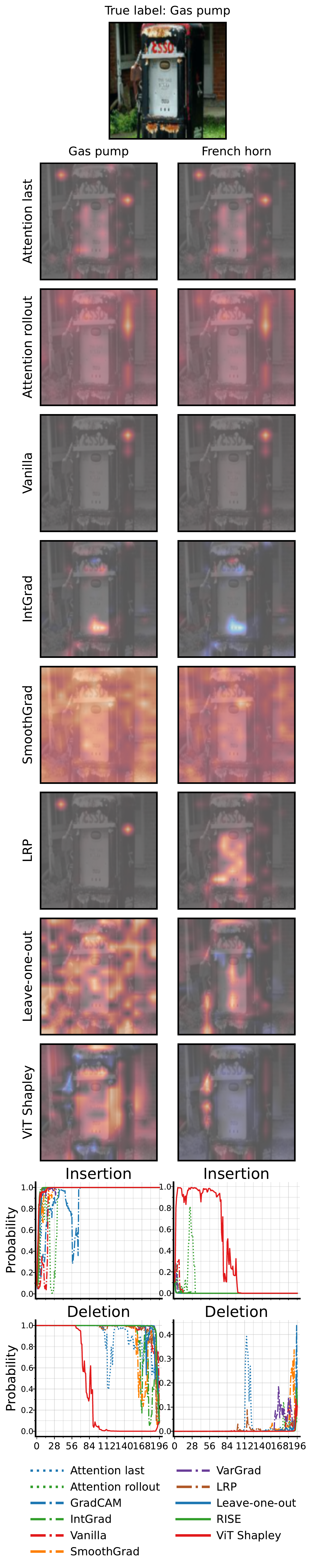}
    \caption{
    ViT Shapley vs. baselines for non-target classes (ImageNette).
    \textbf{Left:} ViT Shapley shows that the grass in the background provides evidence for the golf ball class.
    \textbf{Middle:} ViT Shapley shows that the sky and its reflection on the water provide evidence for the parachute class.
    \textbf{Right:} ViT Shapley shows that the metallic gas pump handle provides evidence for the French horn class.
    }
    \label{fig:qualitative_baselines_nontarget}
\end{figure}

%% file: main.bbl
\begin{thebibliography}{}

\bibitem[Abnar and Zuidema, 2020]{abnar2020quantifying}
Abnar, S. and Zuidema, W. (2020).
\newblock Quantifying attention flow in transformers.
\newblock In {\em Proceedings of the 58th Annual Meeting of the Association for
  Computational Linguistics}, pages 4190--4197.

\bibitem[Adebayo et~al., 2018]{adebayo2018sanity}
Adebayo, J., Gilmer, J., Muelly, M., Goodfellow, I., Hardt, M., and Kim, B.
  (2018).
\newblock Sanity checks for saliency maps.
\newblock {\em Advances in Neural Information Processing Systems}, 31.

\bibitem[Agarwal and Nguyen, 2020]{agarwal2020explaining}
Agarwal, C. and Nguyen, A. (2020).
\newblock Explaining image classifiers by removing input features using
  generative models.
\newblock In {\em Proceedings of the Asian Conference on Computer Vision}.

\bibitem[Ancona et~al., 2018]{ancona2018towards}
Ancona, M., Ceolini, E., {\"O}ztireli, C., and Gross, M. (2018).
\newblock Towards better understanding of gradient-based attribution methods
  for deep neural networks.
\newblock In {\em International Conference on Learning Representations}.

\bibitem[Bach et~al., 2015]{bach2015pixel}
Bach, S., Binder, A., Montavon, G., Klauschen, F., M{\"u}ller, K.-R., and
  Samek, W. (2015).
\newblock On pixel-wise explanations for non-linear classifier decisions by
  layer-wise relevance propagation.
\newblock {\em PloS One}, 10(7):e0130140.

\bibitem[Beyer et~al., 2022]{beyer2022better}
Beyer, L., Zhai, X., and Kolesnikov, A. (2022).
\newblock Better plain {ViT} baselines for {ImageNet}-1k.
\newblock {\em arXiv preprint arXiv:2205.01580}.

\bibitem[Bhatt et~al., 2021]{bhatt2021evaluating}
Bhatt, U., Weller, A., and Moura, J.~M. (2021).
\newblock Evaluating and aggregating feature-based model explanations.
\newblock In {\em International Conference on International Joint Conferences
  on Artificial Intelligence}.

\bibitem[Boyd et~al., 2004]{boyd2004convex}
Boyd, S., Boyd, S.~P., and Vandenberghe, L. (2004).
\newblock {\em Convex Optimization}.
\newblock Cambridge University Press.

\bibitem[Brown et~al., 2020]{brown2020language}
Brown, T., Mann, B., Ryder, N., Subbiah, M., Kaplan, J.~D., Dhariwal, P.,
  Neelakantan, A., Shyam, P., Sastry, G., Askell, A., et~al. (2020).
\newblock Language models are few-shot learners.
\newblock {\em Advances in Neural Information Processing Systems},
  33:1877--1901.

\bibitem[Castro et~al., 2009]{castro2009polynomial}
Castro, J., G{\'o}mez, D., and Tejada, J. (2009).
\newblock Polynomial calculation of the {S}hapley value based on sampling.
\newblock {\em Computers \& Operations Research}, 36(5):1726--1730.

\bibitem[Charnes et~al., 1988]{charnes1988extremal}
Charnes, A., Golany, B., Keane, M., and Rousseau, J. (1988).
\newblock Extremal principle solutions of games in characteristic function
  form: core, {C}hebychev and {S}hapley value generalizations.
\newblock In {\em Econometrics of Planning and Efficiency}, pages 123--133.
  Springer.

\bibitem[Chefer et~al., 2021]{chefer2021transformer}
Chefer, H., Gur, S., and Wolf, L. (2021).
\newblock Transformer interpretability beyond attention visualization.
\newblock In {\em Proceedings of the IEEE/CVF Conference on Computer Vision and
  Pattern Recognition}, pages 782--791.

\bibitem[Chen et~al., 2022]{chen2022algorithms}
Chen, H., Covert, I.~C., Lundberg, S.~M., and Lee, S.-I. (2022).
\newblock Algorithms to estimate shapley value feature attributions.
\newblock {\em arXiv preprint arXiv:2207.07605}.

\bibitem[Clark et~al., 2019]{clark2019does}
Clark, K., Khandelwal, U., Levy, O., and Manning, C.~D. (2019).
\newblock What does {BERT} look at? an analysis of {BERT}'s attention.
\newblock {\em arXiv preprint arXiv:1906.04341}.

\bibitem[Covert and Lee, 2021]{covert2021improving}
Covert, I. and Lee, S.-I. (2021).
\newblock Improving {KernelSHAP}: Practical {S}hapley value estimation using
  linear regression.
\newblock In {\em International Conference on Artificial Intelligence and
  Statistics}, pages 3457--3465. PMLR.

\bibitem[Covert et~al., 2021]{covert2021explaining}
Covert, I., Lundberg, S., and Lee, S.-I. (2021).
\newblock Explaining by removing: A unified framework for model explanation.
\newblock {\em Journal of Machine Learning Research}, 22(209):1--90.

\bibitem[Covert et~al., 2020]{covert2020understanding}
Covert, I., Lundberg, S.~M., and Lee, S.-I. (2020).
\newblock Understanding global feature contributions with additive importance
  measures.
\newblock {\em Advances in Neural Information Processing Systems},
  33:17212--17223.

\bibitem[Deng et~al., 2009]{deng2009imagenet}
Deng, J., Dong, W., Socher, R., Li, L.-J., Li, K., and Fei-Fei, L. (2009).
\newblock Imagenet: A large-scale hierarchical image database.
\newblock In {\em IEEE Conference on Computer Vision and Pattern Recognition},
  pages 248--255. IEEE.

\bibitem[Devlin et~al., 2018]{devlin2018bert}
Devlin, J., Chang, M.-W., Lee, K., and Toutanova, K. (2018).
\newblock Bert: Pre-training of deep bidirectional transformers for language
  understanding.
\newblock {\em arXiv preprint arXiv:1810.04805}.

\bibitem[Dosovitskiy et~al., 2020]{dosovitskiy2020image}
Dosovitskiy, A., Beyer, L., Kolesnikov, A., Weissenborn, D., Zhai, X.,
  Unterthiner, T., Dehghani, M., Minderer, M., Heigold, G., Gelly, S., et~al.
  (2020).
\newblock An image is worth 16x16 words: Transformers for image recognition at
  scale.
\newblock In {\em International Conference on Learning Representations}.

\bibitem[Ethayarajh and Jurafsky, 2021]{ethayarajh2021attention}
Ethayarajh, K. and Jurafsky, D. (2021).
\newblock Attention flows are {S}hapley value explanations.
\newblock {\em arXiv preprint arXiv:2105.14652}.

\bibitem[Fong and Vedaldi, 2017]{fong2017interpretable}
Fong, R.~C. and Vedaldi, A. (2017).
\newblock Interpretable explanations of black boxes by meaningful perturbation.
\newblock In {\em Proceedings of the IEEE International Conference on Computer
  Vision}, pages 3429--3437.

\bibitem[Frye et~al., 2020]{frye2020shapley}
Frye, C., de~Mijolla, D., Begley, T., Cowton, L., Stanley, M., and Feige, I.
  (2020).
\newblock Shapley explainability on the data manifold.
\newblock In {\em International Conference on Learning Representations}.

\bibitem[Ghorbani and Zou, 2019]{ghorbani2019data}
Ghorbani, A. and Zou, J. (2019).
\newblock Data {S}hapley: Equitable valuation of data for machine learning.
\newblock In {\em International Conference on Machine Learning}, pages
  2242--2251. PMLR.

\bibitem[Ghorbani and Zou, 2020]{ghorbani2020neuron}
Ghorbani, A. and Zou, J.~Y. (2020).
\newblock Neuron {S}hapley: Discovering the responsible neurons.
\newblock {\em Advances in Neural Information Processing Systems},
  33:5922--5932.

\bibitem[Gildenblat and contributors, 2021]{jacobgilpytorchcam}
Gildenblat, J. and contributors (2021).
\newblock {PyTorch} library for {CAM} methods.
\newblock \url{https://github.com/jacobgil/pytorch-grad-cam}.

\bibitem[He et~al., 2021]{he2021masked}
He, K., Chen, X., Xie, S., Li, Y., Doll{\'a}r, P., and Girshick, R. (2021).
\newblock Masked autoencoders are scalable vision learners.
\newblock {\em arXiv preprint arXiv:2111.06377}.

\bibitem[Hooker et~al., 2019]{hooker2019benchmark}
Hooker, S., Erhan, D., Kindermans, P.-J., and Kim, B. (2019).
\newblock A benchmark for interpretability methods in deep neural networks.
\newblock {\em Advances in Neural Information Processing Systems}, 32.

\bibitem[Howard and Gugger, 2020]{howard2020fastai}
Howard, J. and Gugger, S. (2020).
\newblock Fast{AI}: A layered {API} for deep learning.
\newblock {\em Information}, 11(2):108.

\bibitem[Jain et~al., 2021]{jain2021missingness}
Jain, S., Salman, H., Wong, E., Zhang, P., Vineet, V., Vemprala, S., and Madry,
  A. (2021).
\newblock Missingness bias in model debugging.
\newblock In {\em International Conference on Learning Representations}.

\bibitem[Jain and Wallace, 2019]{jain2019attention}
Jain, S. and Wallace, B.~C. (2019).
\newblock Attention is not explanation.
\newblock {\em arXiv preprint arXiv:1902.10186}.

\bibitem[Jethani et~al., 2021a]{jethani2021have}
Jethani, N., Sudarshan, M., Aphinyanaphongs, Y., and Ranganath, R. (2021a).
\newblock Have we learned to explain?: How interpretability methods can learn
  to encode predictions in their interpretations.
\newblock In {\em International Conference on Artificial Intelligence and
  Statistics}, pages 1459--1467. PMLR.

\bibitem[Jethani et~al., 2021b]{jethani2021fastshap}
Jethani, N., Sudarshan, M., Covert, I.~C., Lee, S.-I., and Ranganath, R.
  (2021b).
\newblock {FastSHAP}: Real-time {S}hapley value estimation.
\newblock In {\em International Conference on Learning Representations}.

\bibitem[Jumper et~al., 2021]{jumper2021highly}
Jumper, J., Evans, R., Pritzel, A., Green, T., Figurnov, M., Ronneberger, O.,
  Tunyasuvunakool, K., Bates, R., {\v{Z}}{\'\i}dek, A., Potapenko, A., et~al.
  (2021).
\newblock Highly accurate protein structure prediction with {AlphaFold}.
\newblock {\em Nature}, 596(7873):583--589.

\bibitem[Kim et~al., 2018]{kim2018interpretability}
Kim, B., Wattenberg, M., Gilmer, J., Cai, C., Wexler, J., Viegas, F., et~al.
  (2018).
\newblock Interpretability beyond feature attribution: Quantitative testing
  with concept activation vectors ({TCAV}).
\newblock In {\em International Conference on Machine Learning}, pages
  2668--2677. PMLR.

\bibitem[Kokhlikyan et~al., 2020]{kokhlikyan2020captum}
Kokhlikyan, N., Miglani, V., Martin, M., Wang, E., Alsallakh, B., Reynolds, J.,
  Melnikov, A., Kliushkina, N., Araya, C., Yan, S., et~al. (2020).
\newblock Captum: A unified and generic model interpretability library for
  {PyTorch}.
\newblock {\em arXiv preprint arXiv:2009.07896}.

\bibitem[Liu et~al., 2021]{liu2021swin}
Liu, Z., Lin, Y., Cao, Y., Hu, H., Wei, Y., Zhang, Z., Lin, S., and Guo, B.
  (2021).
\newblock Swin transformer: Hierarchical vision transformer using shifted
  windows.
\newblock In {\em Proceedings of the IEEE/CVF International Conference on
  Computer Vision}, pages 10012--10022.

\bibitem[Loshchilov and Hutter, 2018]{loshchilov2018decoupled}
Loshchilov, I. and Hutter, F. (2018).
\newblock Decoupled weight decay regularization.
\newblock In {\em International Conference on Learning Representations}.

\bibitem[Lundberg et~al., 2020]{lundberg2020local}
Lundberg, S.~M., Erion, G., Chen, H., DeGrave, A., Prutkin, J.~M., Nair, B.,
  Katz, R., Himmelfarb, J., Bansal, N., and Lee, S.-I. (2020).
\newblock From local explanations to global understanding with explainable {AI}
  for trees.
\newblock {\em Nature Machine Intelligence}, 2(1):2522--5839.

\bibitem[Lundberg and Lee, 2017]{lundberg2017unified}
Lundberg, S.~M. and Lee, S.-I. (2017).
\newblock A unified approach to interpreting model predictions.
\newblock {\em Advances in Neural Information Processing Systems},
  30:4765--4774.

\bibitem[Naseer et~al., 2021]{naseer2021intriguing}
Naseer, M.~M., Ranasinghe, K., Khan, S.~H., Hayat, M., Shahbaz~Khan, F., and
  Yang, M.-H. (2021).
\newblock Intriguing properties of vision transformers.
\newblock {\em Advances in Neural Information Processing Systems}, 34.

\bibitem[Olah et~al., 2017]{olah2017feature}
Olah, C., Mordvintsev, A., and Schubert, L. (2017).
\newblock Feature visualization.
\newblock {\em Distill}, 2(11):e7.

\bibitem[Parkhi et~al., 2012]{parkhi2012cats}
Parkhi, O.~M., Vedaldi, A., Zisserman, A., and Jawahar, C. (2012).
\newblock Cats and dogs.
\newblock In {\em 2012 IEEE Conference on Computer Vision and Pattern
  Recognition}, pages 3498--3505. IEEE.

\bibitem[Petsiuk et~al., 2018]{petsiuk2018rise}
Petsiuk, V., Das, A., and Saenko, K. (2018).
\newblock {RISE}: Randomized input sampling for explanation of black-box
  models.
\newblock {\em arXiv preprint arXiv:1806.07421}.

\bibitem[Rajpurkar et~al., 2017]{rajpurkar2017mura}
Rajpurkar, P., Irvin, J., Bagul, A., Ding, D., Duan, T., Mehta, H., Yang, B.,
  Zhu, K., Laird, D., Ball, R.~L., et~al. (2017).
\newblock {MURA}: Large dataset for abnormality detection in musculoskeletal
  radiographs.
\newblock {\em arXiv preprint arXiv:1712.06957}.

\bibitem[Ribeiro et~al., 2016]{ribeiro2016should}
Ribeiro, M.~T., Singh, S., and Guestrin, C. (2016).
\newblock "{W}hy should {I} trust you?" {E}xplaining the predictions of any
  classifier.
\newblock In {\em Proceedings of the 22nd ACM SIGKDD International Conference
  on Knowledge Discovery and Data Mining}, pages 1135--1144.

\bibitem[Rogers et~al., 2020]{rogers2020primer}
Rogers, A., Kovaleva, O., and Rumshisky, A. (2020).
\newblock A primer in {BERT}ology: What we know about how {BERT} works.
\newblock {\em Transactions of the Association for Computational Linguistics},
  8:842--866.

\bibitem[Ronneberger et~al., 2015]{ronneberger2015u}
Ronneberger, O., Fischer, P., and Brox, T. (2015).
\newblock U-{N}et: Convolutional networks for biomedical image segmentation.
\newblock In {\em International Conference on Medical Image Computing and
  Computer-Assisted Intervention}, pages 234--241. Springer.

\bibitem[Rudin, 2019]{rudin2019stop}
Rudin, C. (2019).
\newblock Stop explaining black box machine learning models for high stakes
  decisions and use interpretable models instead.
\newblock {\em Nature Machine Intelligence}, 1(5):206--215.

\bibitem[Ruiz et~al., 1998]{ruiz1998family}
Ruiz, L.~M., Valenciano, F., and Zarzuelo, J.~M. (1998).
\newblock The family of least square values for transferable utility games.
\newblock {\em Games and Economic Behavior}, 24(1-2):109--130.

\bibitem[Saporta et~al., 2021]{saporta2021benchmarking}
Saporta, A., Gui, X., Agrawal, A., Pareek, A., Truong, S.~Q., Nguyen, C.~D.,
  Ngo, V.-D., Seekins, J., Blankenberg, F.~G., Ng, A.~Y., et~al. (2021).
\newblock Benchmarking saliency methods for chest x-ray interpretation.
\newblock {\em medRxiv}, pages 2021--02.

\bibitem[Selvaraju et~al., 2017]{selvaraju2017grad}
Selvaraju, R.~R., Cogswell, M., Das, A., Vedantam, R., Parikh, D., and Batra,
  D. (2017).
\newblock Grad-{CAM}: Visual explanations from deep networks via gradient-based
  localization.
\newblock In {\em Proceedings of the IEEE International Conference on Computer
  Vision}, pages 618--626.

\bibitem[Serrano and Smith, 2019]{serrano2019attention}
Serrano, S. and Smith, N.~A. (2019).
\newblock Is attention interpretable?
\newblock In {\em Proceedings of the 57th Annual Meeting of the Association for
  Computational Linguistics}, pages 2931--2951.

\bibitem[Shapley, 1953]{shapley1953value}
Shapley, L.~S. (1953).
\newblock A value for n-person games.
\newblock {\em Contributions to the Theory of Games}, 2(28):307--317.

\bibitem[Shrikumar et~al., 2016]{shrikumar2016not}
Shrikumar, A., Greenside, P., Shcherbina, A., and Kundaje, A. (2016).
\newblock Not just a black box: Learning important features through propagating
  activation differences.
\newblock {\em arXiv preprint arXiv:1605.01713}.

\bibitem[Simon and Vincent, 2020]{simon2020projected}
Simon, G. and Vincent, T. (2020).
\newblock A projected stochastic gradient algorithm for estimating {S}hapley
  value applied in attribute importance.
\newblock In {\em International Cross-Domain Conference for Machine Learning
  and Knowledge Extraction}, pages 97--115. Springer.

\bibitem[Simonyan et~al., 2013]{simonyan2013deep}
Simonyan, K., Vedaldi, A., and Zisserman, A. (2013).
\newblock Deep inside convolutional networks: Visualising image classification
  models and saliency maps.
\newblock {\em arXiv preprint arXiv:1312.6034}.

\bibitem[Smilkov et~al., 2017]{smilkov2017smoothgrad}
Smilkov, D., Thorat, N., Kim, B., Vi{\'e}gas, F., and Wattenberg, M. (2017).
\newblock {SmoothGrad}: Removing noise by adding noise.
\newblock {\em arXiv preprint arXiv:1706.03825}.

\bibitem[{\v{S}}trumbelj and Kononenko, 2010]{vstrumbelj2010efficient}
{\v{S}}trumbelj, E. and Kononenko, I. (2010).
\newblock An efficient explanation of individual classifications using game
  theory.
\newblock {\em Journal of Machine Learning Research}, 11:1--18.

\bibitem[Sundararajan et~al., 2017]{sundararajan2017axiomatic}
Sundararajan, M., Taly, A., and Yan, Q. (2017).
\newblock Axiomatic attribution for deep networks.
\newblock In {\em International Conference on Machine Learning}, pages
  3319--3328. PMLR.

\bibitem[Touvron et~al., 2021]{touvron2021training}
Touvron, H., Cord, M., Douze, M., Massa, F., Sablayrolles, A., and J{\'e}gou,
  H. (2021).
\newblock Training data-efficient image transformers \& distillation through
  attention.
\newblock In {\em International Conference on Machine Learning}, pages
  10347--10357. PMLR.

\bibitem[Vaswani et~al., 2017]{vaswani2017attention}
Vaswani, A., Shazeer, N., Parmar, N., Uszkoreit, J., Jones, L., Gomez, A.~N.,
  Kaiser, {\L}., and Polosukhin, I. (2017).
\newblock Attention is all you need.
\newblock {\em Advances in Neural Information Processing Systems}, 30.

\bibitem[Vig et~al., 2020]{vig2020bertology}
Vig, J., Madani, A., Varshney, L.~R., Xiong, C., Rajani, N., et~al. (2020).
\newblock {BERT}ology meets biology: Interpreting attention in protein language
  models.
\newblock In {\em International Conference on Learning Representations}.

\bibitem[Wang et~al., 2020]{wang2020transformer}
Wang, Y., Mohamed, A., Le, D., Liu, C., Xiao, A., Mahadeokar, J., Huang, H.,
  Tjandra, A., Zhang, X., Zhang, F., et~al. (2020).
\newblock Transformer-based acoustic modeling for hybrid speech recognition.
\newblock In {\em ICASSP 2020-2020 IEEE International Conference on Acoustics,
  Speech and Signal Processing (ICASSP)}, pages 6874--6878. IEEE.

\bibitem[Waskom, 2021]{Waskom2021}
Waskom, M.~L. (2021).
\newblock Seaborn: statistical data visualization.
\newblock {\em Journal of Open Source Software}, 6(60):3021.

\bibitem[Wiegreffe and Pinter, 2019]{wiegreffe2019attention}
Wiegreffe, S. and Pinter, Y. (2019).
\newblock Attention is not not explanation.
\newblock In {\em Proceedings of the 2019 Conference on Empirical Methods in
  Natural Language Processing and the 9th International Joint Conference on
  Natural Language Processing (EMNLP-IJCNLP)}, pages 11--20.

\bibitem[Wightman, 2019]{rw2019timm}
Wightman, R. (2019).
\newblock {PyTorch} image models.
\newblock \url{https://github.com/rwightman/pytorch-image-models}.

\bibitem[Wortsman et~al., 2022]{wortsman2022model}
Wortsman, M., Ilharco, G., Gadre, S.~Y., Roelofs, R., Gontijo-Lopes, R.,
  Morcos, A.~S., Namkoong, H., Farhadi, A., Carmon, Y., Kornblith, S., et~al.
  (2022).
\newblock Model soups: averaging weights of multiple fine-tuned models improves
  accuracy without increasing inference time.
\newblock {\em arXiv preprint arXiv:2203.05482}.

\bibitem[Xu et~al., 2020]{xu2020attribution}
Xu, S., Venugopalan, S., and Sundararajan, M. (2020).
\newblock Attribution in scale and space.
\newblock In {\em Proceedings of the IEEE/CVF Conference on Computer Vision and
  Pattern Recognition}, pages 9680--9689.

\bibitem[Zeiler and Fergus, 2014]{zeiler2014visualizing}
Zeiler, M.~D. and Fergus, R. (2014).
\newblock Visualizing and understanding convolutional networks.
\newblock In {\em European Conference on Computer Vision}, pages 818--833.
  Springer.

\bibitem[Zhou et~al., 2022]{zhou2022feature}
Zhou, Y., Booth, S., Ribeiro, M.~T., and Shah, J. (2022).
\newblock Do feature attribution methods correctly attribute features?
\newblock In {\em Proceedings of the AAAI Conference on Artificial
  Intelligence}, volume~36, pages 9623--9633.

\end{thebibliography}
